\documentclass[10pt,journal,compsoc]{IEEEtran}

\ifCLASSOPTIONcompsoc
\usepackage[nocompress]{cite}
\else
\usepackage{cite}
\fi

\ifCLASSINFOpdf
\usepackage[pdftex]{graphicx}
\else
\usepackage[dvips]{graphicx}
\fi

\usepackage{amsmath,amsfonts,amssymb,amsthm}
\usepackage{array}

\usepackage[utf8]{inputenc}
\usepackage{times}
\usepackage{epsfig}
\usepackage{color}
\usepackage{caption}
\usepackage{subcaption}
\usepackage{xspace}
\usepackage{multirow}
\usepackage{booktabs}
\usepackage{graphbox}  
\usepackage{tabularx}
\usepackage{ragged2e}
\usepackage[lined,boxed]{algorithm2e}  %

\SetCommentSty{mycommfont}
\usepackage{soul}

\usepackage[pagebackref=true,breaklinks=true,letterpaper=true,colorlinks,bookmarks=false]{hyperref}

\newcommand{\bd}{\mathbf{d}}

\newcommand{\bh}{\mathbf{h}}

\newcommand{\bl}{\mathbf{l}}

\newcommand{\bn}{\mathbf{n}}

\newcommand{\bq}{\mathbf{q}}

\newcommand{\bt}{\mathbf{t}}

\newcommand{\bx}{\mathbf{x}}

\newcommand{\bomega}{\boldsymbol{\omega}}

\newcommand{\nR}{\mathbb{R}}

\newcommand{\cC}{\mathcal{C}}
\newcommand{\cD}{\mathcal{D}}

\newcommand{\cG}{\mathcal{G}}

\newcommand{\cI}{\mathcal{I}}

\newcommand{\cM}{\mathcal{M}}
\newcommand{\cN}{\mathcal{N}}

\newcommand{\cP}{\mathcal{P}}

\newcommand{\cR}{\mathcal{R}}

\newcommand{\cX}{\mathcal{X}}

\newcommand{\cZ}{\mathcal{Z}}

\newcommand{\figref}[1]{Fig.~\ref{#1}}
\newcommand{\secref}[1]{Section~\ref{#1}}
\newcommand{\algref}[1]{Algorithm~\ref{#1}}
\newcommand{\eqnref}[1]{Eq.~\eqref{#1}}
\newcommand{\tabref}[1]{Table~\ref{#1}}

\DeclareMathOperator*{\argmin}{argmin~}

\makeatletter
\DeclareRobustCommand\onedot{\futurelet\@let@token\@onedot}
\def\@onedot{.}
\def\eg{e.g\onedot} \def\Eg{E.g\onedot}
\def\ie{i.e\onedot}

\def\wrt{wrt\onedot}

\def\etal{et~al\onedot}

\makeatother

\newcommand{\boldparagraph}[1]{\vspace{0.2cm}\noindent{\bf #1:} }

\definecolor{darkgreen}{rgb}{0,0.7,0}
\definecolor{darkred}{rgb}{0.6,0,0}
\definecolor{blue}{rgb}{0.012,0.263,0.875}
\definecolor{lightblue}{rgb}{0.024,0.604,0.953}
\definecolor{red}{rgb}{0.898,0,0}
\definecolor{orange}{rgb}{1.,0.58,0.031}
\definecolor{green}{rgb}{0.361,0.675,0.176}

\begin{document}	
	\title{Towards Scalable Multi-View Reconstruction \\of Geometry and Materials}
	\author{Carolin~Schmitt$^{\ddag}\quad$
		Bo\v{z}idar~Anti\'{c}$\quad$
		Andrei~Neculai$\quad$
		Joo~Ho~Lee$^{\ddag}\quad$
		Andreas~Geiger
		\IEEEcompsocitemizethanks{
			\IEEEcompsocthanksitem This work has been submitted to the IEEE for possible publication. Copyright may be transferred without notice, after which this version may no longer be accessible.
			\IEEEcompsocthanksitem For this project, all authors were with the Autonomous Vision Group, University~of~Tübingen~and~Max~Planck~Institute~for~Intelligent~Systems,~Tübingen,~Germany. E-mail: \{firstname.lastname\}@tue.mpg.de
			\IEEEcompsocthanksitem Joo Ho Lee is now with Sogang~University,~Seoul,~South~Korea. 
			\IEEEcompsocthanksitem $^{\ddag}$ denotes corresponding authors. \newline
			Emails: carolin.schmitt@tue.mpg.de, jhleecs@sogang.ac.kr
	}}

	\markboth{}%
	{}
	
	\IEEEtitleabstractindextext{%
		\begin{justify}
	\begin{abstract}
		In this paper, we propose a novel method for joint recovery of camera pose, object geometry and spatially-varying Bidirectional Reflectance Distribution Function (svBRDF) of 3D scenes that exceed object-scale and hence cannot be captured with stationary light stages.
		The input are high-resolution RGB-D images captured by a mobile, hand-held capture system with point lights for active illumination.
		Compared to previous works that jointly estimate geometry and materials from a hand-held scanner, we formulate this problem using a single objective function that can be minimized using off-the-shelf gradient-based solvers.
		To facilitate scalability to large numbers of observation views and optimization variables, we introduce a distributed optimization algorithm that reconstructs 2.5D keyframe-based representations of the scene.
		A novel multi-view consistency regularizer effectively synchronizes neighboring keyframes such that the local optimization results allow for seamless integration into a globally consistent 3D model.
		We provide a study on the importance of each component in our formulation and show that our method compares favorably to baselines.
		We further demonstrate that our method accurately reconstructs various objects and materials and allows for expansion to spatially larger scenes. 		
		We believe that this work represents a significant step towards making geometry and material estimation from hand-held scanners scalable.
	\end{abstract}	
\end{justify}

	}
	
	\maketitle
	\IEEEdisplaynontitleabstractindextext
	\IEEEpeerreviewmaketitle

	\begin{figure*}
	\includegraphics[width=\linewidth]{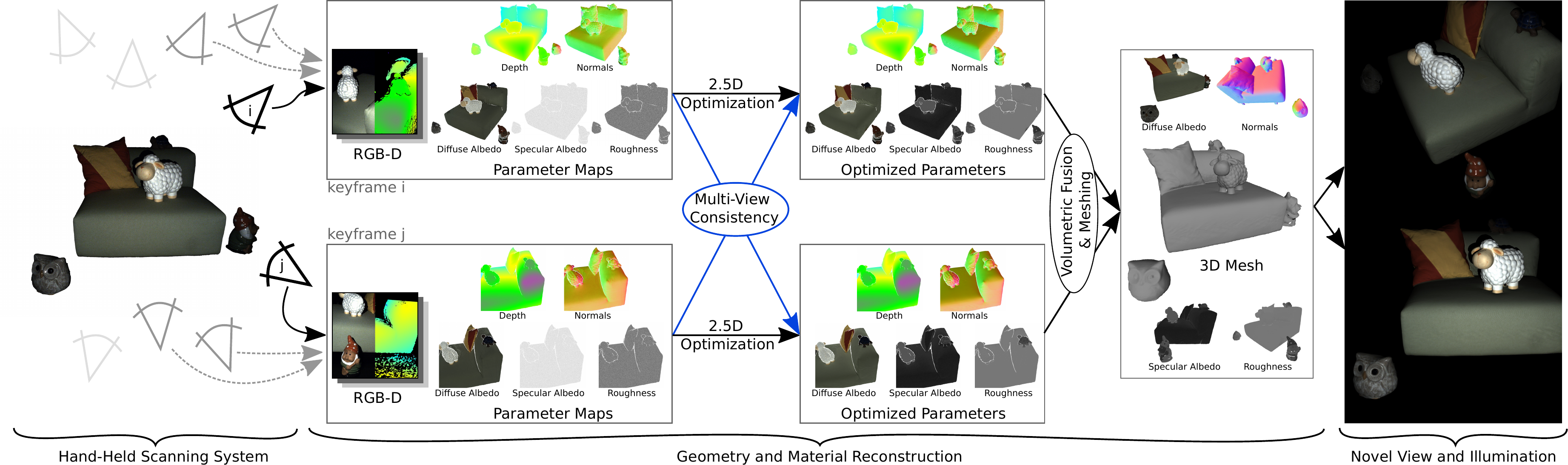}
	\caption{
		\textbf{Globally Consistent Material and Geometry Reconstruction.}
		Given RGB-D images from a mobile hand-held scanner (left), the proposed method uses local 2.5D representations to iteratively reconstruct globally consistent poses, geometry and material parameter maps that can be integreated into a 3D representation which features per voxel normals and material parameters (middle). 
		This allows for rendering novel views under unseen illumination (right).
		Our approach can handle both multiple-object scenes and very specular materials.
		}
		\label{fig:intro_teaser}
\end{figure*}

\section{Introduction}\label{sec:introduction}

As AR/VR technologies are emerging, there is an increasing demand for scanning casual scenes for immersive and interactive virtual experiences.
Both, for a high level of realism and to increase robustness to varying illumination, it is beneficial to reconstruct material and reflectance properties alongside geometry.
Yet, 3D shape and appearance estimation is very challenging due to sparse measurements, large computation and memory requirements and the interlaced and complex correlation of geometric and photometric entities.
Therefore, most works target only small objects which are often captured using complex light stages in a laboratory.
However, in order to generate photo-realistic replica of real environments for training embodied agents or telepresence, this is insufficient.
Instead, we require accurate geometry and material reconstructions for large, complex scenes and the ability to work with data from mobile scanners that capture scenes from arbitrary viewpoints. 

Ideally, object geometry and material properties are inferred jointly:
a good model of light transport allows for recovering geometric detail using shading cues.
An accurate shape model, in turn, facilitates the estimation of material properties.
This is particularly relevant for shiny surfaces and detailed geometries. 
Yet joint optimization of geometry and material from a handheld device poses an inverse rendering problem and is ill-posed and under-determined.
Existing approaches  assume fixed camera poses \cite{Bi2020CVPR,Lichy2021CVPR} or leverage sophisticated pipelines \cite{Higo2009ICCV,Georgoulis2014THREEDV,Nam2018SIGGRAPH,Li2021CVPR} which decompose the problem into smaller problems using multiple decoupled objectives and optimization algorithms that treat geometry and materials separately.
In this work, we provide a novel formulation for this problem which does not rely on sophisticated pipelines or decoupled objective functions.

In order to process larger multi-object scenes, a scalable scene representation is mandatory.
Unfortunately, reconstructing large scenes captured from many viewpoints at high resolution (\eg, 4K) quickly becomes intractable.
We therefore propose to use local 2.5D scene representations and an optimization scheme that encourages global consistency between them.
By optimizing in 2.5D, the proposed model has a constant memory footprint independent of the scene size and allows for reconstructing geometry and materials at larger scales, see \figref{fig:intro_teaser}.

\noindent We summarize the contributions of this paper as follows:
\begin{itemize}
	\item We demonstrate that joint optimization of camera pose, object geometry and materials is possible using a single objective function and off-the-shelf gradient-based solvers.
	\item We propose a distributed optimization scheme over a set of 2.5D scene representations that enables accurate integration of 2.5D reconstructions to full 3D models. 
	We show that despite overlapping fields of view, regularizing multi-view consistency is crucial to attain globally accurate reconstructions without visual artifacts.
	\item We provide a study on the importance of each component in our formulation and a comparison to multiple baselines.
	\item We demonstrate that our model can be used to reconstruct scenes exceeding object-level that include multiple objects with various different materials.
	\item We provide videos of our reconstructed models and make our source code and dataset publicly available at
	\href{https://sites.google.com/view/material-fusion/}{https://sites.google.com/view/material-fusion/}.
\end{itemize}

This journal paper is an extension of a conference paper published at CVPR 2020 \cite{Schmitt2020CVPR} which jointly estimates pose, geometry and svBRDF from handheld data in 2.5D.
In \secref{sec:results_baselines} we demonstrate that simple fusion of the 2.5D parameter maps produced by \cite{Schmitt2020CVPR} is insufficient to obtain an accurate and consistent 3D model. 
We therefore extend our previous model with a distributed multi-view optimization stage which enables fusion of 2.5D representations of geometry and materials into consistent 3D models.
 
Further, in comparison to \cite{Schmitt2020CVPR}, we 
1) model the specular BRDF parameters per pixel for a richer and more flexible material representation, 
2) refine the regularization terms to account for the new material model and better ensure proximity to the measurements, 
3) provide a thorough ablation study of our global multi-view consistent optimization scheme, 
4) add qualitative comparisons to \cite{Schmitt2020CVPR} and \cite{Nam2018SIGGRAPH}, and 
5) show reconstruction results for additional objects and scenes that demonstrate the scalability of our method.

	\section{Related work}
\label{sec:relwork}

We now discuss the most related work on geometry, material as well as joint geometry and material estimation.
We further provide an overview on geometry estimation at scale.

\subsection{Geometry Estimation}

\textbf{Multi-View Stereo (MVS)} reconstruction techniques \cite{Seitz1997CVPR,Lafarge2013PAMI,Ulusoy2015THREEDV,Vogiatzis2005CVPR,Kolmogorov2002ECCV,Furukawa2010PAMI,Schoenberger2016ECCV} recover the 3D geometry of an object from multiple input images by matching feature correspondences across views or by optimizing photo-consistency. As they ignore physical light transport, they cannot recover material properties. Furthermore, they are only able to recover geometry for surfaces which are sufficiently textured.

\textbf{Shape from Shading (SfS)} techniques exploit shading cues for reconstructing \cite{Ikeuchi1981AI,Zhang1999PAMI,Horn1970,Queau2017EMMCVPR,Queau2017ARXIV} or for refining \cite{Haefner2018CVPR,Wu2014SIGGRAPH,Zollhoefer2015SIGGRAPH,Maier2017ICCV} 3D geometry from one or multiple images by relating surface normals to image intensities through Lambert's law.
While early SfS approaches were restricted to objects made of a single Lambertian material, modern reincarnations of these models \cite{Barron2015PAMI,Oxholm2014CVPR,Lombardi2012CVPR} are also able to infer non-Lambertian materials and lighting. Unfortunately, reconstructing geometry from a single image is a highly ill-posed problem, requiring strong assumptions about the surface geometry. Moreover, textured objects often cause ambiguities as intensity changes can be caused by changes in either surface orientation or surface albedo.

\textbf{Photometric Stereo (PS)} approaches \cite{Woodham1980OE,Queau2016CVPR,Queau2015JMIV,Papadhimitri2014BMVC,Holroyd2008SIGGRAPH,Zhou2010ECCV,Tunwattanapong2013SIGGRAPH} assume three or more images captured with a static camera while varying illumination or object pose \cite{Lim2005ICCV,Simakov2003ICCV} to resolve the aforementioned ambiguities. In contrast to early PS approaches which often assumed orthographic cameras and distant light sources, newer works have considered the more practical setup of near light sources \cite{Logothetis2017CVPR,Queau2017SSVM,Xie2015CVPR,Liu2018ICCP} and perspective projection \cite{Mecca2014JIS,Queau2018JMIV,Mecca2014JISa}. To handle non-Lambertian surfaces, robust error functions have been suggested \cite{Queau2017CVPR,Queau2015SSVM} and the problem has been formulated using specularity-invariant image ratios \cite{Mecca2015CG,Mecca2016WACV,Mecca2016JIS,Chandraker2011CVPR}.
The advantages of PS (accurate normals) and MVS (global geometry) have also been combined by integrating normals from PS and geometry from MVS \cite{Lu2010CVPR,Nehab2005SIGGRAPH,Joshi2007ICCV,Fan2018IECON,Park2017PAMI,Shi2014THREEDV,Logothetis2019ICCV,Yoshiyasu2011CVPR} into a single consistent reconstruction.
However, many classical PS approaches are not capable of estimating material properties other than albedo and most PS approaches require a fixed camera which restricts their applicability to lab environments. In contrast, here we are interested in recovering shape and surface materials of larger scenes using a \textit{handheld} mobile scanner.

\subsection{Material Estimation}

\textbf{Intrinsic Image Decomposition} \cite{Barrow1978CVS,Gehler2011NIPS,Barron2015PAMI,Chen2013ICCV} is the problem of decomposing an image into its material-dependent and light-dependent properties. However, only a small portion of the 3D physical process is captured by these models and strong regularizers must be exploited to solve the task.
A more accurate description of the reflective properties of materials is provided by the Bidirectional Reflectance Distribution Function (BRDF) \cite{Nicodemus1992}.

For \textbf{known 3D geometry}, the BRDF can be measured using specialized light stages or gantries \cite{Matusik2003SIGGRAPH,Nielsen2015SIGGRAPH,Lensch2003SIGGRAPH,Schwartz2013EUROGRAPHICS,Holroyd2010SIGGRAPH}.
While this setup leads to accurate reflectance estimates, it is typically expensive, stationary and only works for objects of limited size.
In contrast, recent works have demonstrated that reflectance properties of flat surfaces can be acquired using an ordinary mobile phone \cite{Albert2018EUROGRAPHICS,Riviere2016CGF,Aittala2015SIGGRAPH,Xu2016SIGGRAPH}.
While data collection is easy and practical, these techniques are designed for capturing flat textured surfaces and do not generalize to objects with more complex geometries.

More closely aligned with our goals are approaches that estimate parametric BRDF models for scenes with known geometry based on sparse measurements of the BRDF space
\cite{Dong2014SIGGRAPH,Park2018THREEDV,Wu2015CGF,Wu2016ARXIVa,Wu2016VCG,Melou2017SSVM,Melou2018JMIV,Yu1999SIGGRAPH,Zhou2016SIGGRAPH,Haefner2021THREEDV}.
While we also estimate a parametric BRDF model and assume only sparse measurements of the BRDF domain, we \textit{jointly} optimize for camera pose, object geometry and material parameters. 
As our experiments show, joint optimization allows for recovering  fine geometric structures not present in the initial reconstruction while at the same time improving material estimates compared to a sequential treatment of both tasks.

\subsection{Joint Geometry and Material Estimation}

Several works have addressed the problem of jointly inferring geometry and materials.
By integrating shading cues with multi-view constraints and an accurate model of materials and light transport, this approach has the potential to deliver the most accurate results.
However, joint optimization of all relevant quantities is a challenging task.
Several works have considered extensions of the classic PS setting \cite{Goldman2010PAMI,Alldrin2008CVPR,Hertzmann2005PAMI,Ackermann2010ECCVWORK,Hui2017PAMI,Peng2017ICCVWORK,Birkbeck2006ECCV,Esteban2008PAMI,Zuo2017ICCV,Zhou2013CVPR,Xia2016SIGGRAPH,Bi2020CVPR}.
While some of these approaches consider multiple viewpoints and/or estimate spatially varying BRDFs, all of them require multiple images from the \textbf{same or known viewpoints} as input.
In contrast, we are interested in jointly estimating geometry and materials from \textbf{mobile scanning systems}, enabling applications outside laboratory environments.

In 2011, \cite{Ren2011SIGGRAPH} proposed to exploit low-cost and handheld scanning devices such as a flash camera for reconstructing both BRDFs and geometry from multi-view images.
Like subsequent works \cite{Hui2017ICCV, Kim2017ICCV, Li2018ECCVa, Cheng2021CVPR} they are restricted to flat surfaces, simple shapes or uniform materials.
For \textbf{single objects or small scenes} the following works estimate materials alongside geometry:
Higo \etal~\cite{Higo2009ICCV} estimate a depth, normal and diffuse albedo map of a Lambertian object by graph-cut-based plane sweeping.
Georgoulis \etal \cite{Georgoulis2014THREEDV} optimize 3D geometry and a data-driven BRDF model in an alternating fashion.
Nam \etal~\cite{Nam2018SIGGRAPH} refine a subdivided mesh by alternatively updating positions, normals, and material properties.
Finally, Li \etal~\cite{Li2021CVPR} iteratively optimize for 3D geometry, reflectance, camera pose and environment lighting. %
All these methods decompose the problem into smaller problems by splitting the optimization variables by their property (\ie~geometry, materials, poses) and alternate the optimization over those properties using multiple decoupled objectives.
In contrast, we exploit that spatially separated regions naturally decouple the corresponding optimization variables and therefore, we decompose the problem based on spatial regions instead of separate properties.
This has two advantages: 
1) It enables us to optimize all parameters of each region jointly and to use a single objective function.
Consequently, we can use all information encapsulated in the intricate interplay of geometry and materials and reach high accurate reconstructions.
2) The separation into spatial regions allows us to distribute the optimization over the local 2.5D representations of these regions and thus, facilitates scalability to larger scenes.

Following our conference paper \cite{Schmitt2020CVPR}, Luan \etal~\cite{Luan2021EGSR} proposed another method for jointly optimizing geometry and spatially-varying reflectance.
They represent the geometry of a single object as a mesh and alternate the optimization over mesh vertices and reflectance with re-meshing in a coarse-to-fine process.
Hereby, they use a co-located configuration of a hand-held camera and point light which greatly simplifies the rendering process but also restricts the sample space of the BRDF.
In contrast, we use multiple, alternating light sources in conjunction with explicit shadow modeling to extract most information from the sparse set of samples that is captured with a handheld setup.

Recently, \textbf{neural scene representations}~\cite{BI2020ECCV, Bi2020ARXIV, Cheng2021Arxiv} have been trained to reconstruct surface normals and reflectance properties of complex and multi-colored objects.
Most works are targeting object-centric scenarios with a fixed scanning volume.
\cite{BI2020ECCV} uses voxel representation 
 of deep features that encodes opacity, normals, and materials.
Instead of storing deep features discretely, \cite {Bi2020ARXIV} trains a neural network with positional encoding to represent continuous 3D functions of scene properties.
To track the 2D topology of the 3D surfaces of the object, \cite{Cheng2021Arxiv} optimize the neural transform from the unit sphere to a 3D object.
For a scene with multiple objects, all these methods either require to compromise reconstruction resolution due to a fixed encoding which limits scalability.
Or they are designed to re-train the network individually for each object in an object-centered unit-volume. 
This maximizes reconstruction quality per object but at the cost of a global scene representation.
In contrast, for our method the resolution does not depend on the scene size due to the local 2.5D representations and we optimize all local representations in a single global world coordinate system which does not require any reconfiguration of the representation after initialization.

With our proposed method, we make a step towards reconstructions of geometry, materials and poses beyond object-level.
For scalability, we optimize a set of 2.5D representations instead of a single 3D representation since each local 2.5D representation has constant requirements in terms of memory and optimization variables, independent of the scene size.
Furthermore, in contrast to methods that assume watertight 3D shapes, our model is able to reconstruct partially scanned 3D environments as it doesn't require closed shapes.

\subsection{Differentiable Rendering}

Differentiable rendering describes a rendering pipeline that allows for computing image pixel changes \wrt~scene parameters.
It is at the heart of most methods that aim to synthesize photo-realistic images from real-world observations. 
The approaches are manifold but many share a similar structure:
Based on real 2D or 2.5D observations, inverse rendering is used to infer a parametric representation of the 3D scene (\eg~for geometry, illumination or BRDF). 
This scene representation is then rendered into images by a forward rendering engine.
In the following, we discuss four method classes implementing this.

\textbf{Classical inverse rendering} approaches use non-learning-based methods to optimize 2D or 3D scene parameters from observation images via gradient descent.
\cite{Liu2019ICCV, Ravi2020ARXIV} use 'soft' rasterization to differentiate a rasterizer, while \cite{Loper2014ECCV, Loubet2019TOG, Li2018TOG, Niemeyer2020CVPR, Luan2021EGSR, NimierDavid2019TOG, Vicini2022TOG} propose solutions to differentiate through ray casting.
All these works use hand-designed rendering functions.
This limits the flexibility of the renderer (as compared to learning-based approaches, discussed afterwards) but has the advantage of physically-based rendering functions which are interpretable and enable rendering a scene under changed conditions (\eg~novel viewpoint, different illumination, or edited materials).
In the proposed pipeline, we use such a classical optimization approach with a hand-designed rendering engine since we aim for physically correct reconstructions.

\textbf{Neural inverse rendering} pipelines \cite{Li2020CVPRa, Li2018ECCVa, Deschaintre2018SIGGRAPH, Boss2020CVPR, Wimbauer2022CVPR, Li2022ECCV, Li2022CVPR, Zhu2022CVPR, Wang2021ICCV} train neural networks to predict scene parameters from observations and then use an analytical differentiable rendering layer to synthesize images.
The network has the potential to learn to ignore transient objects, adapt to varying illumination conditions in the observations or disentangle the parameters of complex scenes from data.
But it also introduces additional parameters and thus, requires data to train.
Especially when considering more complex reflectance settings, this is not easy to obtain.
In contrast, our approach does not require any large dataset but solely takes the captures of a scene as input.

\textbf{Neural rendering} (or 2D neural rendering) refers to methods that use classical surface or volume representations and replace the differentiable rendering engine by a generative model to learn the image formation function.
Exemplary tasks are changing the camera viewpoint \cite{Eslami2018Science, Meshry2019CVPR, Thies2020ICLR, Nguyen-Phuoc2018NIPS, Hedman2018TOG, Xu2019TOG, Sitzmann2019CVPR} or relighting \cite{Gao2020TOG, Xu2018TOG, Philip2019TOG}.
The generative network has the potential to synthesize high-quality novel images, learn visibility constraints, deal with incomplete or inconsistent input representations %
or depict complex illumination effects like inter-reflections or multiple bounces.
However, the rendering network is non-deterministic and how to enforce physical plausibility of the reconstructions is unclear -- which is the goal of this paper.
Therefore, the proposed approach relies on a classical rendering engine instead of a neural renderer.

Last, \textbf{neural scene representations} (or 3D neural rendering, learnable 3D representations) encode the scene parameters in a neural network and combine them with classical differentiable rendering engines.
A very well-known example is NeRF~\cite{Mildenhall2020ECCV} and its follow-ups. 
\Eg~\cite{Zhang2021CVPR, Shafiei2021BMVC, Boss2021ICCV} enable relighting and \cite{Zhang2021TOG, Srinivasan2021CVPR, Bi2020ARXIV, Boss2021NEURIPS, Boss2022NEURIPS, Kuang2021CORR, Munkberg2021ARXIV, Zhang2022CVPR, Yao2022ECCV, Zhang2021NEURIPS} include full BRDFs to encode material reflectance.
All these approaches are trained or fine-tuned per scene and require either prior knowledge on materials, like pre-trained reflectance or transmittance priors, or strong regularizers like compression to low dimensional latent spaces.
While our method is also optimized per scene, we do not assume any prior knowledge on materials.
Instead, we predict material parameters per pixel and use regularizers to propagate reflectance information across pixels.
Additionally, our 2.5D representation is faster to optimize than an MLP, naturally scales to large scenes and distributes modeling capacity equally across all selected keyframes.

\subsection{Geometry Estimation at Scale}

To date, there exists no solution to reconstruct accurate geometry and materials at scale.
In this paragraph, we therefore review existing work on scalable geometry-only reconstruction.
Crucial for a scalable model is a memory efficient scene representation that allows for accurate and dense reconstructions.

One approach is to keep the \textbf{full reconstruction in memory} by supporting efficient compression of connected surface data. 
\cite{Huang2017SIGGRAPH} represent the scanned environment by a light-weight mesh using plane primitives and \cite{Gallup2010JPRS} fits a multi-layer heightmap to a volume of occupancy votes.
Those representations support scene completion and can scale efficiently to larger scenes, but fail to reconstruct complex 3D structures.
In the context of Image Based Rendering (IBR), \cite{Hedman2016SIGGRAPH} calculate a global mesh from a pointcloud reconstruction but then refine per-view depth maps, sacrificing global consistency for local accuracy.

Another approach is to use \textbf{scene representations that allow for subdivision} into parts/segments or to introduce hierarchies to facilitate memory efficient processing by keeping only relevant scene parts in memory.
Hereby, a common strategy is to first reconstruct the geometry of individual and overlapping scene segments, parts or frames and then integrate those segments in 3D world space while employing sophisticated pose registration, alignment and outlier filtering techniques. 
Multiple \textbf{implicit volumetric models} extent the seminal works of \cite{Curless1996SIGGRAPH} or \cite{Newcombe2011ISMAR} (which rely on memory inefficient regular voxel grids) to larger environments by introducing efficient volumetric data structures like volume windows~\cite{Whelan2012RSSWORK}, patch volumes~\cite{Henry2013THREEDV}, a hierarchical volume structure~\cite{Chen2013SIGGRAPH} or spatial hashing~\cite{Niesner2013SIGGRAPH, Dai2017SIGGRAPH}.
This increases spatial efficiency.
Non-volumetric approaches represent scene segments by, \eg, per frame \textbf{2.5D depth maps} \cite{Pollefeys2008IJCV,Donne2019CVPR} or \textbf{3D mesh fragments} \cite{Choi2015CVPR}.
All aforementioned subdivision methods enable final reconstructions which would exceed memory limitations during processing.
And most employ global pose or texture refinements similar to \cite{Zhou2014SIGGRAPH, Bi2017SIGGRAPH, Huang2017SIGGRAPH}. 
The common challenge of non-global geometry reconstruction methods is to assure consistency between local reconstructions.

Our method represents the scene as a collection of 2.5D parameter maps from multiple keyframe views.
This representation is memory efficient and we actively encourage consistency between overlapping regions.
As demonstrated in \secref{sec:results}, this leads to accurate and well-aligned reconstructions, eliminating the need for post-processing or refinement.
\\

In the following 4 sections we describe our method in detail. 
First, we introduce our scene representation and parameterizations of the optimization variables in \secref{sec:scene_representation}.
We then present the model formulation and optimization objective in \secref{sec:optimization_objective} before discussing the multi-view consistent optimization scheme in \secref{sec:optimization}.
The mesh generation step that integrates the 2.5D optimization results into a full 3D model is explained in \secref{sec:postprocessing}.

\begin{figure}
	\begin{tabular}{cc}
		\centering 
		\includegraphics[width=0.45\linewidth]{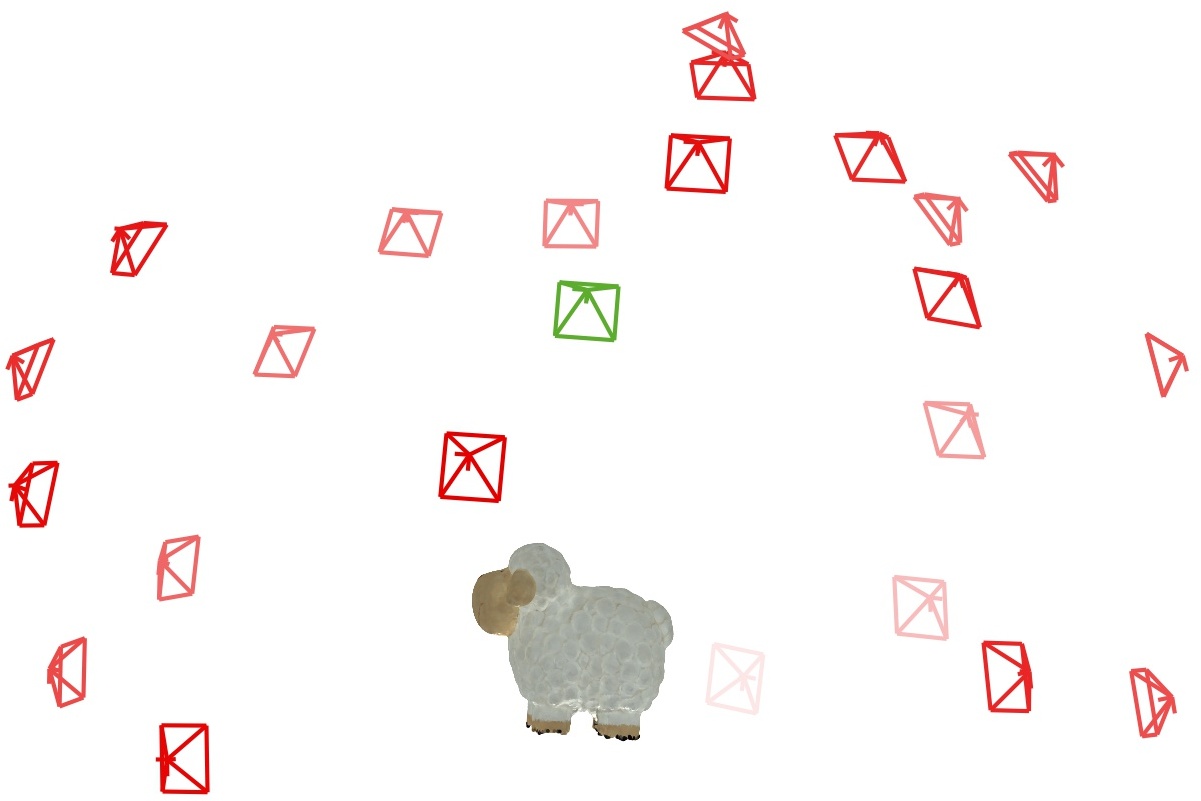} &
		\includegraphics[width=0.45\linewidth]{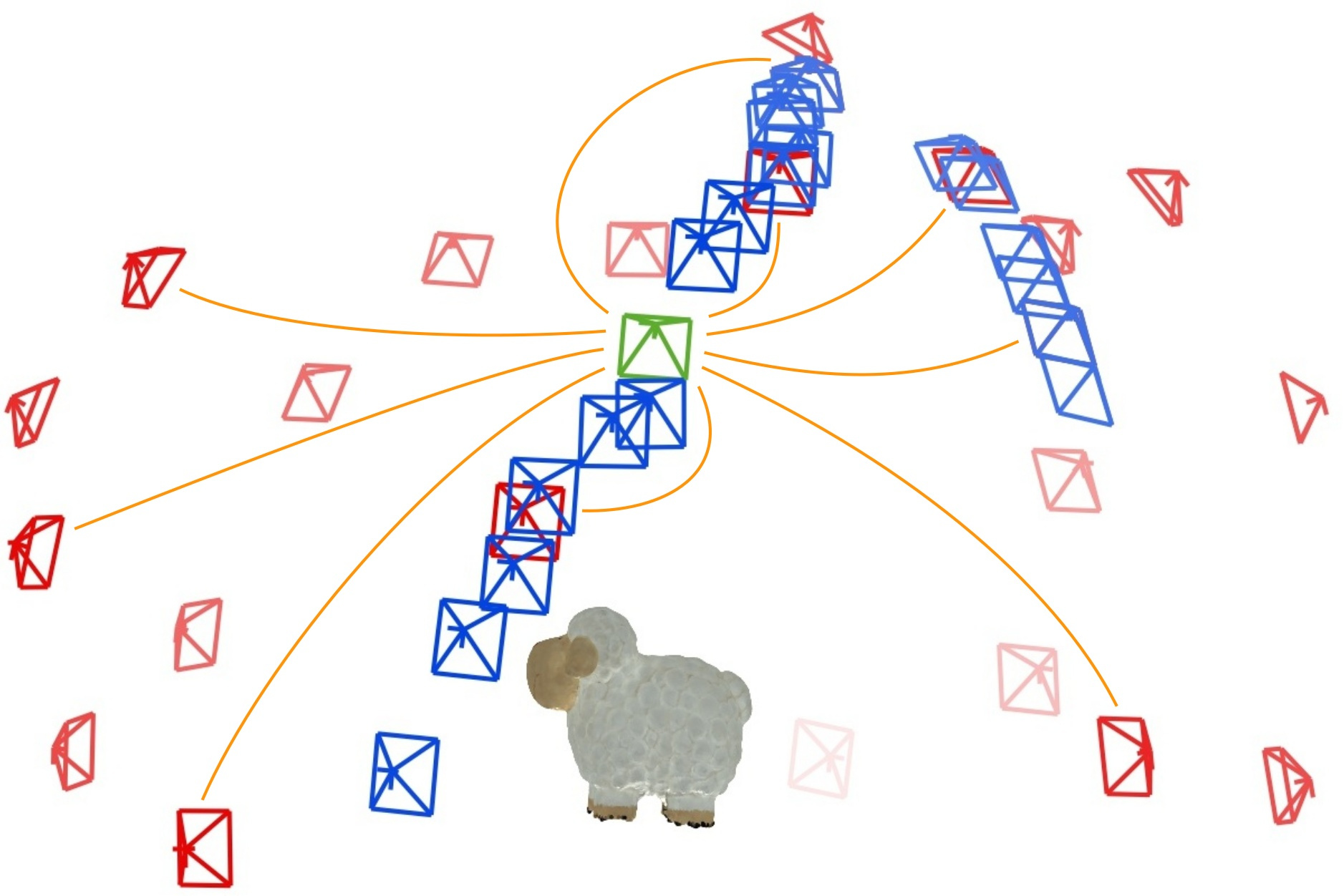} \\
		(a) Keyframes & (b) Neighbor Views
	\end{tabular}
	\caption{
		\textbf{Keyframe and Neighbor View Selection:} 
		(a) Visualized are all \textcolor{red}{keyframes $K$ (red, }\textcolor{green}{green)} for the object `Sheep' 
		- they represent the scene as a set of 2.5D maps for efficient optimization.
		(b) Our method optimizes the parameter set of one \textcolor{green}{keyframe $k \in K$ (green)} guided by photometric and geometric constraints from \textcolor{blue}{neighboring observation views $N_k$ (blue cameras)} and consistency constraints from \textcolor{orange}{neighboring keyframes $\bar{N}_k$ (orange lines)}.
	}
	\label{fig:representation_view_selection}
\end{figure}

\begin{figure*}
	\centering
	\includegraphics[width=0.75\linewidth]{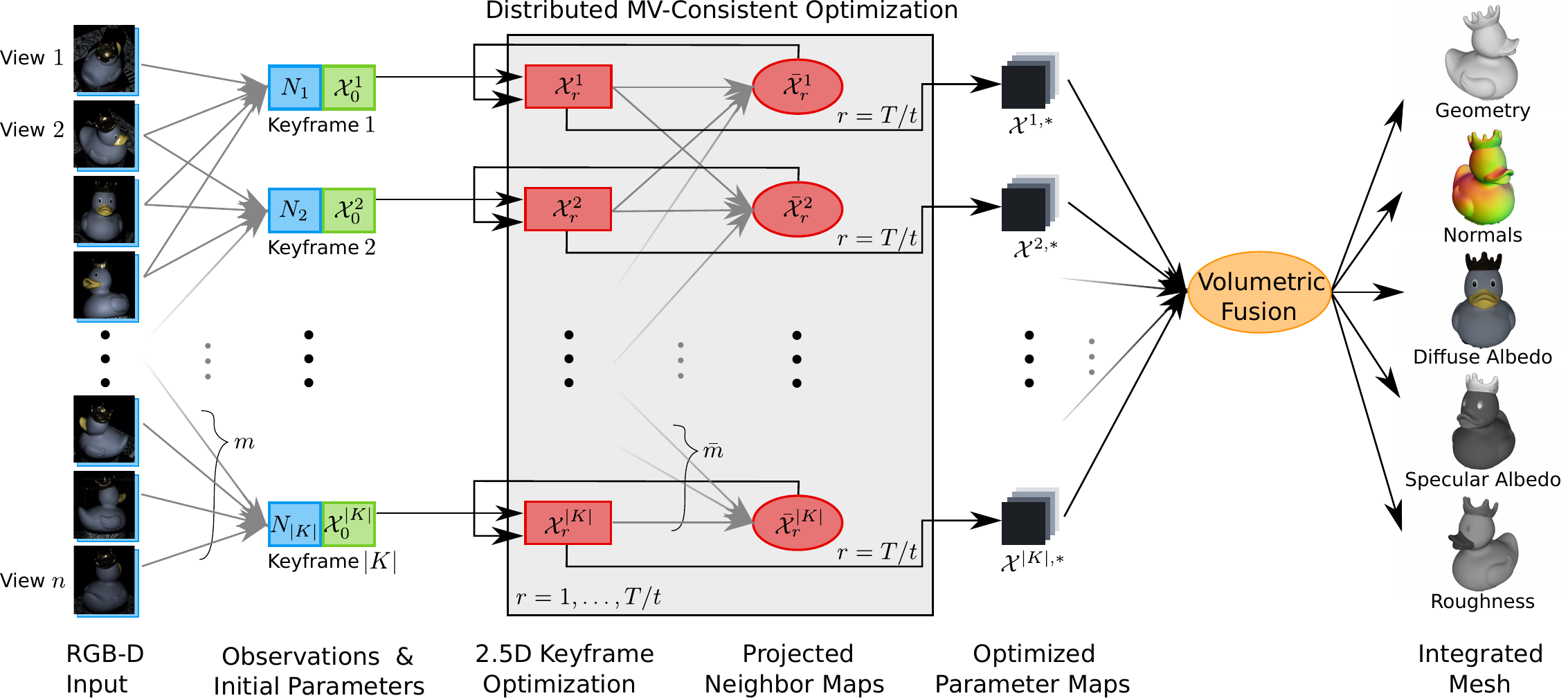}
	\vspace{2mm}
	\caption{
		\textbf{Pipeline Overview.}
		The input to our model are $n$ RGB-D images from which we select a subset of well distributed keyframes $K$. 
		For each keyframe $k\in K$, we select $m$ neighboring observation views $N_k$ and initialize the set of optimization parameters $\cX_0^k$.
		During optimization, we then iterate the following rounds $r$ (gray box):
		After optimizing each keyframe representation independently for $t$ iterations, we project the current parameter maps of all neighboring keyframes $\bar{N_k}$ into each keyframe $k$ and use the resulting set $\bar{\cX_r^k}$ as additional constraint to the optimization of the next round $r+1$.
		The sets of $\bar{m}$ neighboring keyframes $\{\bar{N_k}\}_k$ are defined at the start for all keyframes $k$.
		After $r=T/t$ rounds, the resulting sets of optimized 2.5D parameter maps $\{\cX^{k, *}\}_k$ are integrated into a full 3D model, represented by a mesh with per vertex normal, diffuse and specular albedo as well as roughness parameters.
	}
	\label{fig:pipeline_overview}
\end{figure*}

\section{Scene Representation}
\label{sec:scene_representation}

Our goal is to reconstruct geometry, material properties and camera poses from RGB-D data.
Unfortunately, representing an entire scene in memory is computationally demanding, in particular when using memory-limited but computationally efficient GPUs for optimization.
Towards scalable scene reconstruction, we therefore exploit a keyframe-based 2.5D representation which locally describes and optimizes geometry, materials and poses.
In particular, we adopt alternating block coordinate optimization of keyframes to minimize photometric errors while encouraging consistency between adjacent keyframes using soft constraints.
An overview of our method is shown in \figref{fig:pipeline_overview}.

The input to our model is an RGB-D sequence captured with a handheld scanner, as shown in \figref{fig:results_sensor}, that consists of a color image $\cI_i: \nR^2\to \nR^3$ and a depth map $\cZ_i: \nR^2\to \nR$ at each frame $i\in N=\{1,\dots,n\}$. We assume that each image is illuminated by exactly one point light source and that global and ambient illumination effects are negligible. Moreover, we assume the images to be undistorted, de-vignetted and the black frame to be subtracted. 

We represent the scene as a set of 2D parameter maps defined at several keyframes of the RGB-D sequence. 
More specifically, at each keyframe we store the geometry in terms of a depth and normal map, and the materials as BRDF parameter maps.
Additionally, each keyframe is linked to a set of camera poses of its respective neighbor views.
In the following, we first describe the process of keyframe and neighbor view selection, followed by the representations for poses, geometry and materials.

\subsection{Keyframe and Neighbor Selection}

To represent the scene, we define a set of keyframes $K \subseteq N$ that capture the scene tightly.
For each keyframe $k\in K$ we define two sets of neighboring views:
The first set is the set of neighboring observation views $N_k$ which provide photometric and geometric constraints for the local 2.5D multi-view optimizations over the parameter set of keyframe $k$.
Second, we define a set of neighboring keyframes $\bar{N}_k$ from which we project the parameter maps into keyframe $k$ as a soft constraint during optimization to enforce consistency of the local 2.5D reconstructions.
And since all keyframes are connected via the overall optimization graph, these pairwise consistency constraints propagate globally during optimization.
As evidenced by our experiments, this term is crucial for obtaining a consistent result when fusing all 2.5D representations into a global 3D representation of the scene.
All sets of views are visualized in \figref{fig:representation_view_selection} for the capture of the object `Sheep'.

\boldparagraph{Keyframe Selection}
To select a set of diverse keyframes $K \subseteq N$, we iteratively compute the pairwise 3D Euclidean distances between the camera centers of all views and remove the view with the minimum distance to its nearest neighbor until the desired number of keyframes has been reached.
Generally, the number of keyframes is a tradeoff between accuracy and time and it grows with the scale of the scene.
But increasing the number of keyframes is unproblematic for our method since most computations run per keyframe in parallel, with fixed memory requirements independent of the scene size.
	We ablate the number of keyframes in \secref{sec:results_ablation}.

\boldparagraph{Neighboring Observation Views $N_k$}
To optimize geometry, pose and spatially-varying material parameters, the set of keyframe observations contains too little samples.	Therefore, we define $m$ neighboring observation views $N_k \subset N$ with $m=|N_k|$ per keyframe $k\in K$ and minimize the photoconsistency error between these and the predictions of our model.
To select the neighbor observation views we only consider views that are within a $40^\circ$ cone around keyframe $k$ with respect to (\wrt) the object center. 
For larger scenes, we additionally remove views with a view direction that deviates more than $45^\circ$ from the keyframes' view direction.
We then choose $m$ views that cover the cone around keyframe $k$ as uniformly as possible by removing the views which are closest to their neighbors.

\boldparagraph{Neighboring Keyframes $\bar{N}_k$}
For each keyframe $k \in K$, we define a set of neighbor keyframes $\bar{N}_k\subseteq K\setminus\{k\}$ of size $\bar{m} = \vert\bar{N}_k \vert$.
During optimization, we regularize the parameter maps of keyframe $k$ against those of all neighbor keyframes $i\in \bar{N}_k$ projected into $k$.
This enforces consistent parameter estimates across keyframes. 
To ensure that all neighbors $i\in\bar{N}_k$ share scene content with keyframe $k$, we sample them randomly from all keyframes 
that fulfill two conditions: For keyframes $i$ and $k$, 1) define the \textit{middle point} as the median of all initial geometry points for objects and the first intersection point of the principal ray of camera $k$ with the initial geometry for scenes.
Then the two lines connecting each views' camera position with the \textit{middle point} should form an angle of $\leq 60^\circ$. 
And 2) for scenes, both cameras' view directions form an angle of $\leq 45^\circ$.
Note that per keyframe, we sample up to $\bar{m}$ neighboring keyframes, depending on the availability of valid neighbors.

\subsection{Keyframe Parameterizations}

In this section, we formally describe the keyframe-based parameterization of our model in terms of poses, geometry and materials.
For each keyframe $k\in K$, we define its pixels $P_k$ as the set of all pixels of view $k$ with a non-zero initial depth value. As we bound the depth to be non-negative, this implies $z_p^k > 0$.

\subsubsection{Camera Parameterization}

We use a perspective pinhole camera model and assume constant intrinsic camera parameters that have been calibrated in advance using established calibration procedures \cite{Zhang1999ICCV}.
We denote the projective mapping for observation $i\in N_k$ and keyframe $k\in K$ as: $\pi_i^k : \nR^3 \to \nR^2$ and represent the extrinsic component (camera pose) of this mapping in world coordinates by a unit quaternion $\bq_i^k \in SO(3)$ and a translation vector $\bt_i^k \in \nR^3$.
Note that we use a redundant representation (\ie, the camera pose of an observation neighboring multiple keyframes is represented once per keyframe) to enable memory efficient optimization, one keyframe at a time, while enforcing consistency via additional soft constraints.

\subsubsection{Geometry Parameterization}
We parameterize geometry in terms of both depth and normal maps and enforce consistency between them using soft constraints.

\boldparagraph{Depth Map}
For each pixel $p\in P_k$ of keyframe $k$ at 2D location $\big(u_p^k, v_p^k\big)^T$ and associated depth $z_p^k$, the 3D point location $\bx_p^k$ is given by
\begin{equation}
	\bx_p^k = {\big(\pi_{k}^{k}\big)}^{-1}\big(u_p^k, v_p^k, z_p^k\big) 
\end{equation}
where $ {\big(\pi_{k}^{k}\big)}^{-1}$ denotes the inverse projection which takes a pixel coordinate and depth value and returns the 3D point in world coordinates.

\boldparagraph{Normal Map}
We represent normals as 3D vectors $\{\bn_p^k\}_{p\in P_k}$. 
During optimization, we only estimate an angular change \wrt~the normal of the previous iteration to avoid both the unit vector constraint and the gimbal lock problem.

\subsubsection{Material Parameterization}
To model reflectance properties, we use a parametric version of the spatially varying Bidirectional Reflectance Distribution Function $f_{p}(\bn_p, \bomega_\text{in}, \bomega_\text{out})$ and estimate its parameters per pixel/point $p\in P_k$ and keyframe $k$.

\boldparagraph{svBRDF}
The svBRDF $f_{p}(\cdot)$ models the fraction of light that is reflected from incoming light direction $\bomega_\text{in}$ to outgoing light direction $\bomega_\text{out}$ given the surface normal $\bn_p^k$ at each point $p\in P_k$.
We use a modified version of the Cook-Torrance model \cite{Cook1982SIGGRAPH}
\begin{align}
	f_p^k(\bn_p^k, \bomega_\text{in}, \bomega_\text{out}) = \bd_p^k + s_p^k \, \frac{D(r_p^k) \; G(\bn_p^k, \bomega_\text{in}, \bomega_\text{out}, r_p^k)}{4 (\bn_p^k\cdot \bomega_\text{in}) (\bn_p^k\cdot\bomega_\text{out})}
	\label{eq:microfacet_brdf}
\end{align}
with Disney's GTR model \cite{Burley2012} for the microfacet slope distribution $D(\cdot)$ and Mitsuba's Smith's function \cite{Jakob2010} for the geometric attenuation factor $G(\cdot)$.
The parameters of the svBRDF are given by the diffuse albedo $\bd_p^k\in\nR^3$, specular albedo $s_p^k\in\nR$ and surface roughness $r_p^k\in\nR$ for pixel/point $p\in P_k$ and keyframe $k$.
As in prior work \cite{Nam2018SIGGRAPH}, we ignore the Fresnel effect which cannot be observed using an active handheld illumination setup.

\section{Optimization Objective}
\label{sec:optimization_objective}

To jointly optimize geometry, materials and pose parameters for each keyframe $k \in K$, we minimize the photometric error between rendered predictions and neighbor view observations while employing multiple additional loss functions for regularization.\\

For a single keyframe $k$ and itspixels/points $p\in P_k$ and neighbor observation views $i\in N_k$, we wish to estimate the depth $z_p^k$, geometric surface normals $\bn_p^k$, svBRDF parameters $\bd_p^k, r_p^k, s_p^k$ as well as the camera poses $\pi_i^k$.
Denoting the parameter set as 
\[\cX=\{\{z_p^k, \bn_p^k, \bd_p^k, r_p^k, s_p^k\}_{p\in P_k}, \{\pi_i^k\}_{i\in N_k}\}_{k\in K}\]
we define our objective function as follows
\begin{equation}
	\cX^*=\argmin_{\cX} \psi_\cP + \psi_\cD + \psi_\cC + \psi_\cG + \psi_\cM
	\label{eq:objective}
\end{equation}
The individual terms encourage photo-consistency $\psi_\cP$, depth-consistency $\psi_\cD$ and multi-view consistency $\psi_\cC$, impose regularization on the geometry $\psi_\cG$ and enforce material smoothness $\psi_\cM$.
Note that we omit the dependency on $\cX$ and the relative weights between the individual terms for clarity.
The full formulation can be found in the supplement.

\subsection{Photo and Depth Consistency}
We introduce the photo and depth consistency terms in the following.
For better readability, we denote $\cI_i(\pi_i^k(\bx_p^k))$ as the observation $\cI_i$ at the 2D location where 3D point $\bx_p^k$ is observed in image $i$.
For fractional image coordinates, we use bilinear interpolation.
Similarly, we write $\cZ_i(\pi_i^k(\bx_p^k))$ for depth measurements.

\boldparagraph{Photo Consistency}
The photo-consistency term ensures that the prediction of our model matches the observation $\cI_i$ for every neighbor view $i\in N_k$ and all visible and illuminated  ($\varphi_p^{ki} = 1$) pixels $p$:
\begin{align}
	\nonumber &\psi_\cP(\cX) =\label{eq:photoconsistency_term}\\
	&\sum_{i\in N_k} \sum_{p\in P_k} {\left\Vert\,\varphi_p^{ki} \, w_p^i \, \left[\cI_i(\pi_i^k(\bx_p^k)) - \cR_i(\bx_p^k, \bn_p^k, f_p^k)\right]\,\right\Vert}_1
\end{align}
Here, $\cR_i$ denotes the rendering equation \cite{Kajiya1986SIGGRAPH} for image $i$. 
Since we assume a single point light source, $\cR_i$ simplifies to
\begin{align}
	\nonumber \cR_i&(\bx_p^k, \bn_p^k, f_p^k) = \\
	&f_p^k\left(\bn_p^k, \bomega_\text{in}^{i}(\bx_p^k), \bomega_\text{out}^{k}(\bx_p^k)\right)\, \frac{a_{i}(\bx_p^k)\,\bn_p^{k^T} \bomega_\text{in}^{i}(\bx_p^k)}{{d_{i}(\bx_p^k)}^2}~L
\end{align}
where $\bomega_\text{in}^{i}(\bx_p^k)$ and $\bomega_\text{out}^{k}(\bx_p^k)$ denote the in- and out-going light directions for the surface point $\bx_p^k$, $a_{i}(\bx_p^k)$ is the angle-dependent light attenuation, $d_{i}(\bx_p^k)$ the distance between $\bx_p^k$ and the light source and $L$ denotes the radiant intensity of the light.
To downweight observations at grazing angles, we use a weight $w_p^i \propto \bn_p^i \bl_p^i$ proportional to the angle between the surface normal $\bn_p^i$ and light direction $\bl_p^i$.

We calculate the visibility term $\varphi_p^{ki}\in\{0,1\}$ of surface point $\bx_p^k$ in observation view $i$ by reconstructing a rough 3D model of the scene using volumetric fusion of the depth observations and performing a zbuffer test to validate if 3D point $\bx_p^k$ is both visible in view $i$ and illuminated by the (calibrated) light source corresponding to keyframe $k$.

\boldparagraph{Depth Consistency}
We further constrain the depth estimate $\{z_p^k\}_p$ against the depth measurements $\cZ_i$ of all neighboring views $i\in N_k$:
\begin{equation}
\psi_\cD(\cX) = \sum_{i\in N_k} \sum_{p\in P_k} \; \varphi_p^{ki} \; {\left\Vert \, z_p^i- \cZ_i(\pi_i^k(\bx_p^k))\right\Vert}_{2}^{2}
\end{equation}
Here, $z_p^i$ denotes the depth of the 3D point $\bx_p^k$ of keyframe $k$ when projected to the neighbor view $i$ via $\pi_i^k(\bx_p^k)$. 
As above, $\varphi_p^{ki}$ ensures that surface point $\bx_p^k$ is visible in image $i$.

Note that our model is able to significantly improve upon the initial coarse geometry provided by the structured light sensor by exploiting shading cues. However, as these cues are related to depth variations (\ie, normals) rather than absolute depth, they do not fully constrain the 3D shape of the object.
Our experiments demonstrate that combining complementary depth and shading cues yields reconstructions which are both locally detailed and globally consistent.

\subsection{Multi-View Consistency}
\label{sec:optimization_mv_consistency}
Since our representation is composed of multiple 2.5D views, we must ensure consistency between them. 
Towards this goal, we augment our objective with a multi-view consistency term which encourages the current parameter estimates $\{z_p^k, \bd_p^k, r_p^k, s_p^k\}_{p\in P_k}$, $\{\bt_j^k, \bq_j^k\}_{j\in N_k}$ of keyframe $k$ to agree with those of neighboring keyframes $\{\bar{z}_p^k, \bar{\bd}_p^k, \bar{r}_p^k, \bar{s}_p^k\}_{p\in P_k}, \{\bar{\bt}^i, \bar{\bq}^i\}_{i\in \bar{N}_k}$ projected into the current keyframe:
\begin{align}
\psi_\cC(\cX) = \frac{1}{\vert P_k\vert} \sum_{p\in P_k}& {\big\Vert\bx_p^k - \bar{\bx}_p^k \big\Vert}_2 + {\big\Vert\bd_p^k - \bar{\bd}_p^k \big\Vert}_1 \nonumber\\
&+ {\big\vert r_p^k - \bar{r}_p^k \big\vert} + {\big\vert s_p^k - \bar{s}_p^k \big\vert} \nonumber\\
+ \frac{1}{n} \sum_{i\in \bar{N}_k} \sum_{j \in N_i\cap N_k}& {\big\Vert\bt_j^k - \bt_j^i \big\Vert}_1 + {\left\Vert\left( \left(\bq_j^i\right)^{-1} \otimes \bq_j^k \right)_v\right\Vert}_1 
\label{eq:multi_view_consistency}
\end{align}
Hereby, the projected neighbor parameters $\{\bar{z}_p^k, \bar{\bd}_p^k, \bar{r}_p^k, \bar{s}_p^k\}_{p\in P_k}$ for each surface point $\bx_p^k$ are computed as follows
\begin{equation} \label{eq: neighbor_parameter_map}
	(\bar{\cdot})_p^k = \frac{1}{\sum_i \varphi_p^{ki} \; w_p^i} \; \sum_{i \in \bar{N}_k} \varphi_p^{ki} \; w_p^i \; \text{interp}\left(\{(\cdot)_q^i\}_q, \pi_i(\bx_p^k)\right)
\end{equation}
where $\bar{N}_k$ is the set of neighboring keyframes, $\varphi_p^{ki}$ denotes visibility as defined above, and $w_p^i = \bn_p^i \bl_p^i$ downweights estimates at grazing angles.
The mapping $\text{interp}: \nR^{\vert P\vert} \times \nR^2 \to\nR$ takes a neighboring parameter map $\{(\cdot)_q^i\}_{q\in P_i}$ and a projected 2D pixel location $\pi_i(\bx_p^k)$ and outputs the bilinearly interpolated parameter value.
The poses of all neighboring keyframes $\{\bar{\bt}^i, \bar{\bq}^i\}_{i\in \bar{N}_k}$ are defined as
\begin{equation}
	\bar{\bt}^i = \left\{ \bt_j^i \,\,\vert \; j \in N_i\cap N_k \right\} 
	\;\;\text{and}\;\; \bar{\bq}^i = \left\{\bq_j^i \,\,\vert \; j \in N_i\cap N_k \right\}
\end{equation}
The $\otimes$ operator in the last term of \eqref{eq:multi_view_consistency} denotes the Hamiltonian product for quaternions and calculates the composed rotation. Furthermore, $(\cdot)_v$ denotes the vector part of the quaternion which equals zero for the identity.
Note that we only calculate the pose loss term for cameras which are part of the observations for both the current keyframe $k$ and the neighboring keyframes $i$, \ie, $j \in N_i\cap N_k$.

\subsection{Geometry Regularization}

Our geometry regularizers encourage geometric consistency $\psi_{\cG\cC}$ and normal smoothness $\psi_\cN$ as follows:
\begin{equation}
	\psi_\cG = \psi_{\cG\cC} + \psi_\cN
	\label{eq:objective_geometric_reg}
\end{equation}
As above, we omitted the dependency on $\cX$ and relative weights of the individual terms for clarity.

\boldparagraph{Geometric Consistency}
We enforce consistency between depth $\{z_p^k\}$ and normals $\{\bn_p^k\}$ by maximizing the inner product between the estimated normals $\{\bn_p^k\}$ and the cross product of the surface tangents at $\{\bx_p^k\}$:
\begin{equation}
	\psi_{\cG\cC}(\cX) = - \sum_{p} \left(\bn_p^k\right)^T~\frac{\frac{\partial z_p^k}{\partial x}\times\frac{\partial z_p^k}{\partial y}}{{\left\Vert\frac{\partial z_p^k}{\partial x}\times\frac{\partial z_p^k}{\partial y}\right\Vert}_2}
	\label{eq:consistency_term}
\end{equation}
where the surface tangent $\frac{\partial z_p^k}{\partial x}$ is given by
\begin{equation}
	\frac{\partial z_p^k}{\partial x} \propto \left[1,0, \nabla\cZ_{k}(\pi_{k}(\bx_p^k))^T \left[f/z_p, 0\right]^T\right]^T
\end{equation}
with $\nabla\cZ_{k}(\pi_{k}(\bx_p^k))$ the gradient of the depth map estimated using finite differences.
We refer to \cite{Schmitt2020CVPR} for details.

\boldparagraph{Normal Smoothness}
We further encourage normals of adjacent pixels $p\sim q$ to be similar:
\begin{equation}
	\psi_\cN(\cX) = \sum_{p\sim q} e_{pq}^k \; {\big\Vert{\bn_p^k-\bn_q^k\big\Vert}}_{1}
\end{equation}
Here, $e_{pq}^k$ is an edge-aware weighting term based on a Canny filter~\cite{Canny1986PAMI} to reduce the smoothing at pixels close to edges in the albedo map and facilitate detailed geometry reconstruction.

\subsection{Material Smoothness}
\label{sec:material_smoothness}
To enforce propagation of reflectance parameters across pixels, we constrain the specular albedo $\{s_p^k\}_p$ and roughness $\{r_p^k\}_p$ maps against a bilaterally smoothed version of themselves:
\begin{align}
	\psi_\cM(\cX) = \sum_p &{\left\Vert s_p^k - \frac{\sum_q s_q^k \, w_q^k g_{pq}^k}{\sum_q w_q^k g_{pq}^k }\right\Vert}_1 \nonumber\\
	+ &{\left\Vert r_p^k - \frac{\sum_q r_q^k \, w_q^k g_{pq}^k}{\sum_q w_q^k g_{pq}^k }\right\Vert}_1
	\label{eq:material_regularizer}
\end{align}
Assuming that nearby pixels with similar diffuse behavior also exhibit similar specular behavior, we use a Gaussian kernel $g_{pq}^k$ with both the 3D location $\bx$ and diffuse albedo $\bd$ at pixels $p$ and $q$ as features:
\begin{equation}
	g_{pq}^k = \exp \left( -\frac{(\bx^k_p-\bx^k_q)_2^2}{2\sigma^2_{1}} -\frac{(\bd^k_p-\bd^k_q)_2^2}{2\sigma^2_{2}} \right)
\end{equation}
The weight $w_q^k=\max_{i}\textrm{cos}^{-1}(\bn_q^{k^T} \bh^{ki}_q)$ with half-vector $\bh^{ki}_q$
increases the contribution of pixels $q$ which are observed close to perfect mirror reflection in any view $i$ and are therefore most informative for specular material estimation.
We use the permutohedral lattice~\cite{Adams2010CGF} for efficient evaluation of \eqnref{eq:material_regularizer}.

\SetAlCapSkip{1em}
\begin{algorithm}[t]
	\DontPrintSemicolon
	\SetKwFunction{range}{range}
	\KwData{Color and depth images $\{ \cI_i, \cZ_i\}_{i\in N}$.}
	\KwResult{Mesh $\cM$ featuring per-vertex normals and BRDF parameters.}
	\BlankLine
	Initialize $\forall k\in K$: \tcp*[f]{\secref{sec:method_optimization_initialization}}
	\indent $\quad \cX_0^k = \{z_p^k, \bn_p^k, \bd_p^k, r_p^k, s_p^k\}_{p\in P_k}$ and $\{\pi_i^k\}_{i\in N_k}$ \;
	\indent $\quad\;\: \bar{\cX}_{0}^k = \textit{None}$\;
	\BlankLine
	$t=100, \; T=2000$ \;
	\textit{rounds} $ = T / t$ \;
	\BlankLine
	
	Multi-View Consistent Optimization: \;
	\For{$r=1$ \KwTo \textit{rounds}}{
		\For{\textit{keyframe} $k \in K$}{
			\BlankLine
			Optimize $\cX_{r-1}^k$ given $\bar{\cX}_{r-1}^k$ for $t$ iterations:\;
			$\cX_r^k = \cX_{r-1}^{k,*}$ \tcp*[r]{\eqnref{eq:objective}} 
			\BlankLine
			Project neighbor parameter maps: \tcp*[r]{\eqnref{eq: neighbor_parameter_map}}
			$\bar{\cX}_r^k = \{\bar{z}_p^k, \bar{\bd}_p^k, \bar{r}_p^k, \bar{s}_p^k\}_{p\in P_k}$ 
			and $\{\bar{\pi}_i^k\}_{i\in \bar{N}_k}$
			\BlankLine
		}
	}
	\BlankLine
	Mesh Generation: \tcp*[r]{\secref{sec:postprocessing}}
	Fuse all final keyframe parameter maps $\{\cX_{r=T/t}^k\}_{k\in K}$\;
	into a mesh $\cM$ by volumetric fusion and marching cubes.
	\caption{Pseudo-Code of the proposed Algorithm.}
	\label{alg:optimization_pseudo_code}
\end{algorithm}

\section{Optimization}
\label{sec:optimization}

Direct optimization of the global objective in \eqnref{eq:objective} does not scale to larger scenes due to the large amount of data (variables and observations) that need to be stored and GPU memory limitations.
Instead, we decompose the global reconstruction into multiple keyframe reconstructions and perform decentralized, frame-wise block coordinate descent in parallel on multiple processes.
With this distributed optimization strategy, we drastically reduce the memory footprint since we only ever need to store one block in memory at a time per process.
To keep locally adjacent blocks consistent, we periodically share the state of optimization variables between neighboring keyframes and regularize differences in the reconstructed models.
An overview of the full optimization algorithm is given in \algref{alg:optimization_pseudo_code}.

In the following, we first motivate our decentralized optimization strategy and then elaborate on the block coordinate descent algorithm.
Subsequently, we provide details about the parameter initialization and our implementation.

\subsection{Decentralized Optimization}
\label{sec:method_optimiziation_decentralized}

For large computational problems, a \textbf{distributed optimization strategy} that allows for multiple processes and parallelization is essential.
As per \cite{Assran2020Arxiv}, distributed methods can be categorized into centralized and decentralized algorithms, depending on whether the processes read and update a central copy of the optimization variables or work on independent local copies.
Centralized algorithms require consistent transaction management such as semaphores, cause additional computational cost for centralization and distribution, and exhibit less stable optimization behavior due to potentially contradicting updates to the central optimization variables by different processes.
Therefore we implement a decentralized algorithm that facilitates accurate local reconstructions.
As this might result in multiple different estimates per variable, we introduce a soft regularizer to establish synchronization between processes and encourage consistency across spatially nearby regions.
We reduce the communication overhead by employing a strategy similar to \cite{Wang2020ICLR} and letting each process independently perform a set of base optimization steps in between synchronization.
In the following, we call this set of optimization steps performed for all processes a ``round''.

\subsection{Block Coordinate Descent Optimization}
\label{sec:method_optimiziation_block_coordinate_descent}

For optimization, we represent the target scene as a set of 2.5D parameter maps induced by the keyframes.
The keyframes can be viewed as blocks of variables over which we iterate as described in the following:

We start the first round by optimizing over every block/keyframe for $t$ iterations independently using gradient descent. 
Given these intermediate optimization states $\{\cX_1^k\}_k$ of all keyframes in $K$, the current parameter map estimates are projected into neighboring keyframes. 
We refer to them as $\{\bar{\cX}_1^k\}_k$.
They are used in the subsequent round when the parameters of all blocks/keyframes are re-optimized for $t$ iterations with the additional multi-view consistency regularizer. 
We iterate these rounds of optimizing for $\{\cX_r^k\}_k$ and calculating $\{\bar{\cX}_r^k\}_k$ for a total of $T/t$ rounds and $T$ iterations.
During the first half of optimization, we use a higher geometric smoothing regularizer to help bootstrapping the parameter maps. 
Thereafter, the smoothness regularization is reduced to allow for carving out fine geometric details and modeling sharp specular highlights during the remaining iterations.

\subsection{Initialization}
\label{sec:method_optimization_initialization}
We initialize the \textbf{poses} using the SfM pipeline COLMAP~\cite{Schoenberger2016CVPR, Schoenberger2016ECCV}.
For the \textbf{depth} initialization we pre-integrate the rather coarse data of an active stereo setup into a fused 3D geometry using volumetric fusion~\cite{Zeng2017CVPR} and render an initial depth map per keyframe $k \in K$, see \figref{fig:results_geometry} for an example of the initial geometry. 
Initial \textbf{diffuse albedo} and \textbf{normal} maps can be computed in closed form assuming a Lambertian scene.
Towards this goal, we follow \cite{Higo2009ICCV} and robustly filter outliers due to specularities using RANSAC.
Both \textbf{specular albedo} and \textbf{roughness} parameter maps are initialized by sampling randomly and uniformly from the intervals $\left[0.05, 0.25\right]$ and $\left[0.1, 0.9\right]$, respectively.

\subsection{Implementation details}
\label{sec:method_optimization_implementation}
We have implemented the rendering function $\cR_i$ using PyTorch \cite{Paszke2017NIPSWORK}, exploiting PyTorch's GPU acceleration and auto-differentiation capabilities.
For our experiments we use $\vert K\vert = 24$ keyframes, $m=20$ neighboring observation views and $\bar{m}=10$ neighboring keyframes. 
We project parameter maps into neighboring keyframes every $t=100$ iterations and optimize for $T=2000$ iterations in total.
Please see the supplement for more details about the optimization.

\section{Mesh Generation}
\label{sec:postprocessing}

In order to obtain a full 3D reconstruction of geometry and materials we use a memory efficient, voxel hashing based implementation of volumetric fusion \cite{Curless1996SIGGRAPH} as seen in \cite{Niesner2013SIGGRAPH}. 
Since predictions are most accurate for pixels observed frontally, we weight each pixel contribution by the cosine between surface normal and viewing direction.
We extract the final mesh from the TSDF via marching cubes \cite{Lorensen1987SIGGRAPH}.
As we impose consistency across keyframes during optimization, the fused parameter maps are consistent without need for extra post-processing/alignment steps.
Finally, the resulting mesh allows for extracting a texture map for each svBRDF parameter using Blender and exporting the mesh into the OBJ file format.
During all our experiments, we use a voxel size of 0.5 - 2mm.

	\begin{figure}
	\centering
	\begin{subfigure}{\linewidth}
		\centering
		\begin{tabular}{ccc}
	\toprule
	 & Fixed Poses~ &  Full Model~ \\
	\midrule
	Photometric Test Error   & 1.210 & 1.138 \\
	\bottomrule
\end{tabular}

		\vspace{0.1cm}
	\end{subfigure}
	\begin{subfigure}{\linewidth}
		\centering
		\setlength{\tabcolsep}{1pt}
		\begin{tabular}{ccccc}
			\includegraphics[height=2.5cm]{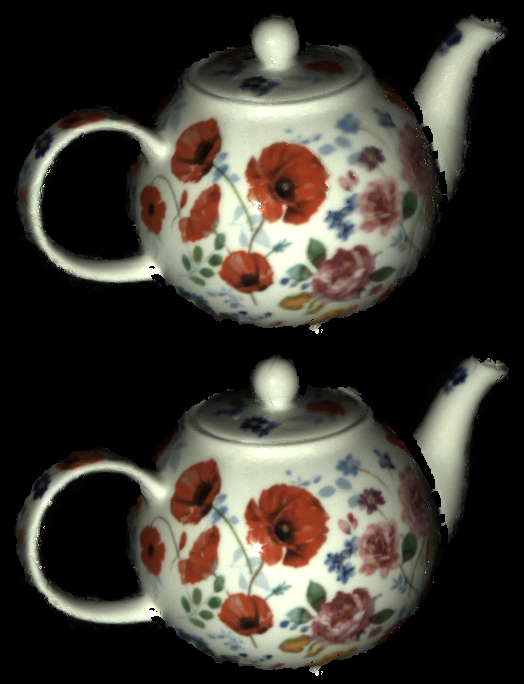}
			&\includegraphics[height=2.5cm]{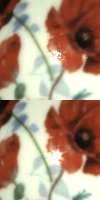}
			&\includegraphics[height=2.5cm]{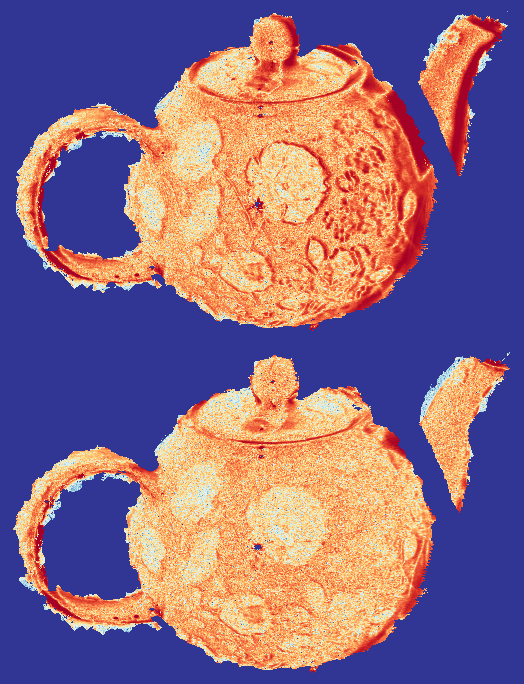}
			&\includegraphics[height=2.5cm]{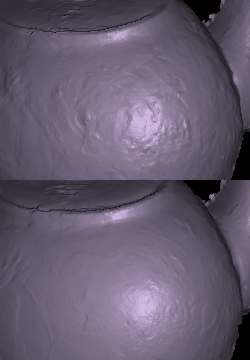}
			&\includegraphics[height=2.5cm]{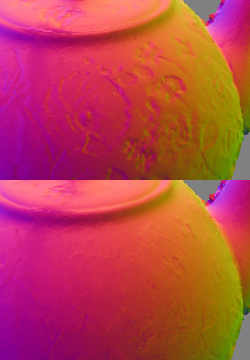}
			\\
			Estimation & Crop & Error & Structure & Normals
		\end{tabular}
	\end{subfigure}
	\caption{
		\textbf{Pose Optimization in 2.5D,} results from Schmitt \etal~\cite{Schmitt2020CVPR}. 
		Compared to using the input poses (top), optimizing the poses (bottom) improves reconstruction quality significantly.
	}
	\label{fig:poses_2.5d}
\end{figure}

\begin{figure}
	\centering
	\begin{subfigure}{0.55\linewidth}
		\begin{minipage}{\linewidth}
			\centering
			\includegraphics[height=4.5cm]{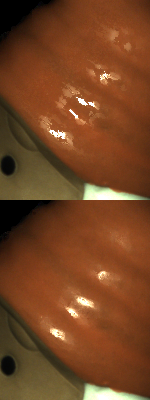}
			\includegraphics[height=4.5cm]{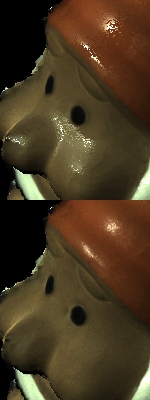}	\\
			Material Regularizer
		\end{minipage}
	\end{subfigure}
	\begin{subfigure}{0.43\linewidth}
		\begin{minipage}{\linewidth}
			\centering
			\includegraphics[height=4.5cm]{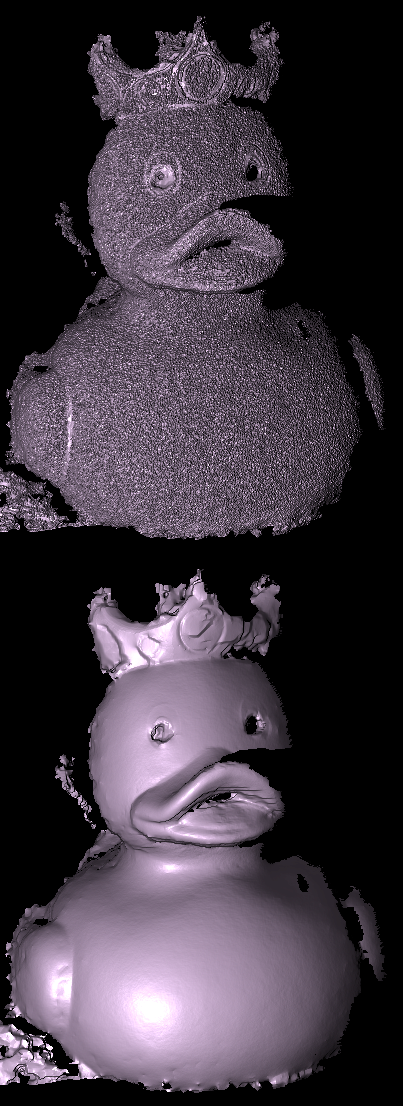}\\
			Geometry Regularizer
		\end{minipage}
	\end{subfigure}
	\vspace{-2mm}
	\caption{
		\textbf{Loss Regularizers in 2.5D,} results from Schmitt \etal~\cite{Schmitt2020CVPR}. 
		Shown are Reconstructions of held-out test views.
		Without regularization (top), appearance and geometry is inconsistent or noisy.
		Using the regularization terms (bottom), information is propagated across the object, successfully generalizing to new illumination conditions on the test set.
	}
	\label{fig:loss_regularizer_2.5d}
	\vspace{-2mm}
\end{figure}

\begin{figure}[t]
	\begin{minipage}{0.52\linewidth}
		\centering
		\includegraphics[width=\linewidth]{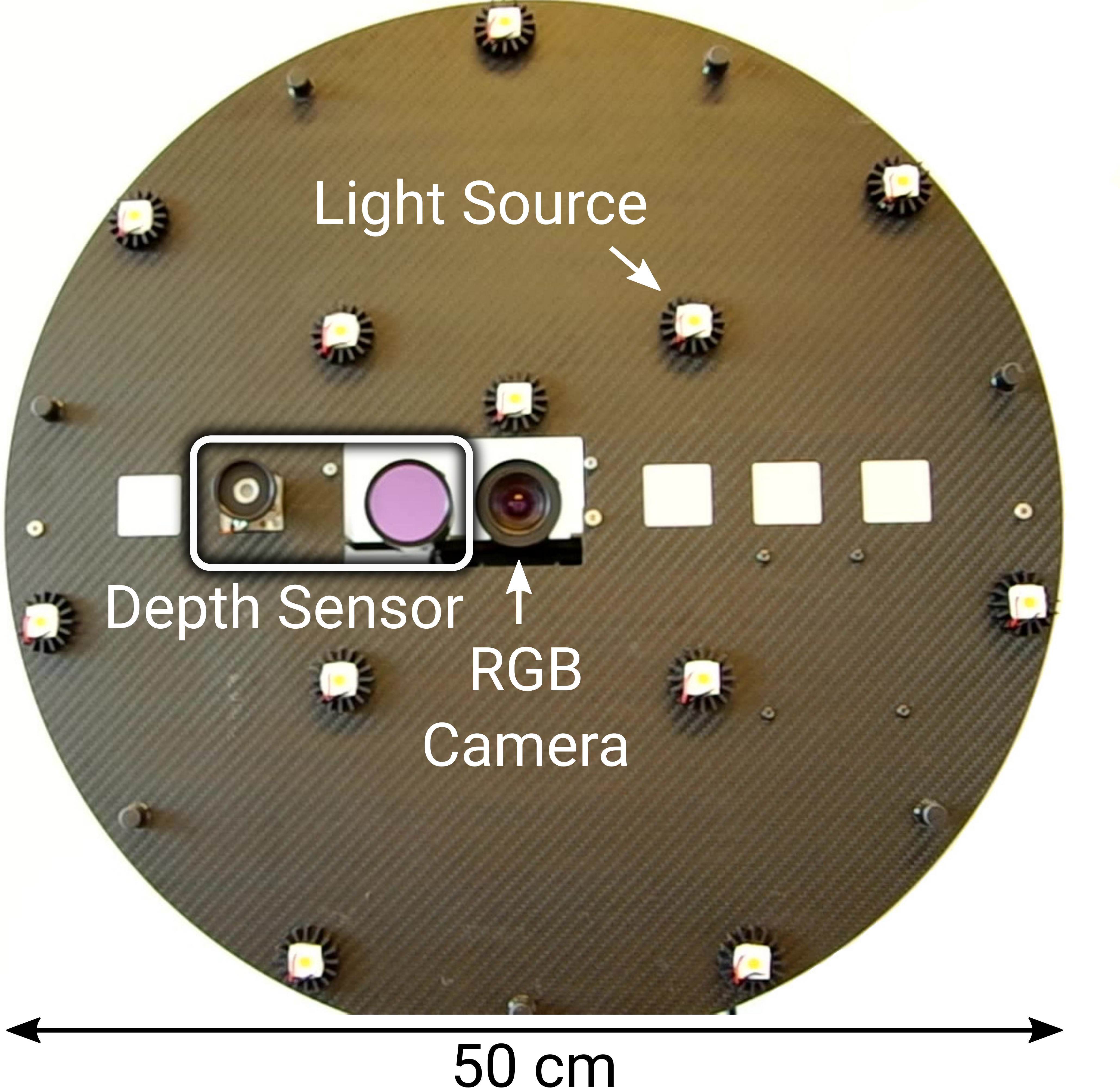}
	\end{minipage}
	\begin{minipage}{0.45\linewidth}
		\centering
		\caption{
			\textbf{Sensor Rig.}
			Our custom-made handheld capture device features a high resolution RGB camera, a Kinect-like active depth sensor and $12$ high-power LEDs (modeled as point light sources) that surround the camera in two circles (with radii $10$ cm and $25$ cm).
			\label{fig:results_sensor}
		}
	\end{minipage}
\end{figure}

\begin{figure}[t]
	\addtolength{\tabcolsep}{-5pt}
	\begin{tabular}{ccccc}
		\includegraphics[height=3cm]{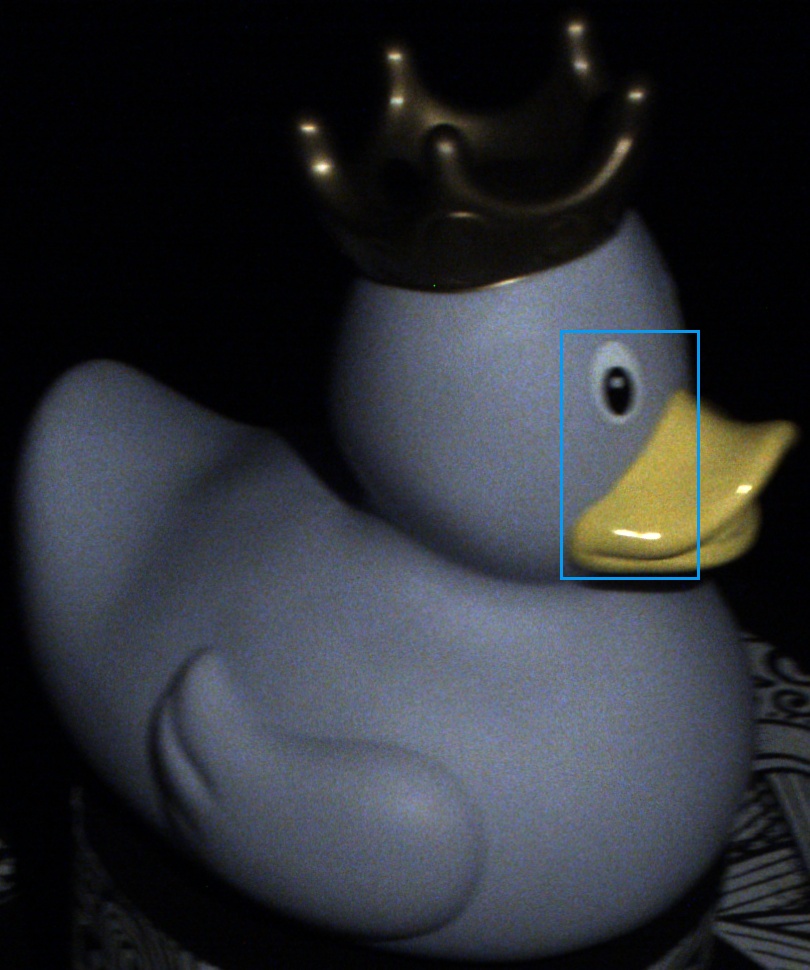} &
		\includegraphics[height=3cm]{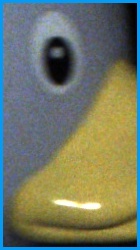} & $\:$ &
		\includegraphics[height=3cm]{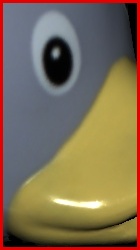} &
		\includegraphics[height=3cm]{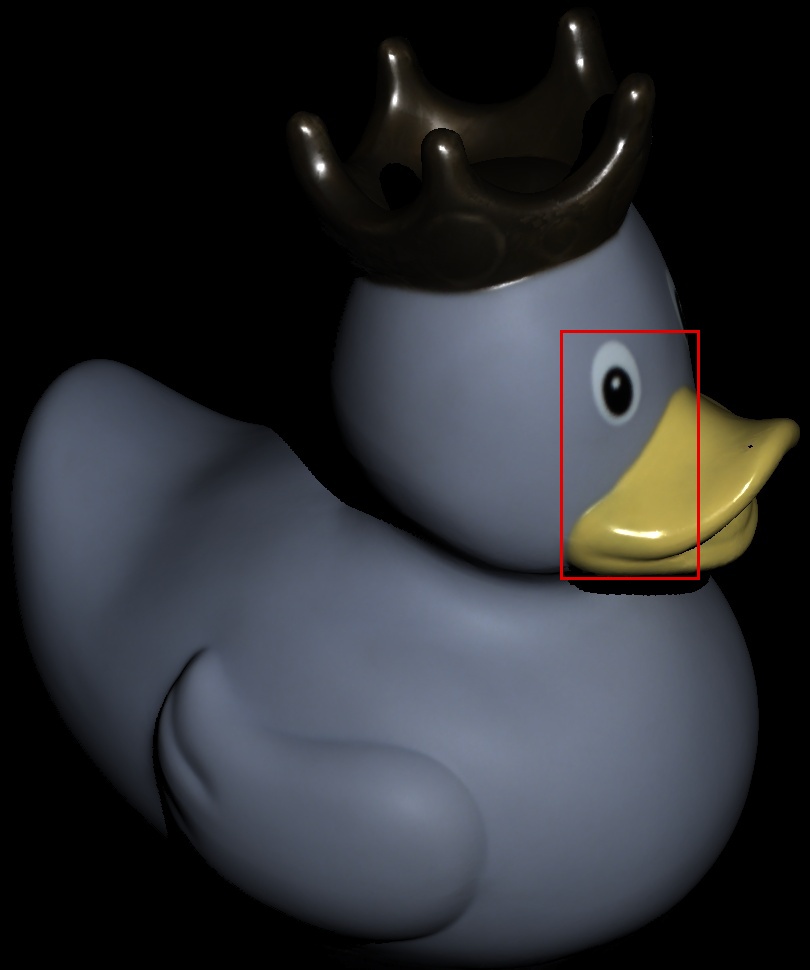}
		\\
		\includegraphics[height=3cm]{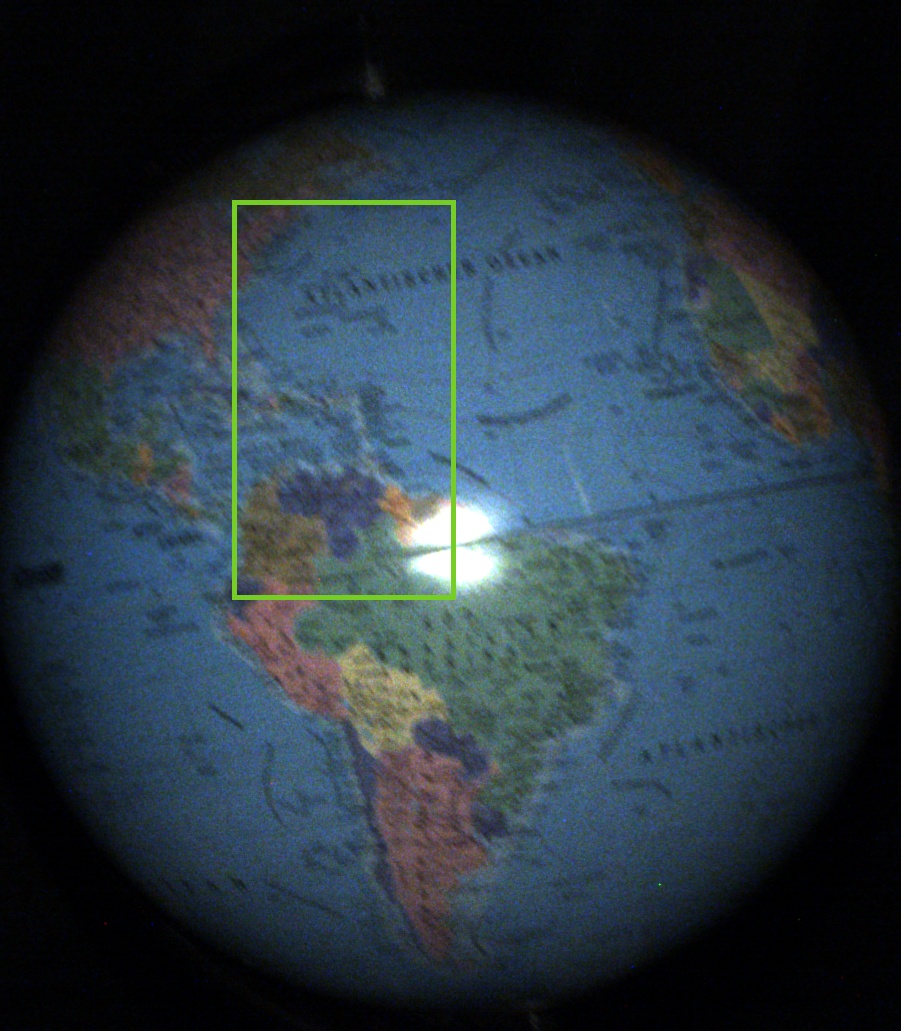} &
		\includegraphics[height=3cm]{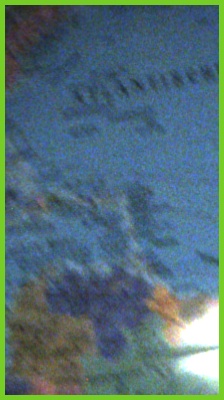} & $\:$ &
		\includegraphics[height=3cm]{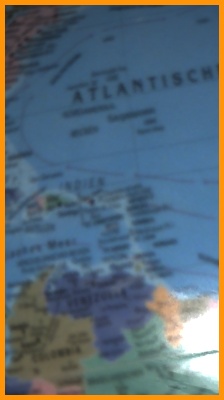} &
		\includegraphics[height=3cm]{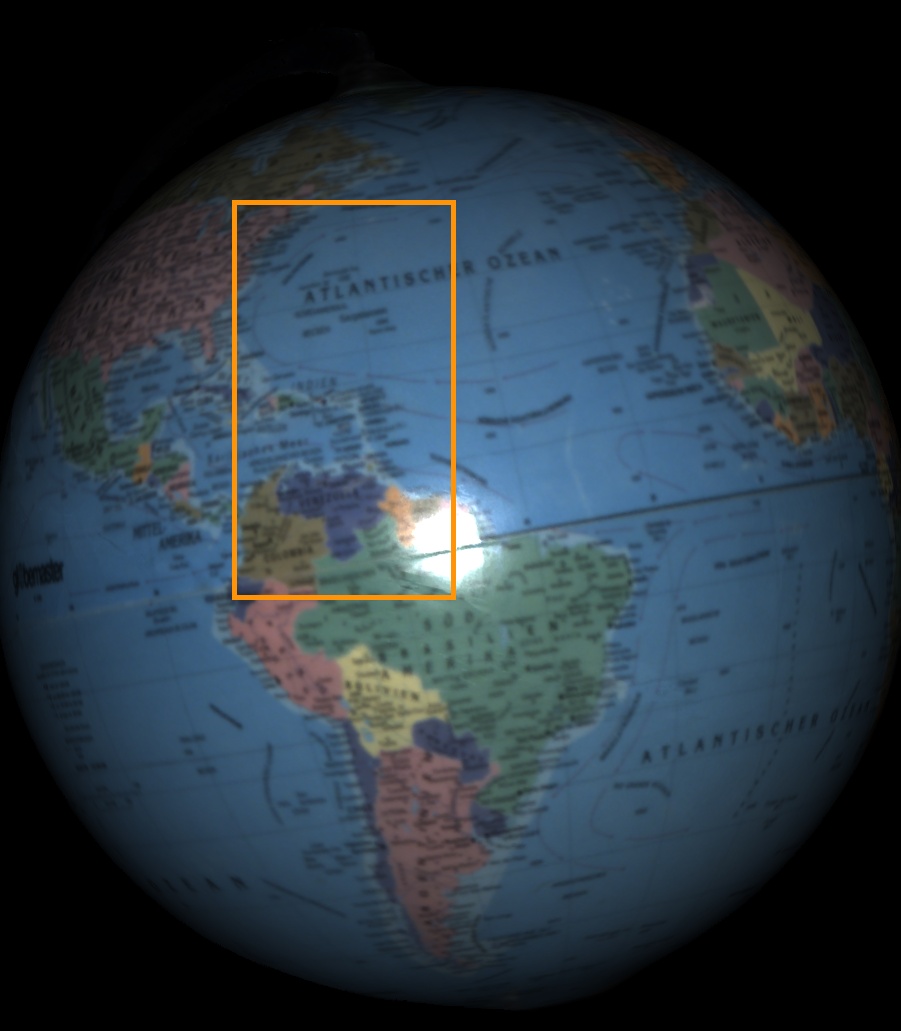}
		\\
		Observation & Zoom-In & & Zoom-In & Reconstruction
	\end{tabular}
	\addtolength{\tabcolsep}{5pt}
	\centering
	\caption{
		\textbf{Super-Resolution and Denoising (3D).}
		With a hand-held capture system, the measured observations exhibit image noise and motion blur (left) which our model is able to remove. 
		The resulting reconstructions appear denoised and sharpened (right).
	}
	\label{fig:results_RGBD_input}
\end{figure}

\begin{table}[t]
	\normalsize
	\centering
	\begin{tabular}{lcc}
\toprule
{} &  Fixed Poses &  Full Model \\
\midrule
Photometric Test Error &       18.767 &     17.8035 \\
\bottomrule
\end{tabular}

	\caption{
		\textbf{Pose Optimization (3D).} Similar to our conclusions in 2.5D (Schmitt \etal~\cite{Schmitt2020CVPR} and \figref{fig:poses_2.5d}), pose optimization also improves the fused 3D results of our full model.
	}
	\label{fig:poses}
\end{table}

\begin{figure}[t]
	\begin{subfigure}{\linewidth}
		\centering
		\begin{tabular}{lcc}
\toprule
{} &  No MV Consistency &  Full Model \\
\midrule
Photometric Test Error &              14.65 &      13.295 \\
\bottomrule
\end{tabular}

		\caption{RMSE on held-out test views, average over 3 objects.}
		\label{fig:ablation_mv_consistency_a}
		\vspace{2mm}
	\end{subfigure}
	\begin{subfigure}{\linewidth}
		\addtolength{\tabcolsep}{-6pt}
		\begin{tabular}{ccccccc}
			\rotatebox{90}{No MV Consist.} & $\;$ &
			\includegraphics[width=0.19\linewidth]{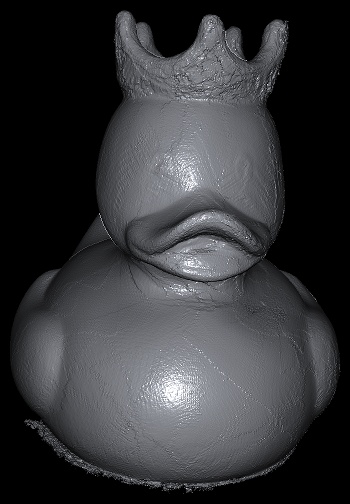} &
			\includegraphics[width=0.19\linewidth]{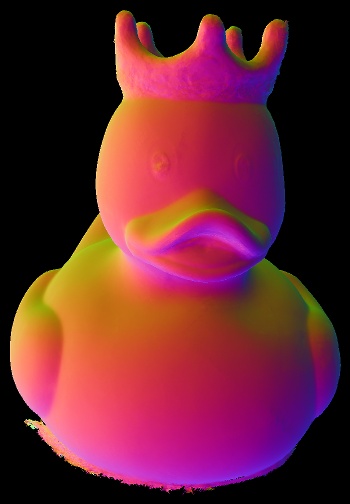} &
			\includegraphics[width=0.19\linewidth]{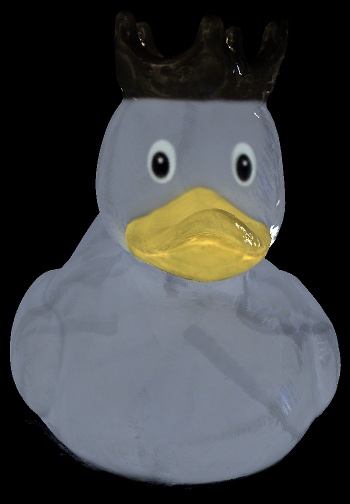} &
			\includegraphics[width=0.19\linewidth]{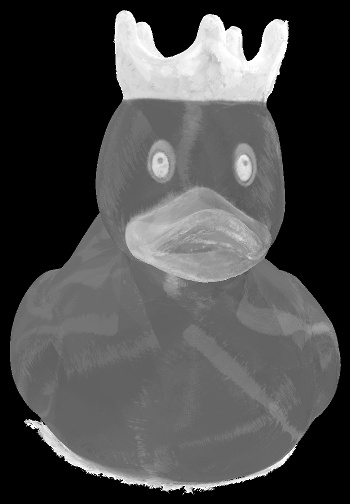} &
			\includegraphics[width=0.19\linewidth]{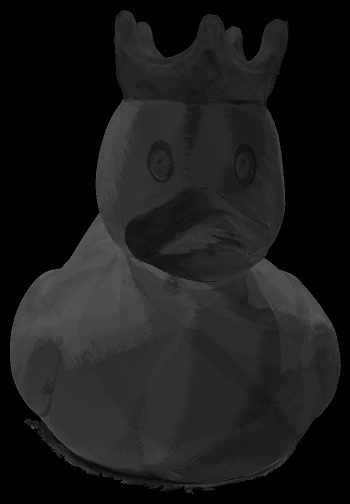} 
			\\
			\rotatebox{90}{$\;$ Full Model} & $\;$ &
			\includegraphics[width=0.19\linewidth]{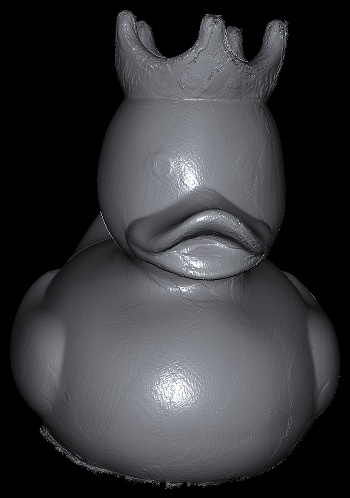} &
			\includegraphics[width=0.19\linewidth]{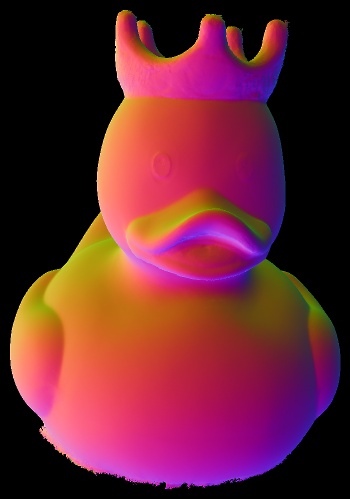} &
			\includegraphics[width=0.19\linewidth]{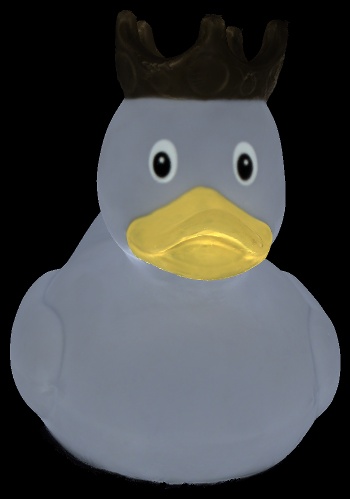} &
			\includegraphics[width=0.19\linewidth]{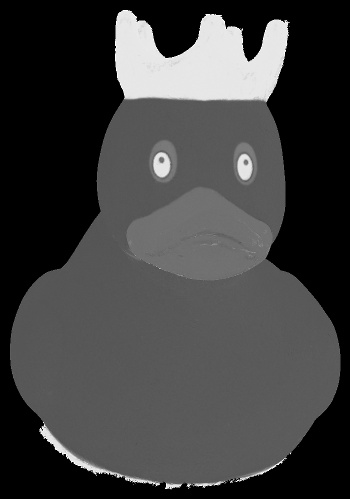} &
			\includegraphics[width=0.19\linewidth]{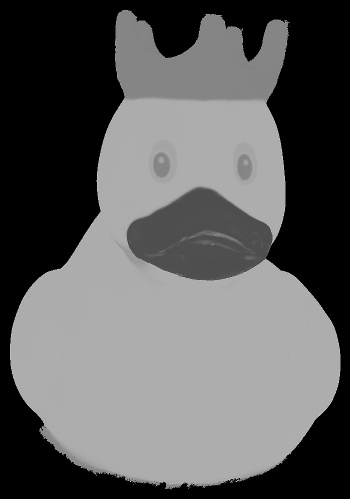} 
			\\
			&& Geometry & Normals & Diff. Alb. & Spec. Alb. & Roughness
		\end{tabular}
		\caption{Qualitative Comparison of Estimated Parameter Maps.}
		\label{fig:ablation_mv_consistency_b}
		\addtolength{\tabcolsep}{6pt}
	\end{subfigure}
	\centering
	\caption{
		\textbf{The Multi-View Consistency Loss (3D)} facilitates consistent parameter predictions across keyframes resulting in more accurate reconstructions (a). 
		In (b) we show parameter maps of the 3D fused mesh.
		Without the multi-view consistency regularizer (top), geometric artifacts are visible and the BRDF parameter maps show patch-like structures as well as baked-in shading information on very glossy object parts (\eg~beak and crown).
		In contrast, with our loss (bottom), the mesh is clean and reflectance maps are homogeneous per object part.
	}
	\label{fig:ablation_mv_consistency}
\end{figure}

\begin{figure}[t]
	\centering
	\addtolength{\tabcolsep}{-5pt}
	\begin{subfigure}{\linewidth}
		\begin{tabular}{ccc}
			\includegraphics[width=0.55\linewidth]{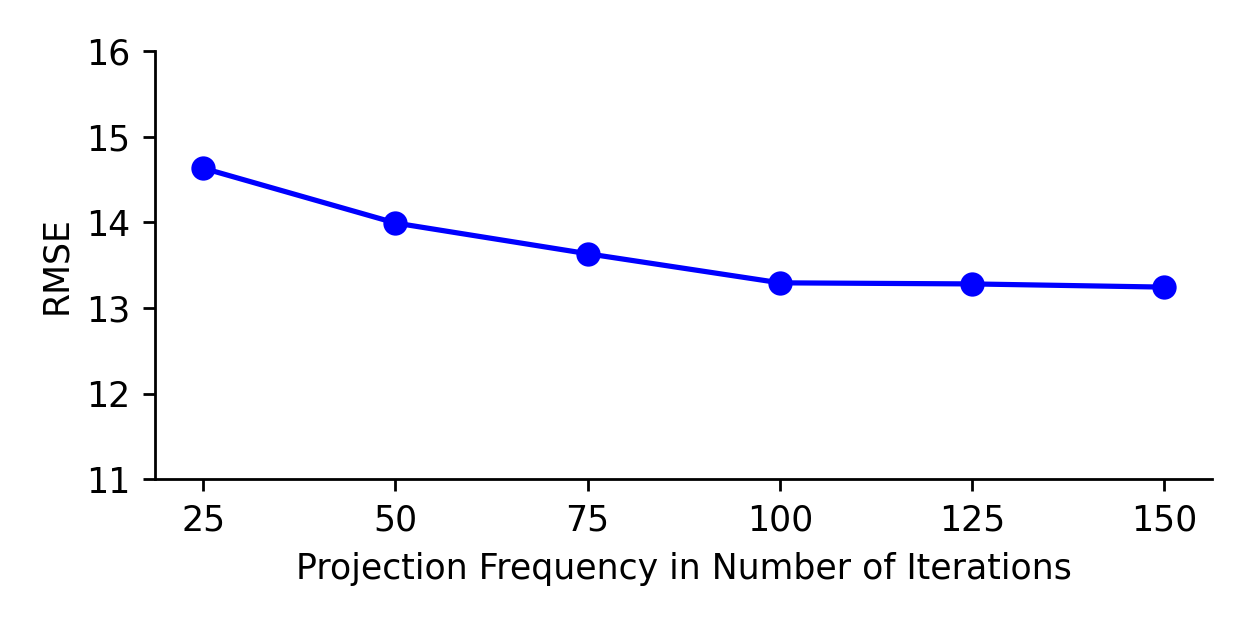} &
			\includegraphics[width=0.2\linewidth]{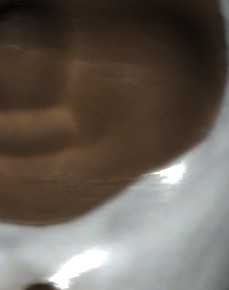} &
			\includegraphics[width=0.2\linewidth]{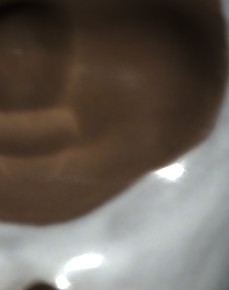}
			\\
			& $t=25$ & $t=100$ 	
		\end{tabular}
		\vspace{-2mm}
		\caption{Multi-View Projection Frequency}
		\vspace{1mm}
		\label{fig:ablation_mv_frequency}
	\end{subfigure}

	\begin{subfigure}{\linewidth}
		\begin{tabular}{ccc}
			\includegraphics[width=0.55\linewidth]{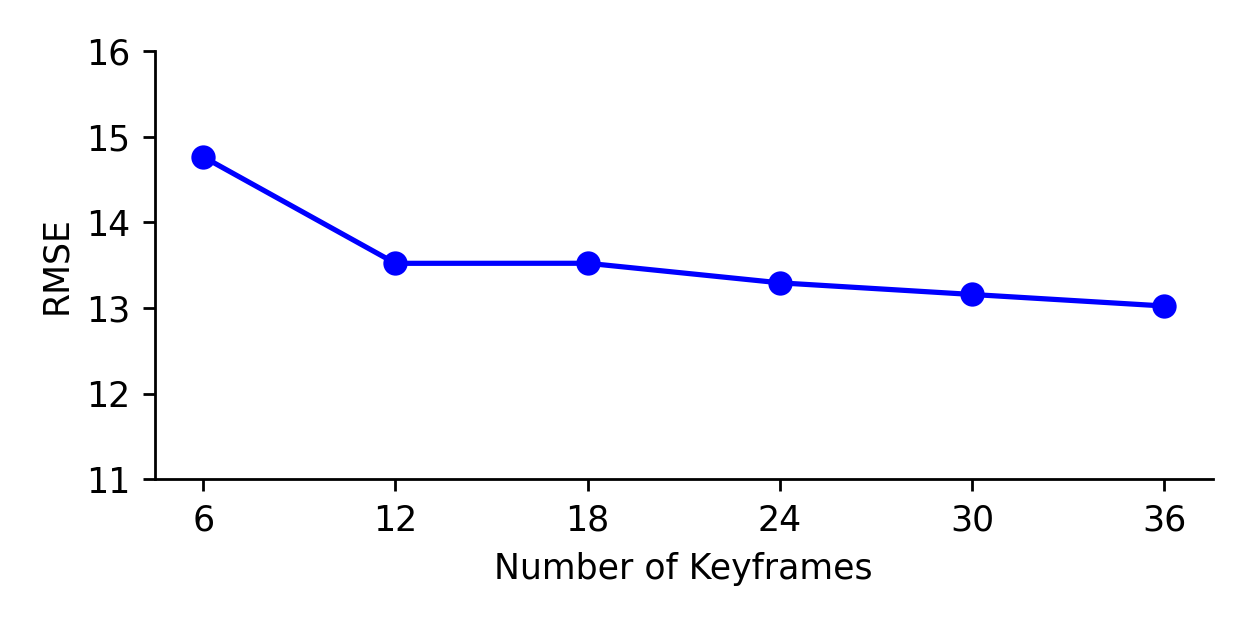} &
			\includegraphics[width=0.2\linewidth]{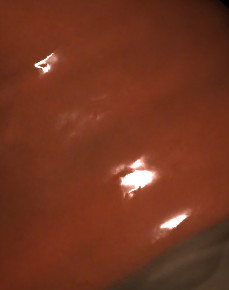} &
			\includegraphics[width=0.2\linewidth]{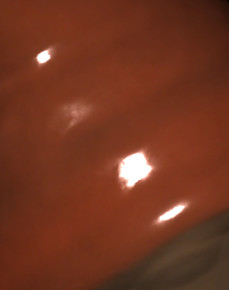}
			\\
			& $\vert K \vert =6$ & $\vert K \vert=24$ 	
		\end{tabular}
		\vspace{-2mm}
		\caption{Number of Keyframes}
		\vspace{1mm}
		\label{fig:ablation_keyframes}
	\end{subfigure}
	\begin{subfigure}{\linewidth}
		\begin{tabular}{ccc}
			\includegraphics[width=0.55\linewidth]{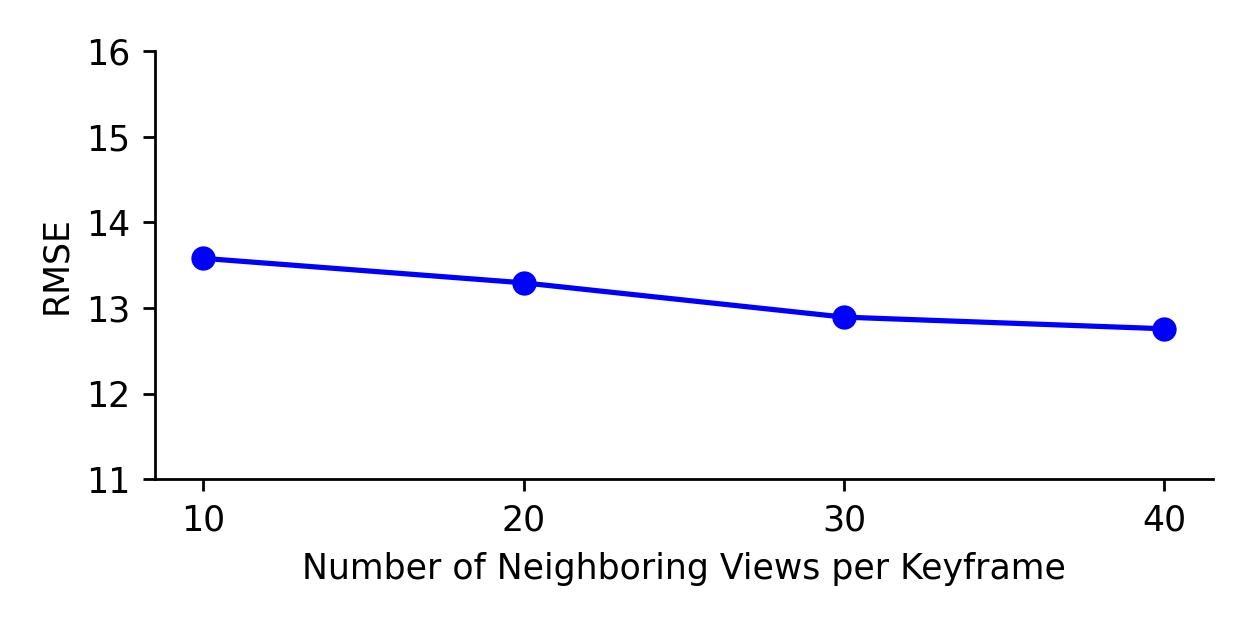} &
			\includegraphics[width=0.2\linewidth]{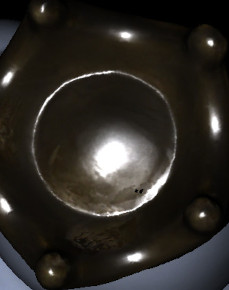} &
			\includegraphics[width=0.2\linewidth]{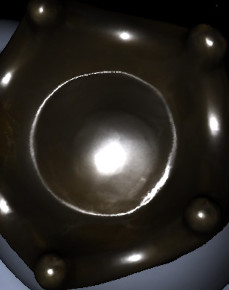}
			\\
			& $m=10$ & $m=20$ 	
		\end{tabular}
		\vspace{-2mm}
		\caption{Number of Neighboring Observation Views per Keyframe}
		\label{fig:ablation_neighbors}
	\end{subfigure}
	\addtolength{\tabcolsep}{5pt}
	\caption{
		\textbf{Multi-View Optimization (3D)}. 
		For multiple parameters we show the average error over 3 objects \wrt~the parameter on the left, and example predictions on the right.
		Hereby, the left image is a rendered result for the worst choice of parameters and the right image shows a rendered result for our chosen parameters.
	}
\end{figure}

\section{Experimental Evaluation}
\label{sec:results}

The proposed method is an extension of the 2.5D reconstruction algorithm presented by Schmitt \etal~\cite{Schmitt2020CVPR} to full 3D models.
Due to space limitations, we do not repeat all experiments conducted in our conference paper, but  instead review the main conclusions and insights. For further details, we kindly refer the reader to \cite{Schmitt2020CVPR}.

In \cite{Schmitt2020CVPR}, we propose a formulation for joint recovery of camera pose, object geometry and spatially-varying BRDF from handheld capture data.
Instead of using multiple decoupled objectives and treating materials and geometry separately as done before, we demonstrate in the conference paper that this problem can be formulated using a single objective function and off-the-shelf gradient-based solvers.
Except for a few minor differences described at the end of this section, this model corresponds to the model described in \secref{sec:optimization_objective} when excluding the multi-view consistent optimization and instead optimizing only a single keyframe, yielding a 2.5D result. As shown in the ablation study of Schmitt \etal~\cite{Schmitt2020CVPR}, two components are particularly important for accurate appearance and geometry reconstructions:
1) Optimizing poses jointly with the other parameters is crucial for disambiguating geometric properties from materials. 
We repeat the ablation results in \figref{fig:poses_2.5d}. 
2) The proposed regularization terms for the geometric parameters and material maps enable joint optimization over all parameters using a single objective function. 
Hereby, the material smoothness term is able to propagate material information over large distances and compensates for the sparse measurements of the BRDF per pixel.
And the geometric consistency term enforces consistency between depth and normals and prevents high-frequency structure artifacts. 
The results are shown in \figref{fig:loss_regularizer_2.5d}.

In this paper, we demonstrate that simple fusion of multiple 2.5D reconstructions obtained using \cite{Schmitt2020CVPR} is insufficient to obtain accurate 3D models of an object.
In particular, the geometry from different keyframes does not align well enough and material predictions from different viewpoints often differ noticeably due to the ambiguities present in this inverse problem.
We therefore introduce a multi-view consistent optimization scheme in this extension, and demonstrate that this enables consistent and accurate reconstructions of 3D models.
We further demonstrate that this allows for modeling larger scenes beyond single objects.
In comparison to \cite{Schmitt2020CVPR}, we also make several minor improvements to the model which we empirically found to be useful:
1) We change the material model to a more flexible and practical solution for larger scenes that does not require material clustering and model selection, 
2) we regularize the depth maps against all neighboring depth maps instead of a single one to stay closer to the measurements and
3) we include an edge-aware weighting term in the normal smoothness loss to facilitate reconstruction of small details.
\\

In the following, we present the results of our extended model in 3D and provide an evaluation on captures of real objects and scenes from a custom built handheld sensor rig.
We first introduce our hardware system and the data capture procedure and then provide details on our evaluation protocol.
Afterwards, we conduct an ablation study of the components of our method.
We then provide qualitative and quantitative comparisons to related approaches and conclude with reconstruction results for our captured dataset.
Note that unlike in \cite{Schmitt2020CVPR} and in \figref{fig:poses_2.5d}, \ref{fig:loss_regularizer_2.5d}, all of the following results show fused 3D models unless explicitly stated otherwise.
Further, we present results of our method on synthetic data in the supplement.

\subsection{Setup}
\label{sec:results_setup}

For capturing data, we use a custom-built handheld sensor rig as shown in \figref{fig:results_sensor}.
We thoroughly calibrate the system in advance, both geometrically and photometrically. 
We estimate camera instrinsics, response, distortion and vignetting, the relative positions of the depth sensor and light sources \wrt~the RGB camera, and the angular attenuation behavior and radiant intensities of the light sources.

We slowly move our sensor around the scene and alternate the illumination such that each image is illuminated by exactly one light source.
We assume a completely darkened room with negligible ambient light. 
Examples of captured raw data is shown in \figref{fig:results_RGBD_input} (left).
Note that due to the handheld setup, we need to accept a certain amount of image noise to trade-off motion blur.
But we show in \figref{fig:results_RGBD_input} (right) that our model is able to predict denoised and sharpened reconstructions.
If the scene is not sufficiently textured, we additionally add texture patterns to the scene to ensure enough feature points and obtain more reliable initial pose estimates.

We use full image resolution (4K) for all objects and half image resolution (2K) for scenes due to GPU memory limitations.
For all our objects and scenes, we captured between $800$ - $1400$ images.

\boldparagraph{Evaluation Protocol} 
\label{sec:results_evaluation}
For quantitative evaluation, we render the final model in 10 held-out test views and compute the photometric loss with respect to the observation. 

As our model can deviate from the coarse initial COLMAP poses during optimization, we first align the test view poses to the predicted model. 
Next, we compute the Root Mean Square Error over all pixels that have a non-zero color prediction for each test view and report the mean loss. 
Since for real capture data there exist no ground truth object masks, we define valid image pixels as pixels with non-zero prediction (prediction mask).
For validation, we draw the observation masks by hand for multiple objects and find that the RMS errors for both evaluation masks differ by $<5\%$.
Therefore, we use the prediction mask in the following for all experiments for simplicity.

\subsection{Ablation Study}
\label{sec:results_ablation}

In this section, we ablate the important parts of our model, both qualitatively and quantitatively. 
An additional ablation on the loss weights can be found in the supplementary.

\boldparagraph{Multi-View Consistency Loss}
\label{sec:results_ablation_consisteny_loss}
Multiple local reconstructions of our method share the coarse but consistent captured depth maps, but observe different samples of the reflectance function since the captured images contain only sparse measurements thereof.
That implies that while the 2.5D keyframe optimizations lead to consistent geometry with respect to immediate neighbors, they may lead to inconsistencies \wrt~keyframes that are not optimized. 
And these inconsistencies cannot be resolved reliably using volumetric fusion which merely averages multiple geometries.
Thus, there is no guarantee for consistent reconstruction results without explicitly enforcing consistency between keyframes.

\figref{fig:ablation_mv_consistency} demonstrates the effectiveness of the multiview consistency loss. 
It encourages both regularization and propagation between keyframes.
Synchronization of parameter estimates during optimization enables our method to find an equilibrium of the variables which is consistent with not only its neighboring observation views but also the neighboring observation views of nearby keyframes.
Hereby, these connections between neighboring keyframes form a connected graph over all keyframes. Therefore, consistency between any two keyframes can be penalized during optimization, which is enforcing global consistency.
Without the multi-view consistency loss term, the fused model is not aligned well enough to form one coherent surface or consistent BRDF parameter maps.
In contrast, we observe that our distributed multi-view consistent optimization leads to globally consistent results without blending artifacts as illustrated in \figref{fig:ablation_mv_consistency_b}.
We note that, in particular, specular properties (specular albedo and roughness) are robustly reconstructed despite the sparsely sampled reflectance function and the high sensitivity of specular highlights to angular configurations.
This leads to lower reconstruction errors as evidenced in \tabref{fig:ablation_mv_consistency_a}.

\boldparagraph{Pose Optimization} \label{sec:ablation_poses}
Misaligned camera poses yield wrong correspondences between view pairs and can cause various reconstruction artifacts such as ghosting, blur, and texture/geometry bleeding between front and back surfaces.
With respect to material reconstruction, wrong pose estimates lead to errors in the prediction of angular relations between the surface normals, view and light directions.
This causes estimated specular highlights to not align with the mirror reflection direction and hence leads to highlights not being recovered and bake-in effects of specular appearance into predicted texture and normal maps. 
Such pose alignment problems are particularly crucial when working with a moving handheld scanner.
Therefore, we optimize the camera poses jointly with the other parameters leading to more consistent results and lower reconstruction errors.
We show these findings quantitatively for our fused 3D models in \tabref{fig:poses}, confirming the 2.5D results in \figref{fig:poses_2.5d} from Schmitt \etal~\cite{Schmitt2020CVPR}.

\boldparagraph{Multi-View Parameter Projection Frequency} 
\label{sec:results_ablation_mv_frequency}
In a single optimization round, we optimize the parameters of all keyframes for $t$ iterations before synchronizing with neighboring keyframes. 
Therefore, $t$ balances local reconstruction quality and global parameter consistency.
A low number of $t$ or high synchronization frequency hinders the local optimizations to fit the neighboring observations as shown in \figref{fig:ablation_mv_frequency} (right), whereas a low frequency or no synchronization ($t=T$) prevents consistency among the \textit{current} parameter estimates of neighboring keyframes, as discussed in \figref{fig:ablation_mv_consistency}.
We found that $t = 100$ leads to both accurate parameter estimates and consistent results across keyframes.

\boldparagraph{Number of Keyframes} 
\label{sec:results_ablation_keyframes}
\figref{fig:ablation_keyframes} plots test accuracy against the number of keyframes $\vert K \vert$. 
We observe that generally, more keyframes lead to more accurate reconstructions and most affected by a small number of keyframes is the quality of the predicted highlights.
This offers several insights as it indicates that 
1) local keyframe results are globally consistent also for larger numbers of keyframes, 
2) details are preserved and the geometry is not noticeably blurred during mesh fusion and 
3) synchronization with neighboring keyframes is important for correct reflectance estimation (with a reduced number of keyframes, the number of possible neighboring keyframes decreases as well).
We use $\vert K \vert = 24$ in the following for all single objects as the performance gain becomes very small thereafter.

\boldparagraph{Number of Neighboring Observation Views} 
\label{sec:results_ablation_neighbors}
Our goal is to estimate the spatially varying BRDF but we only observe a very sparse set of samples for each surface point $\bx$. 
As expected, we see in \figref{fig:ablation_neighbors} that reducing the number of neighboring observation views $m$ worsens this problem.
However, interestingly, this effect is quite small and the error degrades gracefully.
We attribute this to our multi-view consistent optimization scheme which regularly provides information from neighboring keyframes during optimization.
Therefore, given a sufficiently large number of keyframes, our method produces accurate predictions already for $m=20$.

\begin{figure}[t]
	\centering
	\begin{subfigure}{\linewidth}
		\addtolength{\tabcolsep}{-5pt}
		\begin{tabular}{ccccc}
			& &  Owl & Sheep & Globe
			\\ 
			\rotatebox{90}{$\quad\:$ Observation} & 
			\rotatebox{90}{$\qquad\;\;\;$ (2D)} &
			\includegraphics[height=2.65cm]{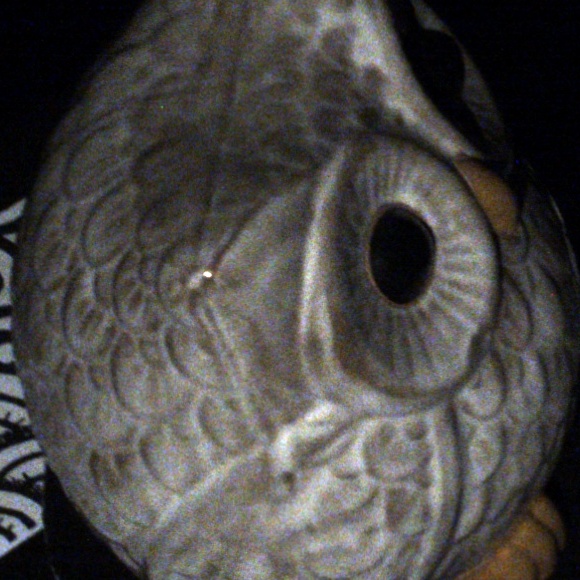} &
			\includegraphics[height=2.65cm]{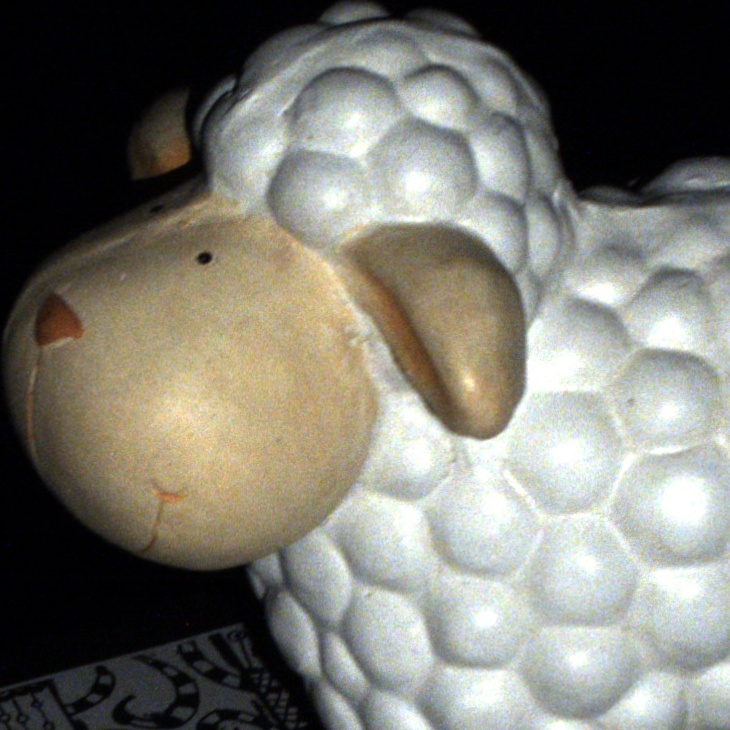} &
			\includegraphics[height=2.65cm]{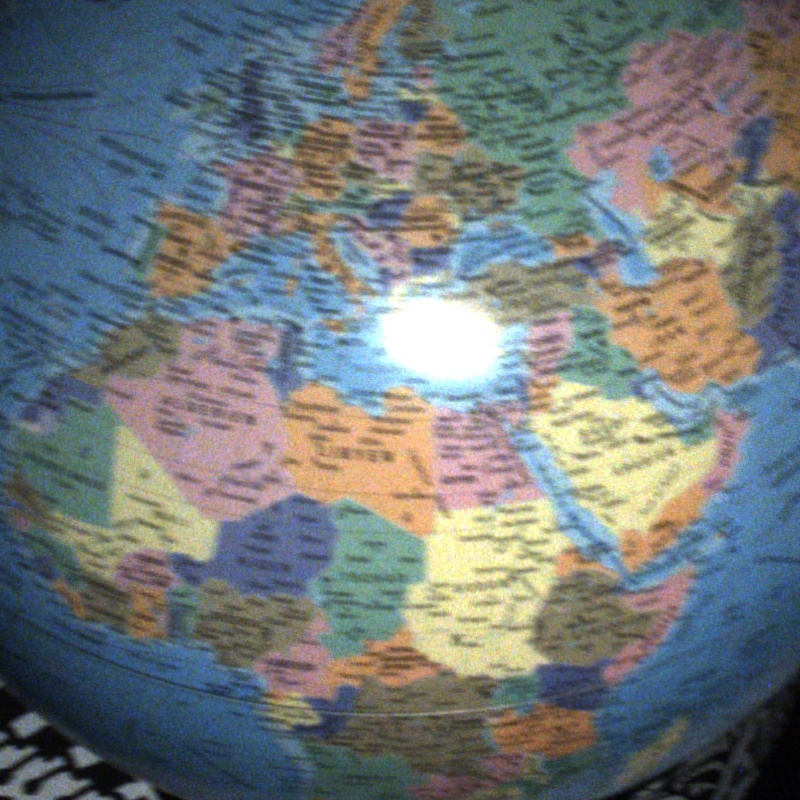} 
			\\
			\rotatebox{90}{TSDF Fusion \cite{Zeng2017CVPR}} &
			\rotatebox{90}{$\qquad\;\;\;$ (3D)} &
			\includegraphics[height=2.65cm]{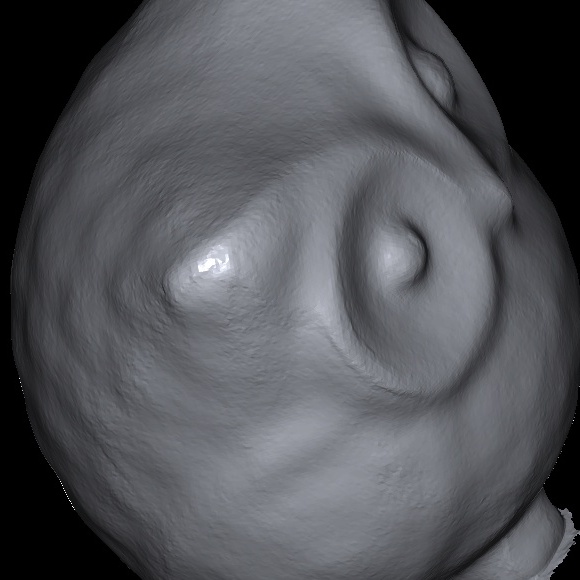} &
			\includegraphics[height=2.65cm]{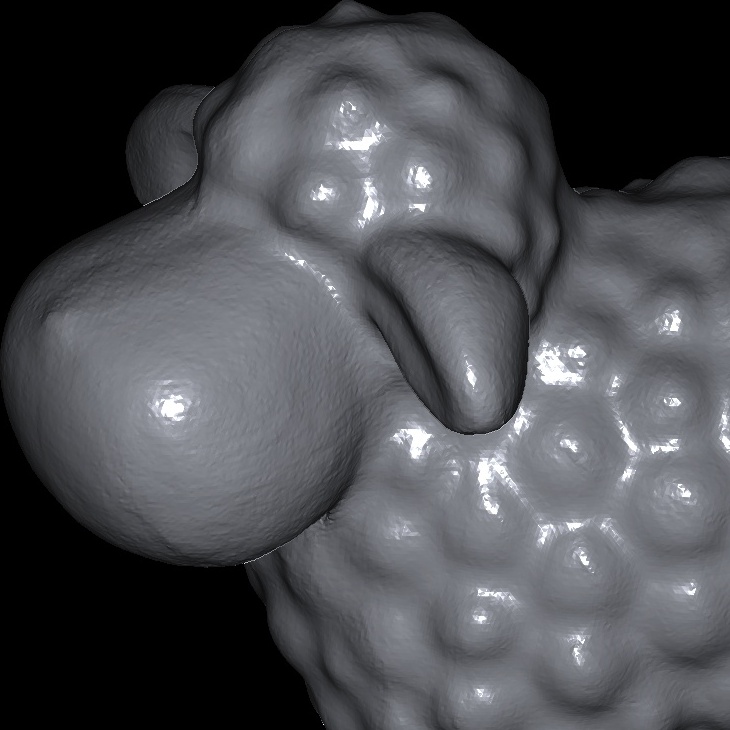} &
			\includegraphics[height=2.65cm]{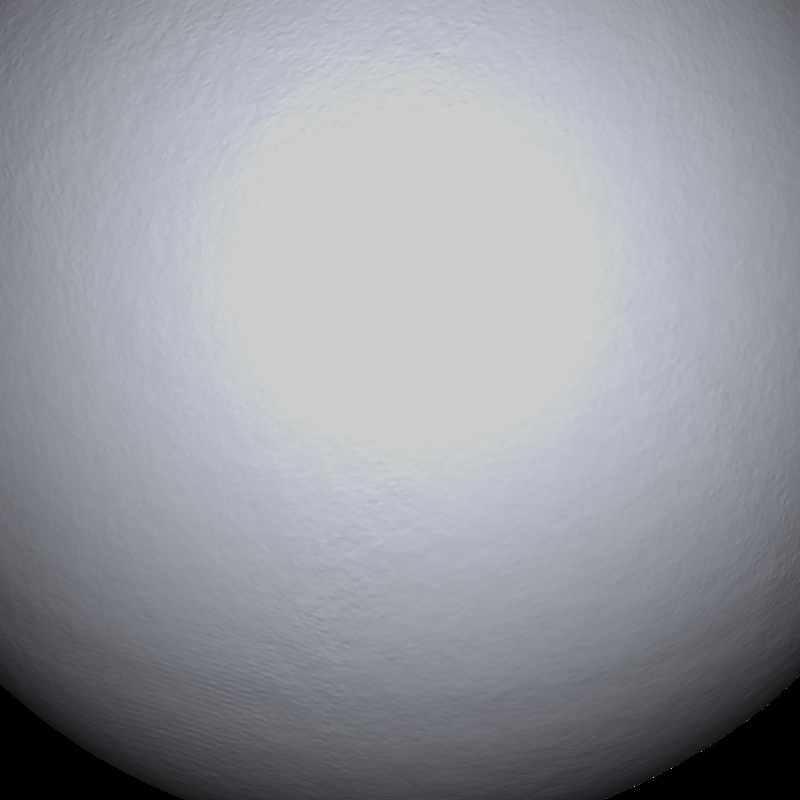} 
			\\
			\rotatebox{90}{$\:$ Schmitt \etal \cite{Schmitt2020CVPR}} &
			\rotatebox{90}{$\qquad\;$ (2.5D)} &
			\includegraphics[height=2.65cm]{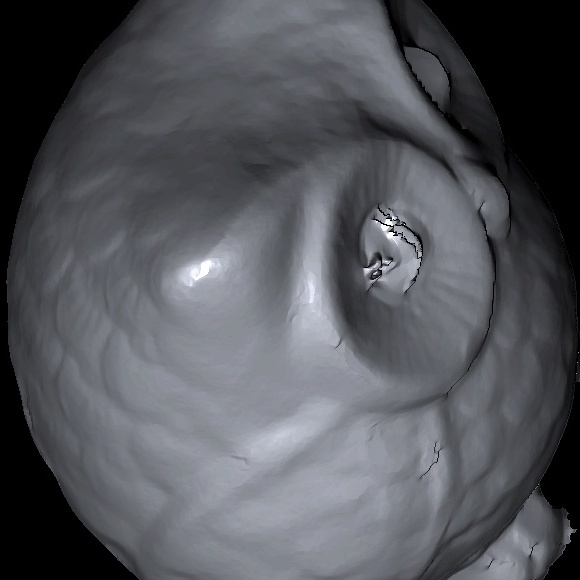} &
			\includegraphics[height=2.65cm]{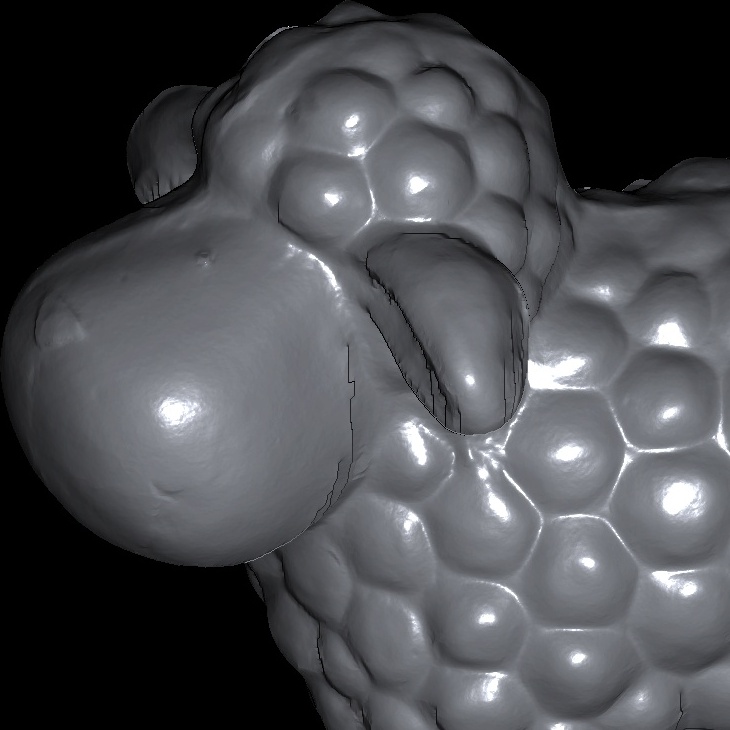} &
			\includegraphics[height=2.65cm]{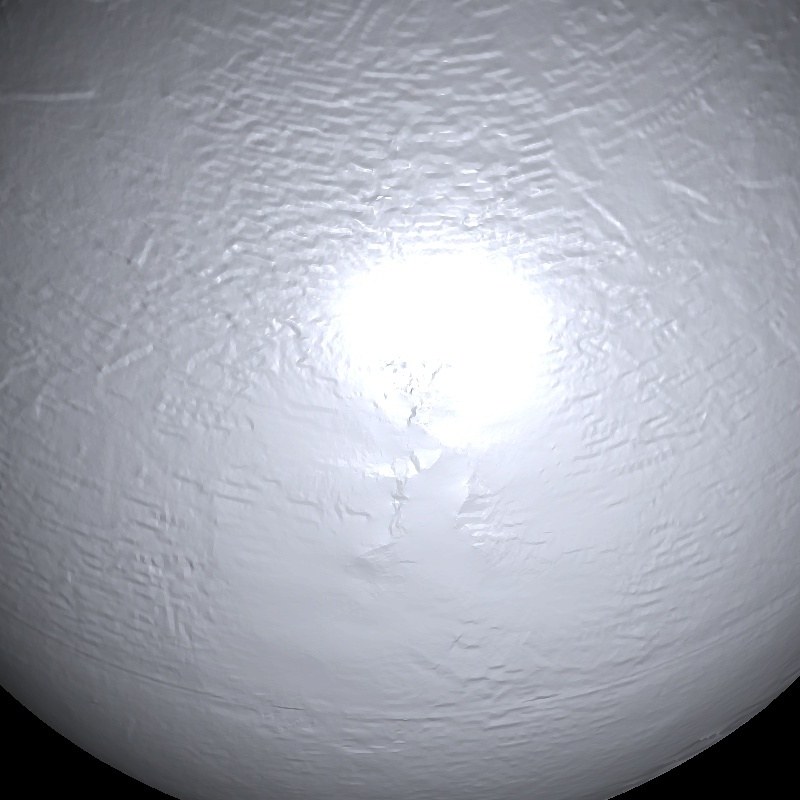} 
			\\
			\rotatebox{90}{$\;$ Single Keyframe} &
			\rotatebox{90}{$\qquad\;$ (2.5D)} &
			\includegraphics[height=2.65cm]{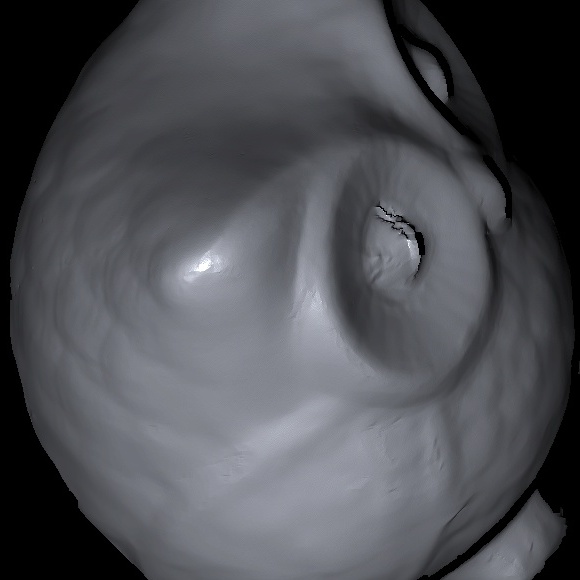} &
			\includegraphics[height=2.65cm]{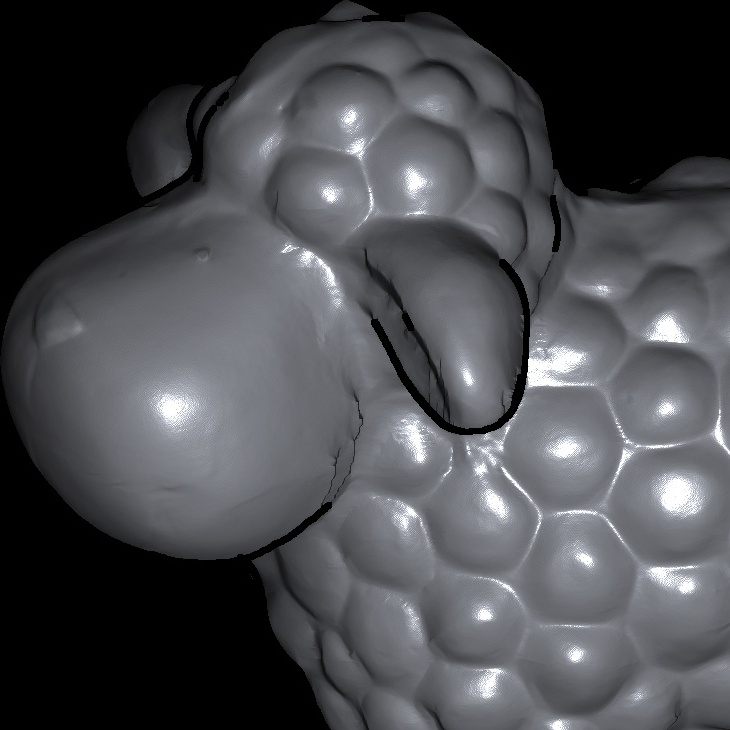} &
			\includegraphics[height=2.65cm]{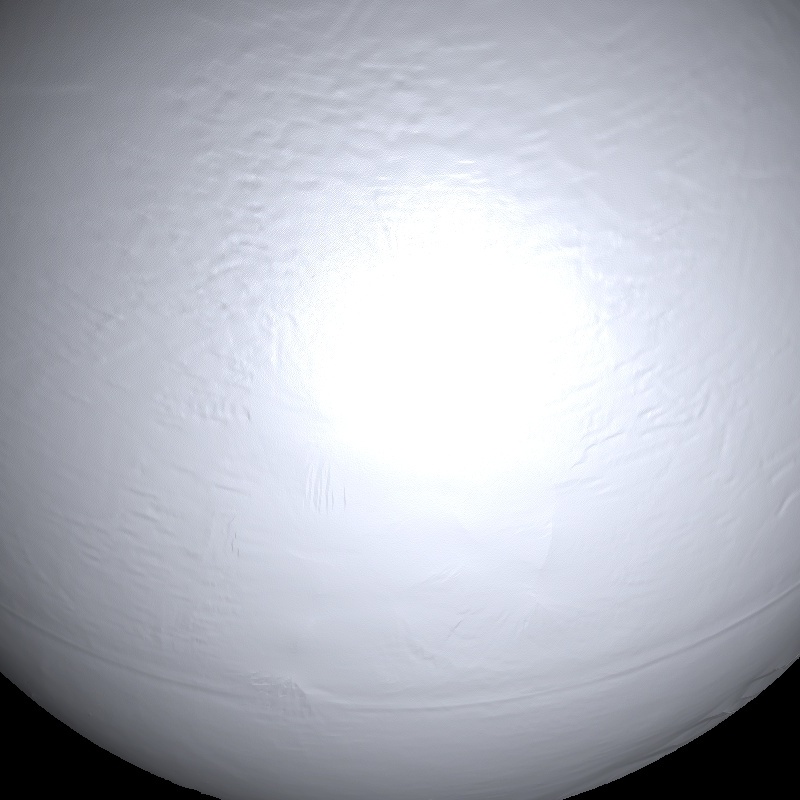} 
			\\
			\rotatebox{90}{$\qquad$ Proposed}&
			\rotatebox{90}{$\qquad\;\;\;$ (3D)} &
			\includegraphics[height=2.65cm]{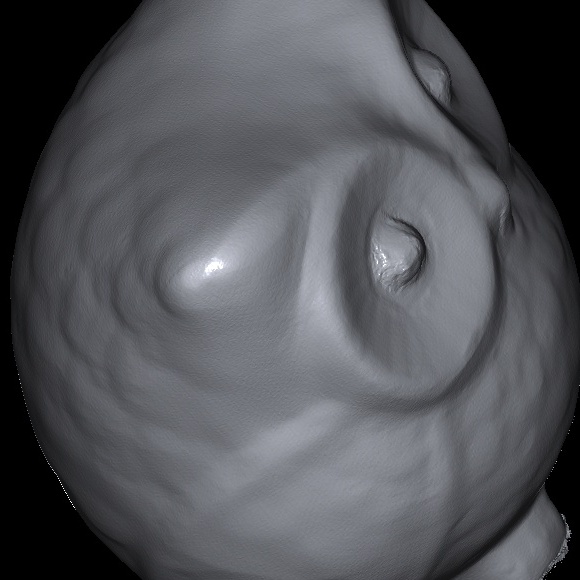} &
			\includegraphics[height=2.65cm]{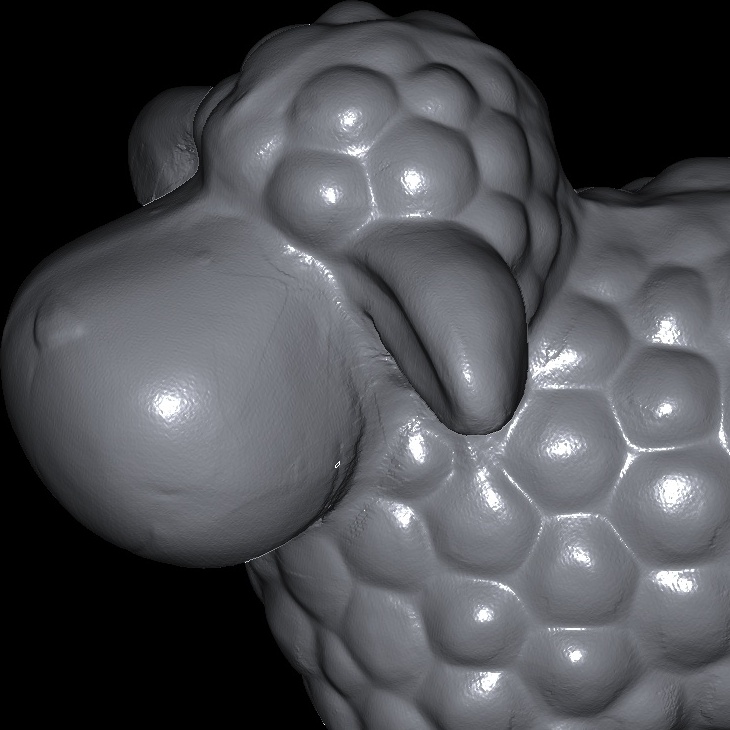} &
			\includegraphics[height=2.65cm]{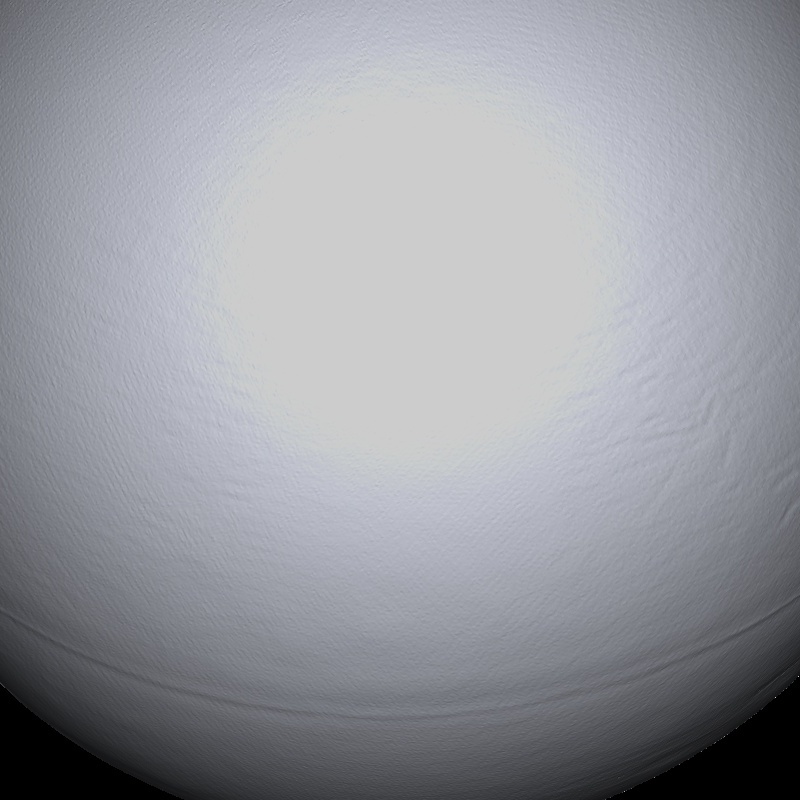} 
		\end{tabular}
		\addtolength{\tabcolsep}{5pt}
	\end{subfigure}
	\caption{
		\textbf{Qualitative Geometry Comparison (2.5D vs. 3D).} 
		We show, for each object, the rendered depth map (shaded based on estimated surface normals) for the 3D model after TSDF fusion, the 2.5D model by Schmitt \etal~and both a 2.5D keyframe as well as the full 3D model of our proposed method.
		We observe that the photometric approaches recover more details than na\"{i}ve TSDF fusion of the input geometry.
		Thanks to our multi-view consistent optimization scheme, the single keyframe result of the proposed approach contains less textural artifacts in the geometry than Schmitt \etal~(see \eg~the `Globe').
		Further, we observe that for our model the resulting global 3D geometry is as detailed as the geometry of the 2.5D keyframe, and additionally resolves artifacts present in the local reconstruction, \ie~the eye of the `Owl'.
	}
	\label{fig:results_geometry}
\end{figure}

\begin{figure}[t]
	\centering
	\begin{subfigure}{\linewidth}
		\addtolength{\tabcolsep}{-5pt}
		\begin{centering}
		\begin{tabular}{cccc}
			& Schmitt \etal~\cite{Schmitt2020CVPR} & Proposed & Observation \\
			& (Fused in 3D) & (3D) & (2D) 
			\\
			\rotatebox{90}{$\;\;$ Girl} &
			\includegraphics[width=0.3\linewidth]{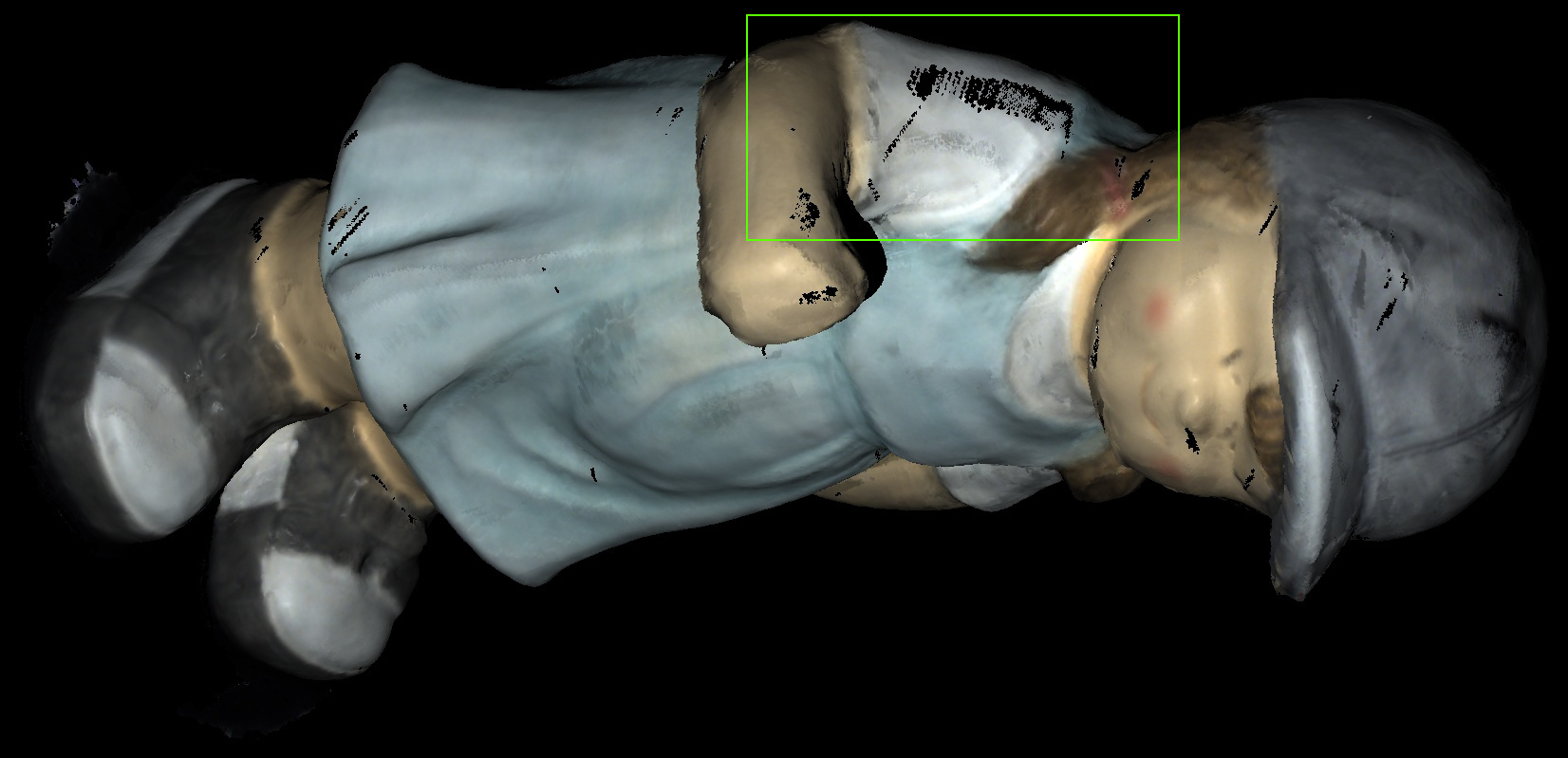} &
			\includegraphics[width=0.3\linewidth]{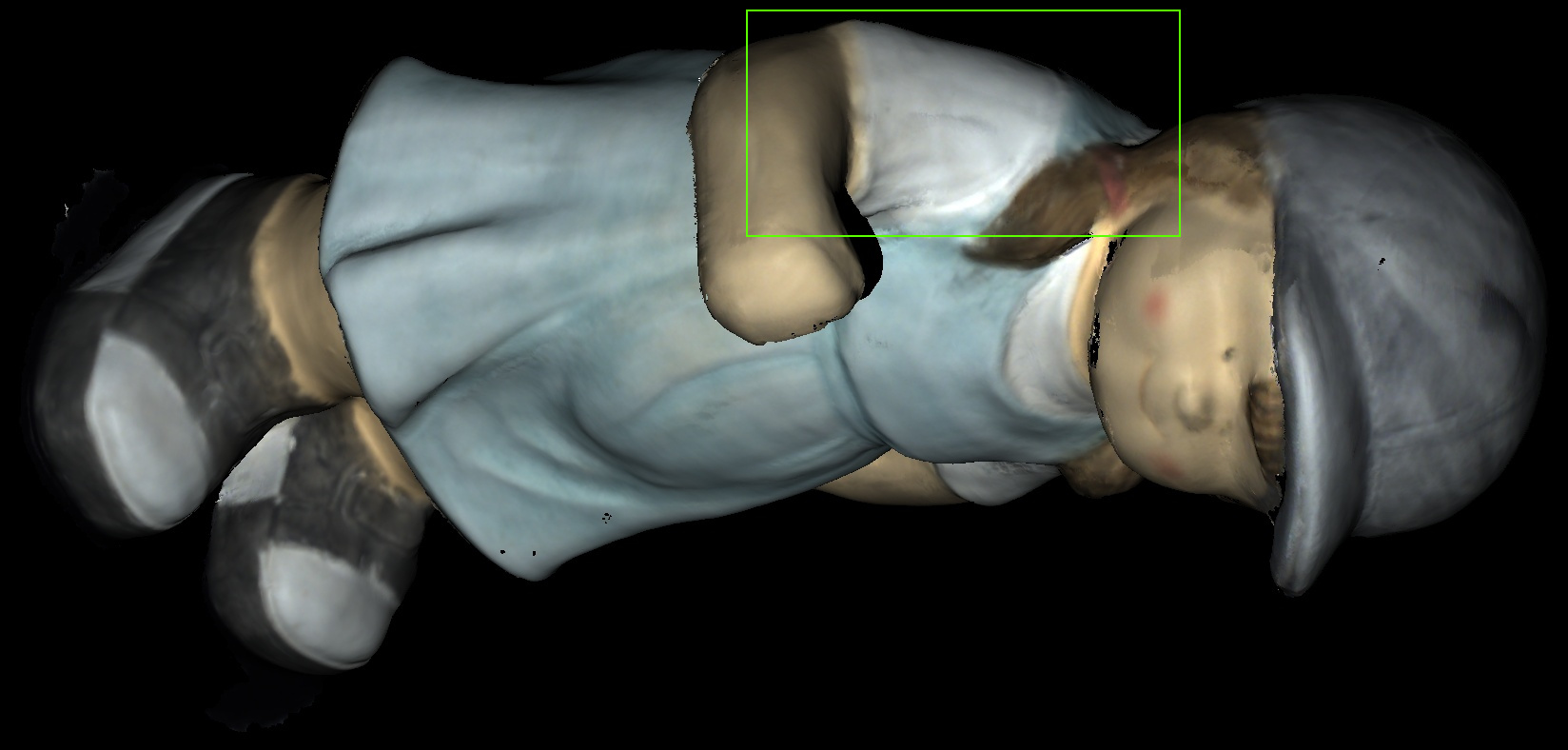} &
			\includegraphics[width=0.3\linewidth]{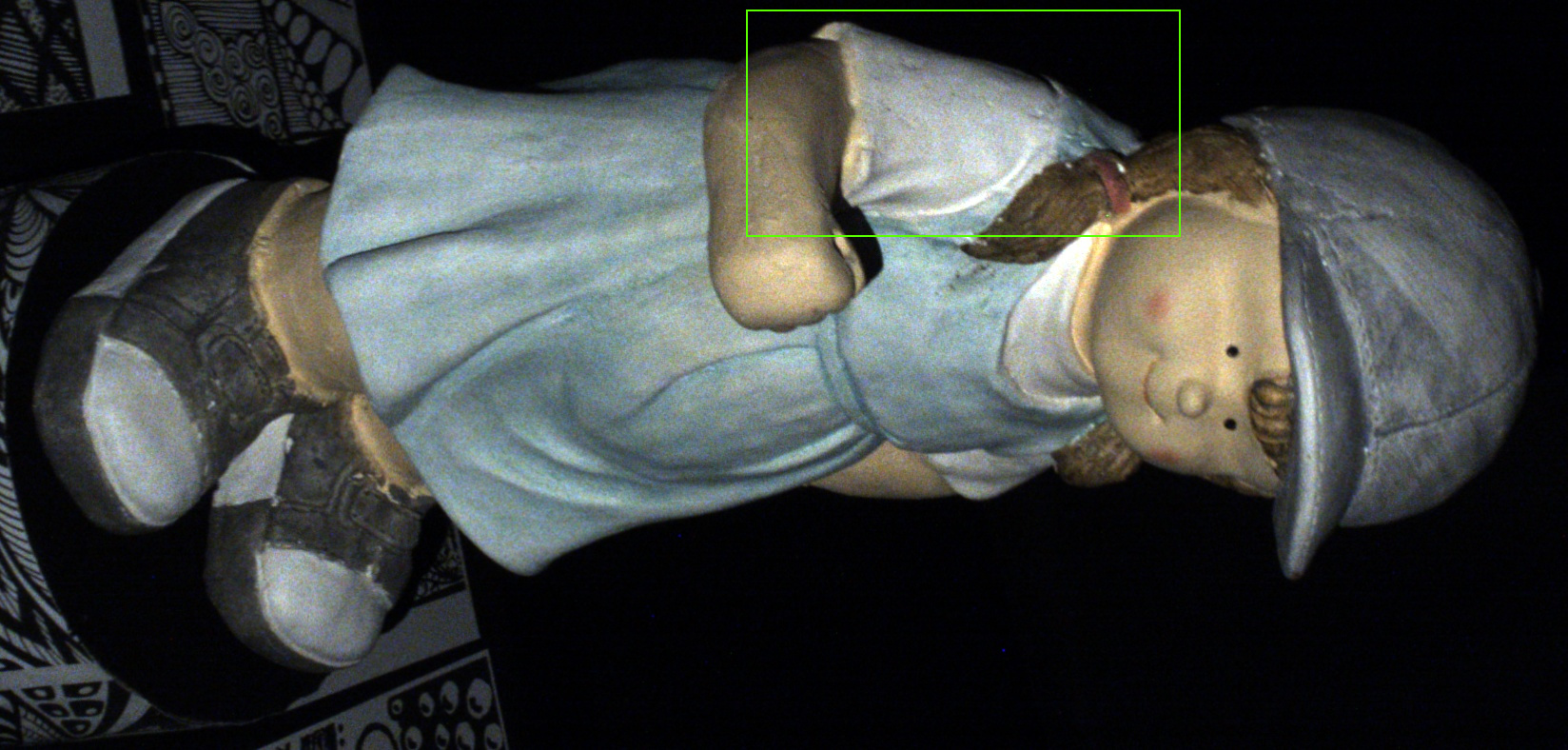} 
			\\
			\rotatebox{90}{$\quad\;\;$ Teapot} &
			\includegraphics[width=0.3\linewidth]{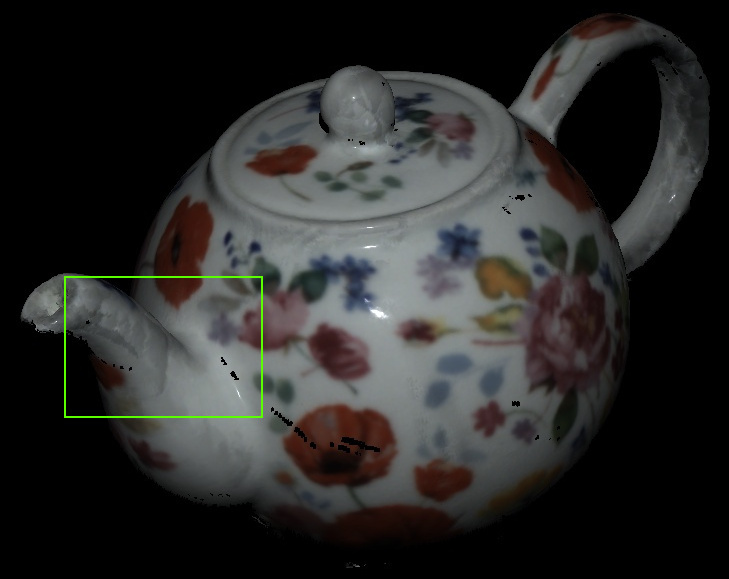} &
			\includegraphics[width=0.3\linewidth]{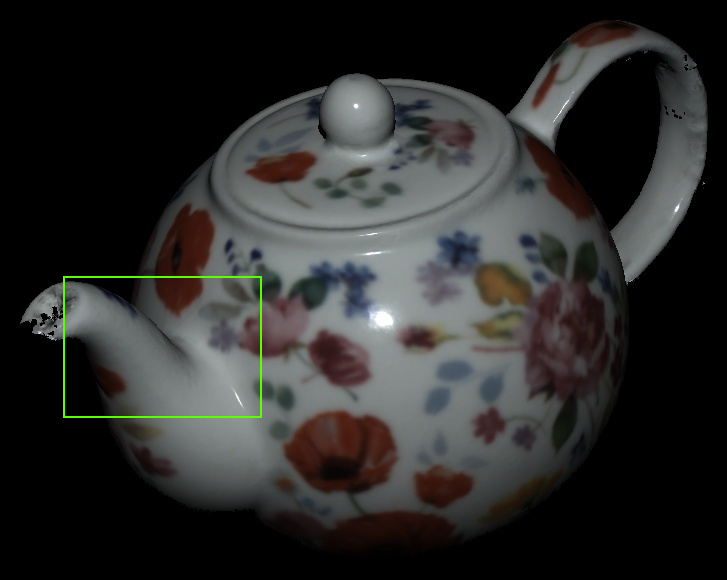} &
			\includegraphics[width=0.3\linewidth]{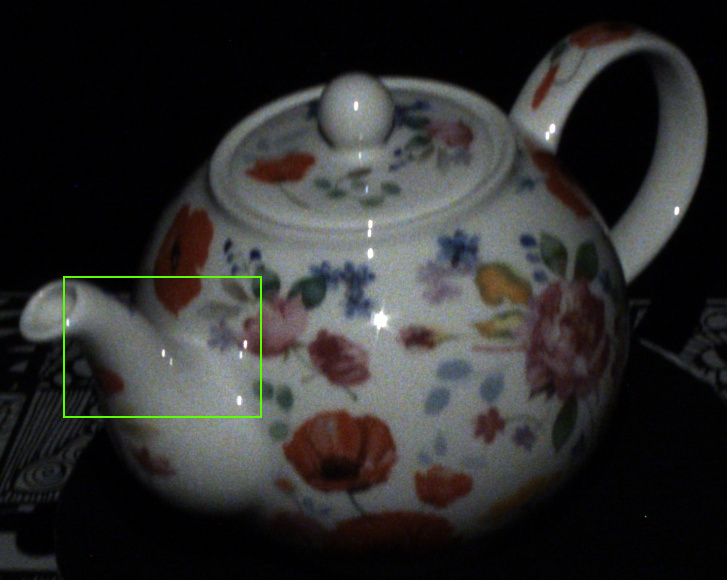} 
			\\
			\rotatebox{90}{$\qquad\quad$ Duck} &
			\includegraphics[width=0.3\linewidth]{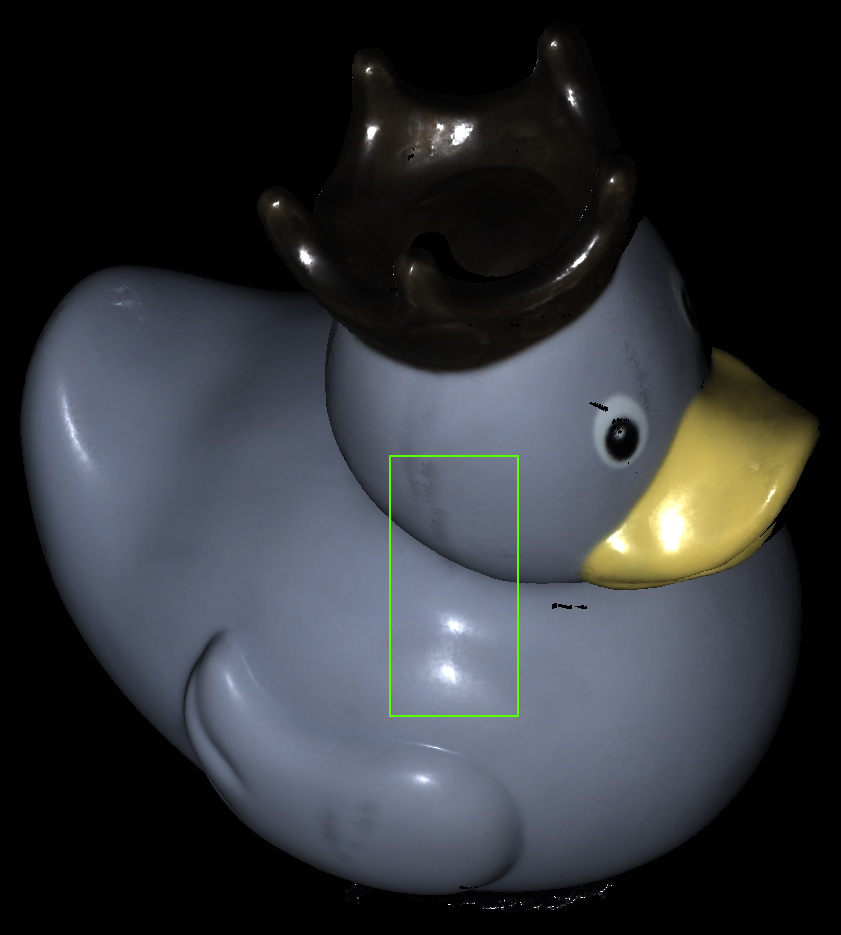} &
			\includegraphics[width=0.3\linewidth]{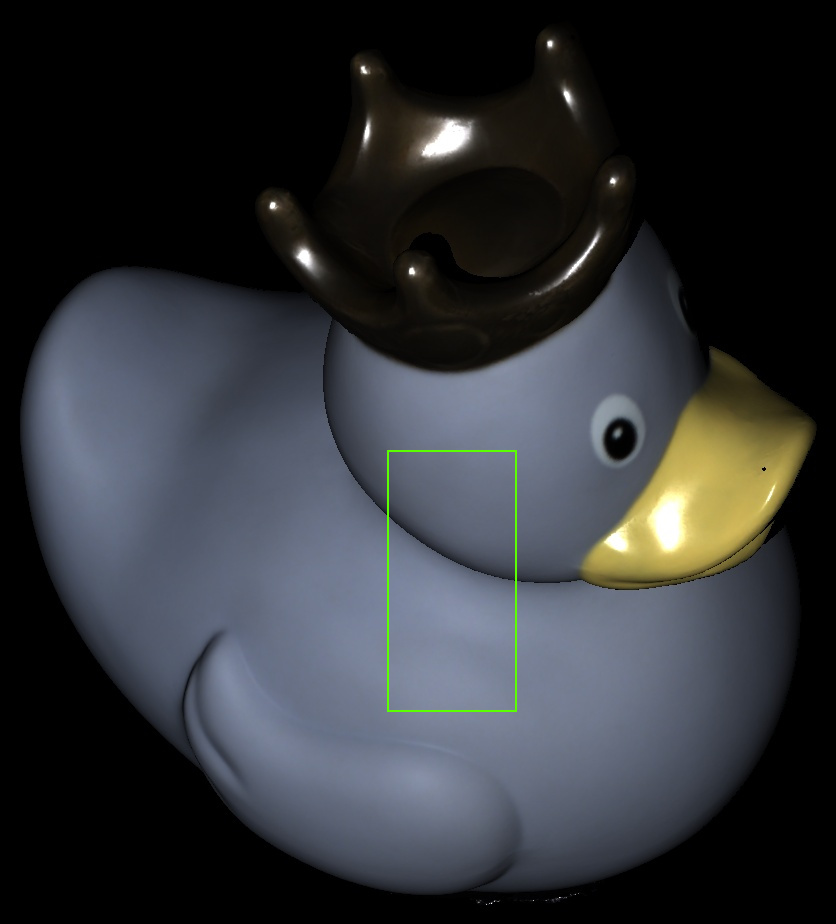} &
			\includegraphics[width=0.3\linewidth]{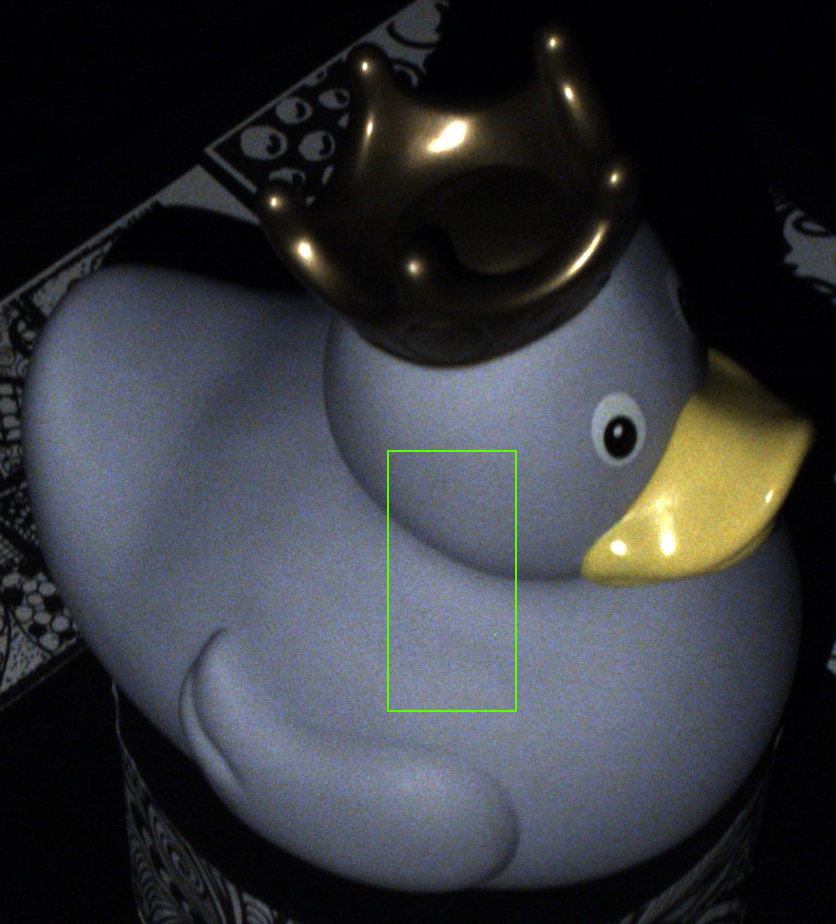} 
			\\
			\rotatebox{90}{$\qquad\qquad\;\;$ Gnome} &
			\includegraphics[width=0.3\linewidth]{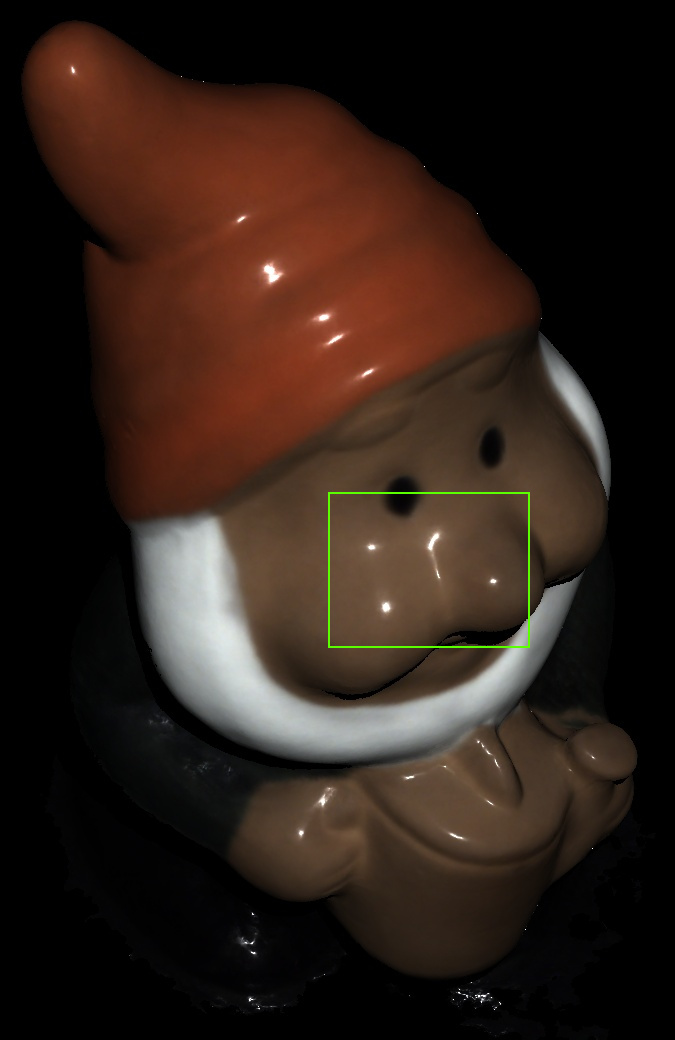} &
			\includegraphics[width=0.3\linewidth]{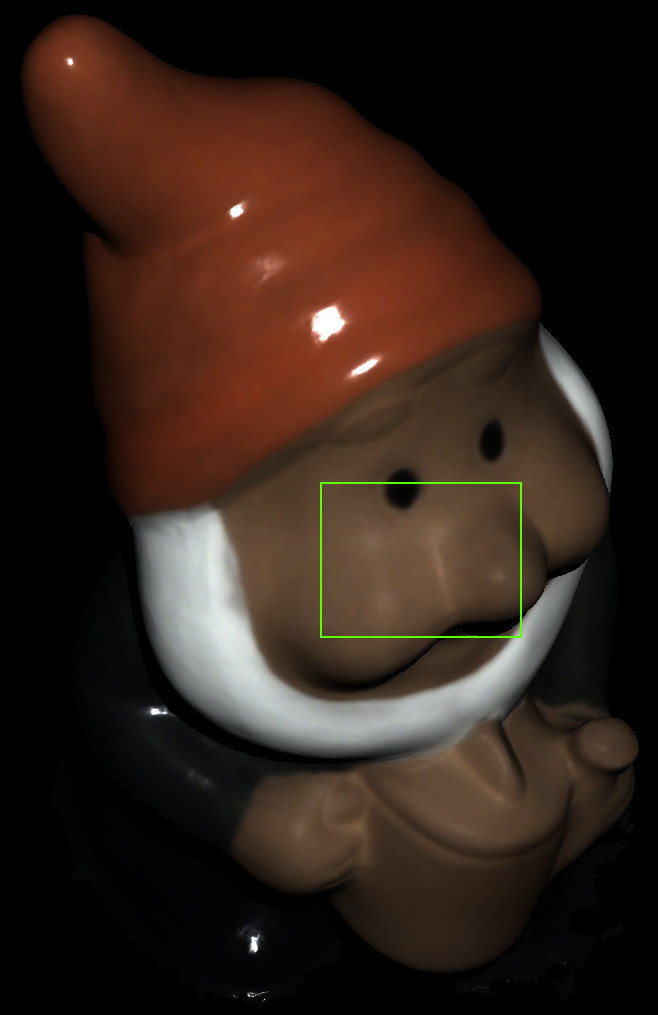} &
			\includegraphics[width=0.3\linewidth]{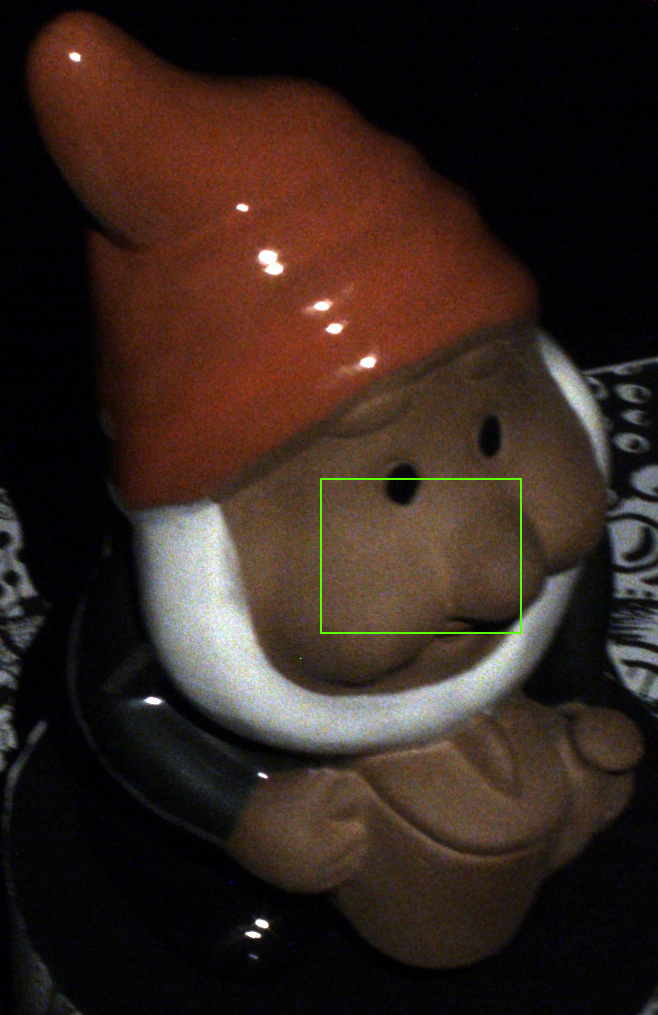} 
			\\
			\rotatebox{90}{$\qquad\;$ Globe} &
			\includegraphics[width=0.3\linewidth]{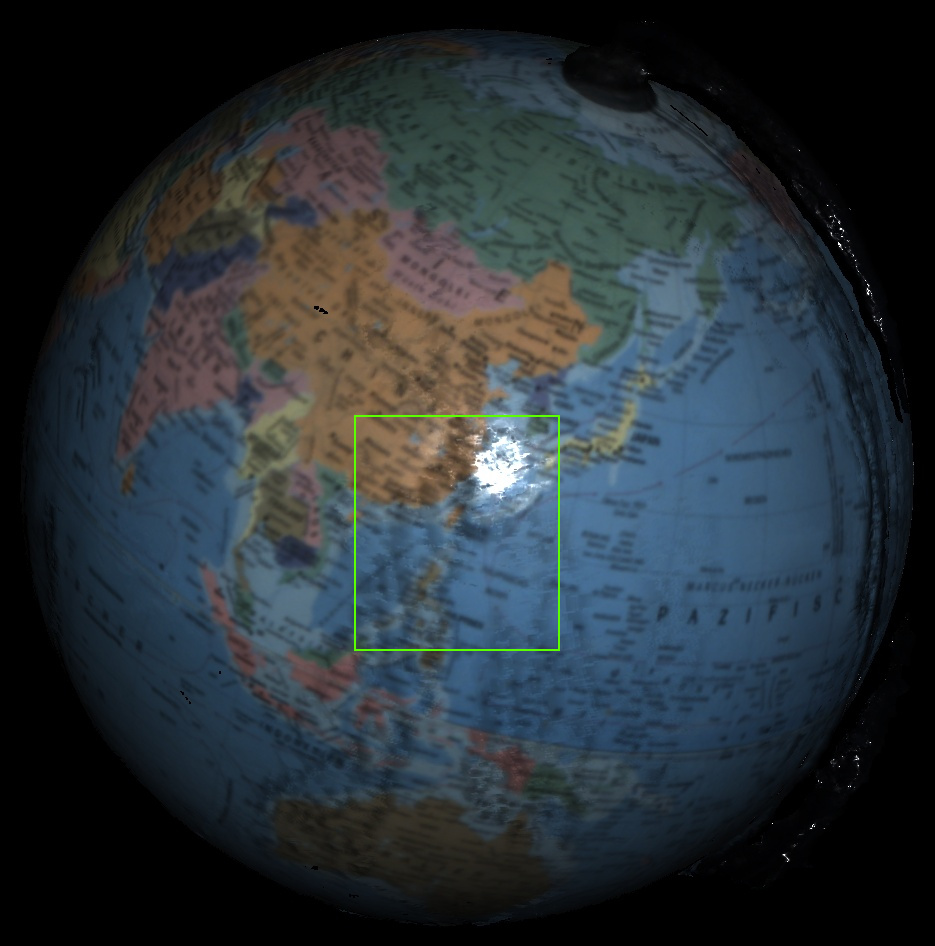} &
			\includegraphics[width=0.3\linewidth]{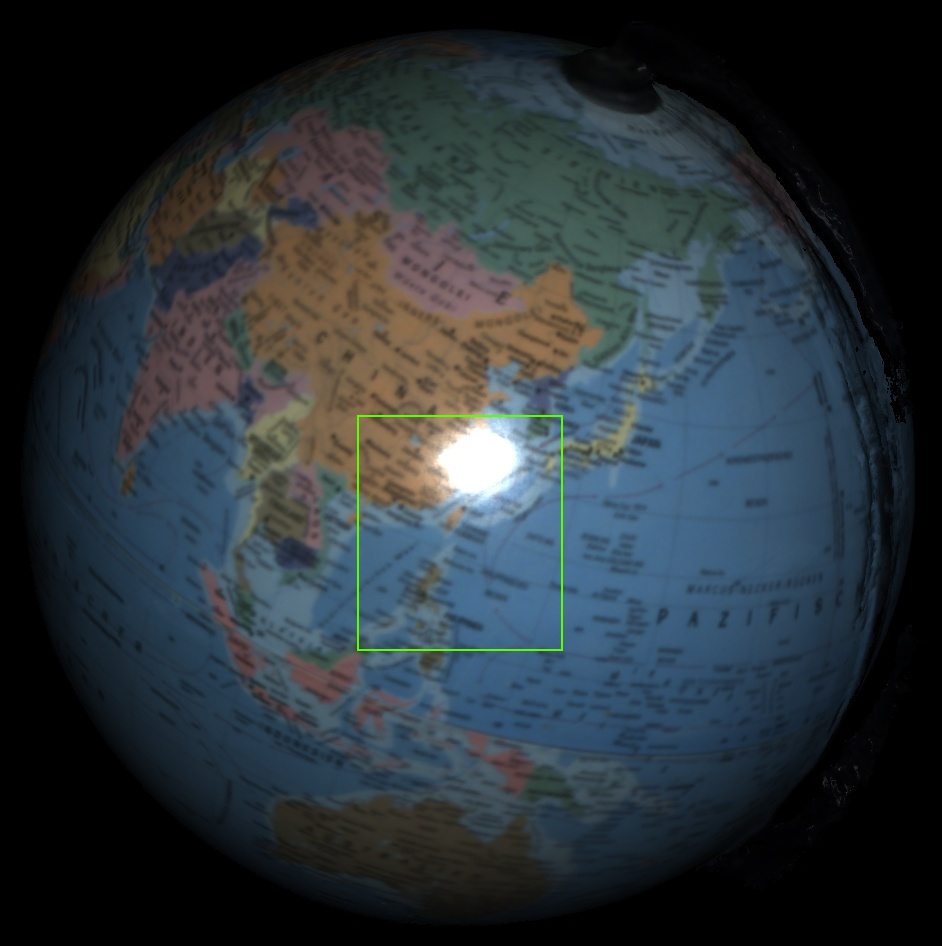} &
			\includegraphics[width=0.3\linewidth]{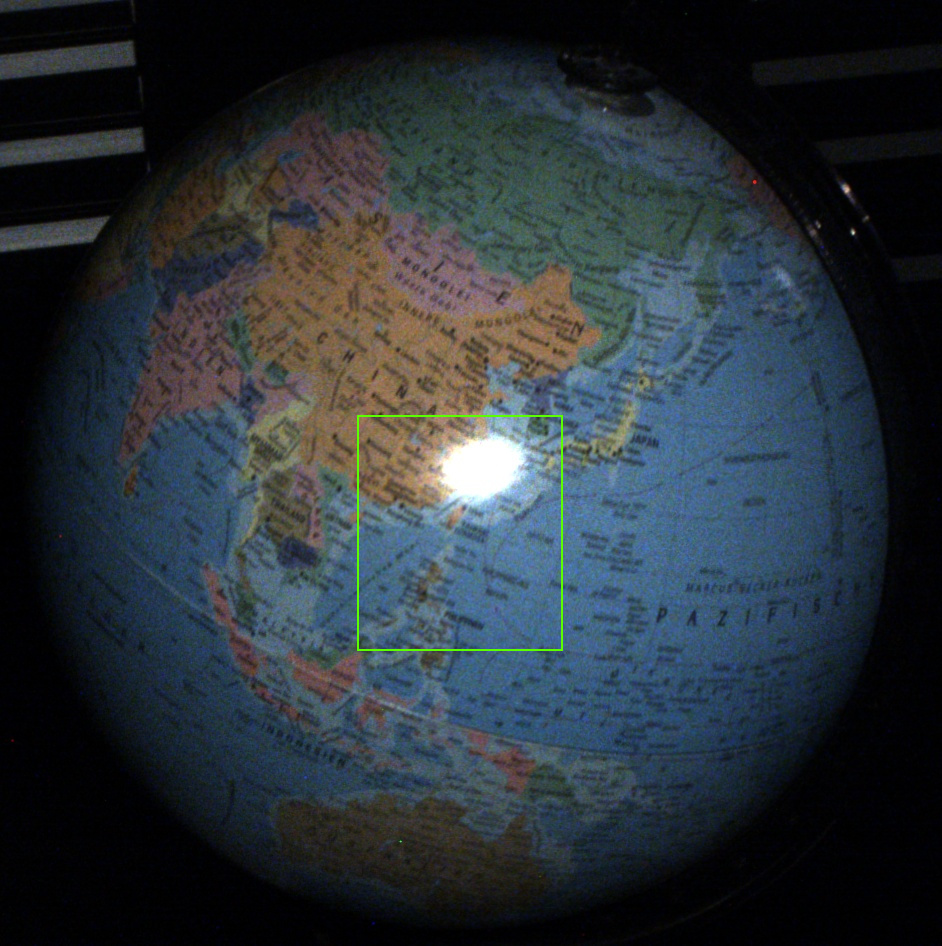} 
		\end{tabular}
		\end{centering}
		\addtolength{\tabcolsep}{5pt}
		\vspace{-2mm}
		\caption{Qualitative Results for Held-Out Test Views}
		\vspace{2mm}
		\label{tab:results_CVPR20_a}
	\end{subfigure}
	\begin{subfigure}{\linewidth}
		\centering
		\begin{tabular}{lccccc}
	\toprule
	{} &    Girl &  Teapot &    Duck &   Gnome &   Globe \\
	\midrule
	Schmitt et al. &  15.196 &  43.755 &  31.212 &  64.032 &  92.765 \\
	Proposed           &  12.281 &  18.128 &  10.366 &  17.483 &  10.510 \\
	\bottomrule
\end{tabular}
	
		\vspace{-1mm}
		\caption{Quantitative Results: Photometric Test Error}
		\label{tab:results_CVPR20_b}
	\end{subfigure}
	\caption{
		\textbf{Comparison to Schmitt \etal~\cite{Schmitt2020CVPR} (3D):} 
		For the same set of keyframes, we executed the proposed method and the independent 2.5D reconstructions as presented in the conference paper \cite{Schmitt2020CVPR}. 
		We show both models after volumetric fusion, qualitatively (a) and quantitatively (b) on held-out test views.
		While our method leads to realistic appearance reconstructions, the predictions of the method by Schmitt \etal~(independent optimizations) cannot resolve inconsistencies between keyframes and tend to over-estimate specular parameters (see highlighted regions of `Duck', `Gnome' and `Girl').
		Please see \figref{tab:results_CVPR20_c} for details.
	}
	\vspace{-1mm}
	\label{tab:results_CVPR20}
\end{figure}

\subsection{Comparisons to Existing Approaches}
\label{sec:results_baselines}

We compare our model with TSDF fusion \cite{Zeng2017CVPR}, the 2.5D optimization approach by Schmitt et al.~\cite{Schmitt2020CVPR} and the 3D reconstruction method from Nam et al.~\cite{Nam2018SIGGRAPH}. 
Hereby, we qualitatively evaluate our reconstructions in terms of geometric details, material modeling as well as overall appearance prediction.

\boldparagraph{Geometry Reconstruction} 
In \figref{fig:results_geometry} we compare the geometry reconstruction capabilities of our 3D model to the 2.5D method by Schmitt \etal~\cite{Schmitt2020CVPR} and na\"ive 3D TSDF fusion~\cite{Zeng2017CVPR} of the raw depth maps.
We show that both photometric approaches are able to recover fine geometric structures that are not present in the initial reconstruction.
Further, in contrast to Schmitt \etal~\cite{Schmitt2020CVPR}, the proposed method recovers the geometry for shiny and dark surfaces like the eyes of the `Owl'. 
Such materials are very challenging for photometric approaches since the signal-to-noise ratio of the diffuse component is low and the signal from specular highlights is very sparse.
Since our multi-view consistent optimization shares information between keyframes, it is often able to reconstruct such problematic regions.

\begin{figure}[t]
	\centering	
	\begin{subfigure}{\linewidth}
		\addtolength{\tabcolsep}{-4pt}
		\begin{tabular}{cccc}
			\rotatebox{90}{Schmitt \etal~\cite{Schmitt2020CVPR}} &
			\includegraphics[width=0.145\linewidth]{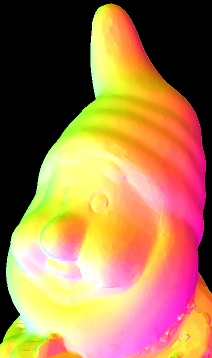}
			\includegraphics[width=0.145\linewidth]{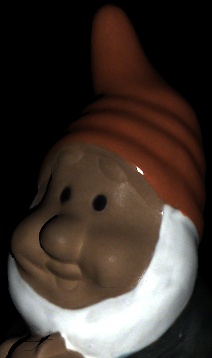} & 
			\includegraphics[width=0.145\linewidth]{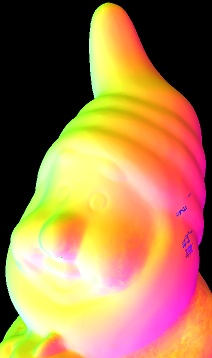}
			\includegraphics[width=0.145\linewidth]{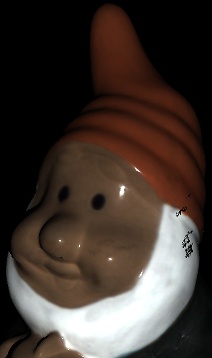} &
			\includegraphics[width=0.145\linewidth]{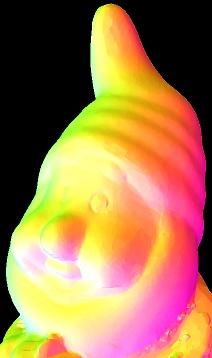}
			\includegraphics[width=0.145\linewidth]{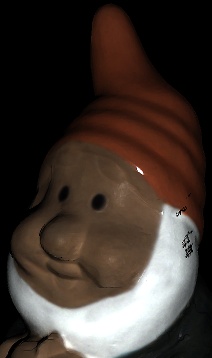}
			\\
			& 2.5D Keyframe & 3D Mesh & 3D + 2.5D Normals
		\end{tabular}
		\addtolength{\tabcolsep}{4pt}
	\end{subfigure}
	\caption{
		\textbf{Reconstruction Ambiguities for Schmitt \etal~\cite{Schmitt2020CVPR} (2.5D vs. 3D):} 
		Integration of independent 2.5D reconstructions, as presented in the conference paper \cite{Schmitt2020CVPR}, into a fused 3D mesh leads to incorrect appearance predictions, see \figref{tab:results_CVPR20}.
		This can be attributed to unresolved ambiguities between geometry and materials:
		Shown are the normal maps and rendered predictions for 
		(left) a single 2.5D reconstruction as presented in \cite{Schmitt2020CVPR}, 
		(middle) the fused 3D mesh after integration of independent 2.5D reconstructions as shown in \figref{tab:results_CVPR20_a}, left column and 
		(right) the fused 3D mesh (as in the middle) with only the normal map loaded from the 2.5D keyframe (from the left).
		We observe that the appearance of the 3D mesh (middle) shows artifacts. 
		But when using the noisier normal map from the 2.5D keyframe reconstruction, these artifacts are reduced noticeably (right). 
		That indicates that the model in Schmitt \etal~\cite{Schmitt2020CVPR} cannot resolve ambiguities in the normal and specular material estimation.
		It tends to over-estimate specular parameters (\eg~on the face of the gnome) and slightly perturbs the normal maps as compensation, resulting in good appearance.
	}
	\label{tab:results_CVPR20_c}
\end{figure}

\begin{figure*}
	\centering
	\addtolength{\tabcolsep}{-5.5pt}
	\begin{tabular}{ccccccccc}
		\rotatebox{90}{$\;\;\:$ Schmitt \etal~\cite{Schmitt2020CVPR}} &
		\rotatebox{90}{$\quad\;$ (Fused in 3D)} &
		\includegraphics[height=2.95cm]{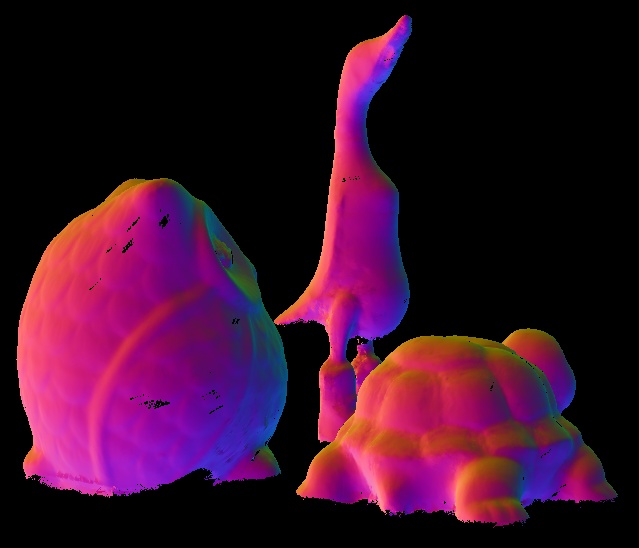} &
		\includegraphics[height=2.95cm]{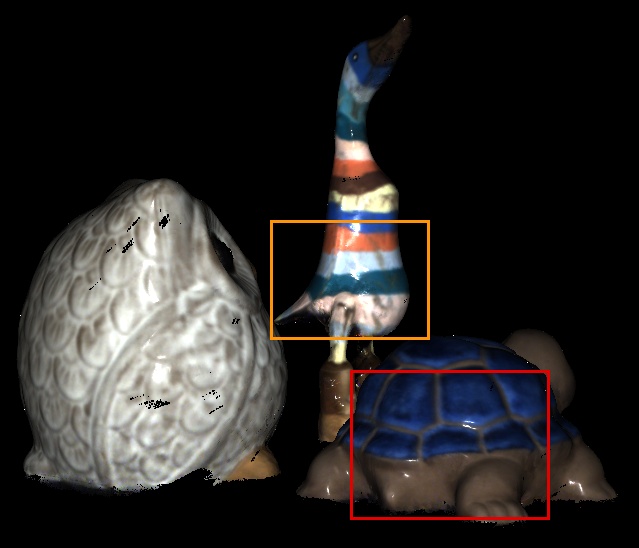} &
		\includegraphics[height=2.95cm]{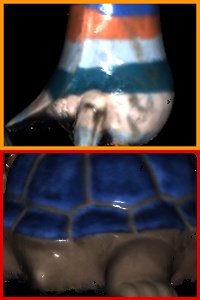} & $\:$ &
		\includegraphics[height=2.95cm]{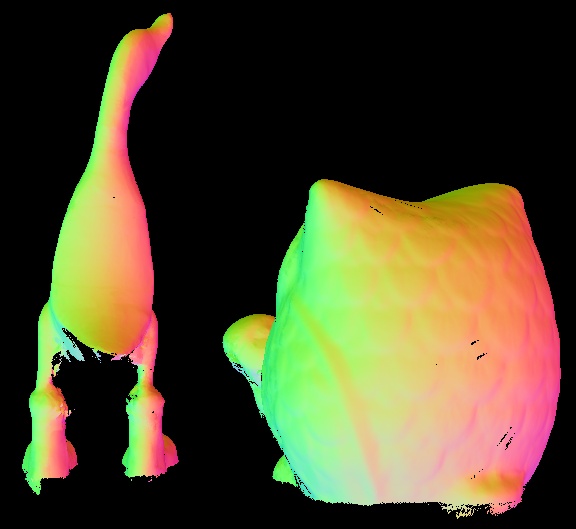} &
		\includegraphics[height=2.95cm]{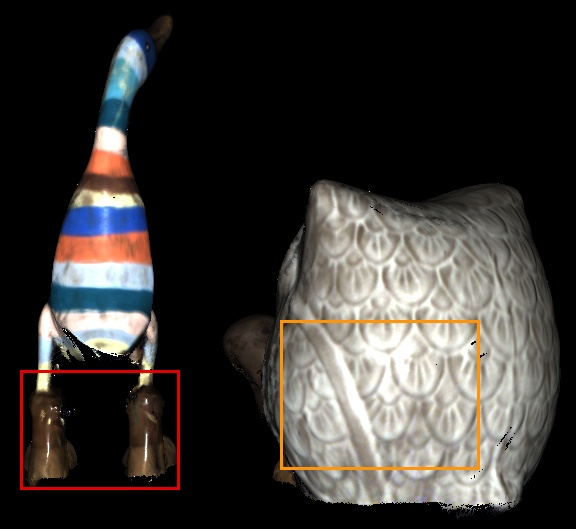} &
		\includegraphics[height=2.95cm]{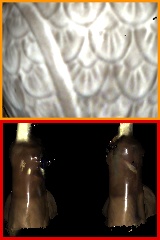} 
		\\
		\rotatebox{90}{$\quad\;$ Nam \etal~\cite{Nam2018SIGGRAPH}} &
		\rotatebox{90}{$\qquad\quad\;$ (3D)} &
		\includegraphics[height=2.95cm]{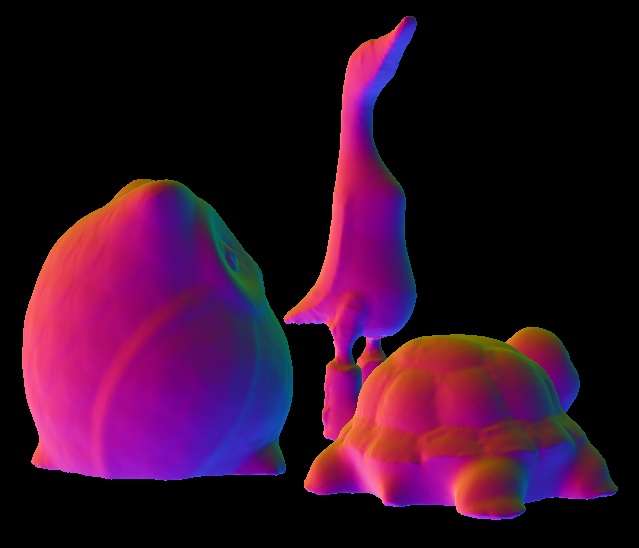} &
		\includegraphics[height=2.95cm]{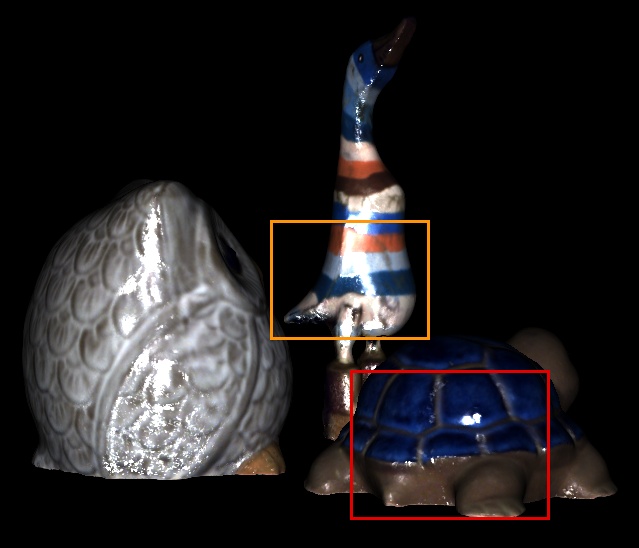} &
		\includegraphics[height=2.95cm]{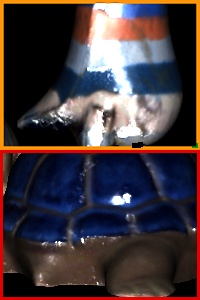}  & $\:$ &
		\includegraphics[height=2.95cm]{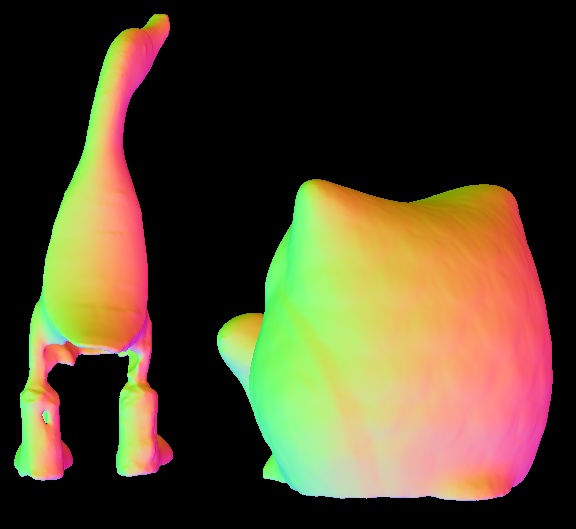} &
		\includegraphics[height=2.95cm]{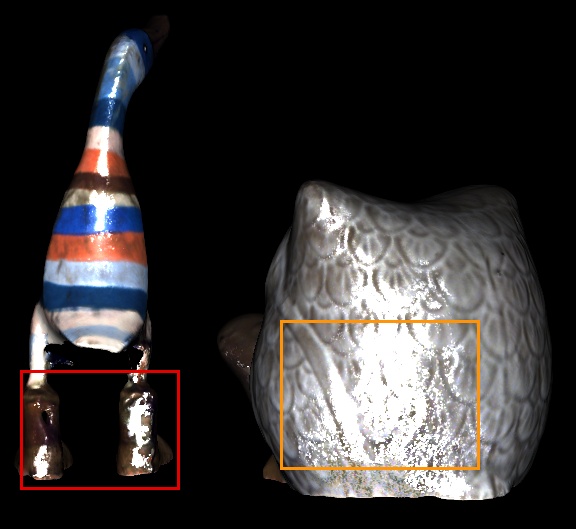} &
		\includegraphics[height=2.95cm]{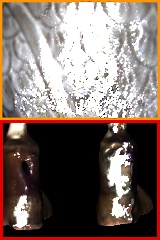} 
		\\
		\rotatebox{90}{$\qquad\;$ Proposed} &
		\rotatebox{90}{$\qquad\quad\;$ (3D)} &
		\includegraphics[height=2.95cm]{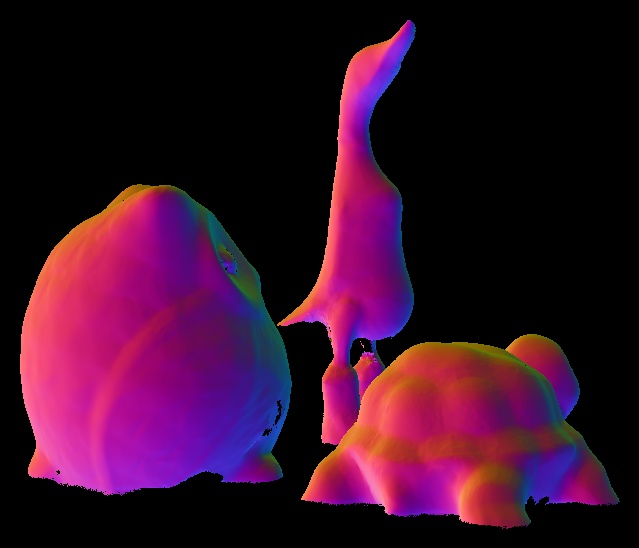} &
		\includegraphics[height=2.95cm]{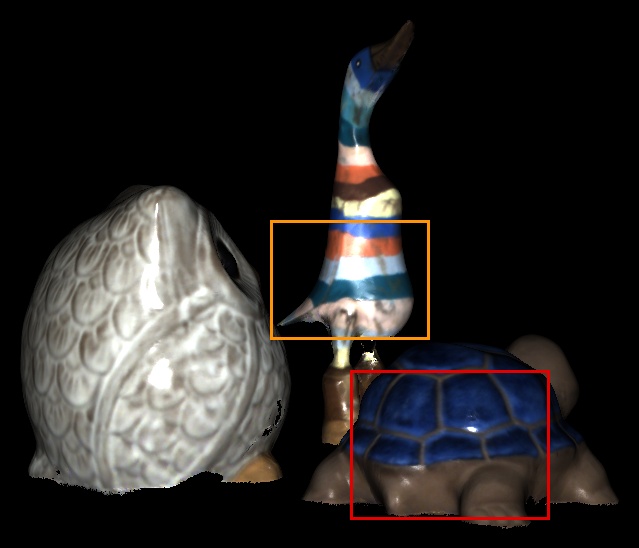} &
		\includegraphics[height=2.95cm]{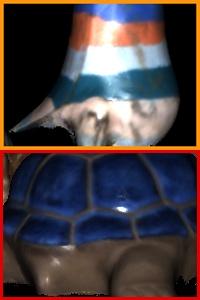}  & $\:$ &
		\includegraphics[height=2.95cm]{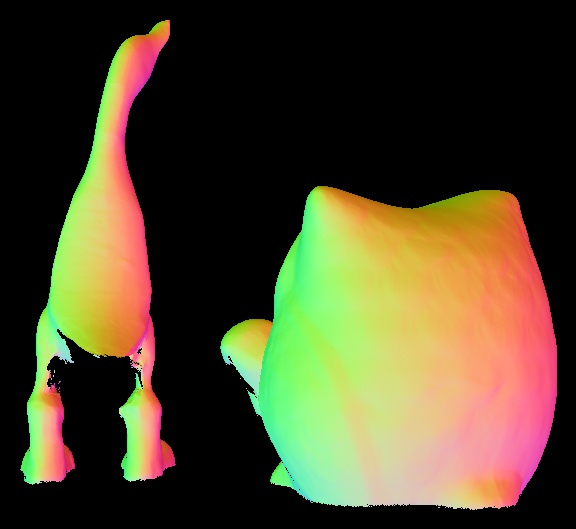} &
		\includegraphics[height=2.95cm]{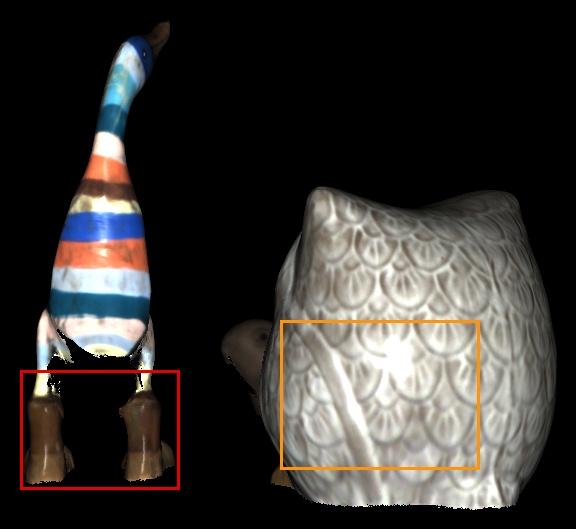} &
		\includegraphics[height=2.95cm]{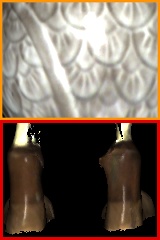} 
		\\
		& & 3D Normal Map & Rendered Prediction & Zoom-In & & 3D Normal Map & Rendered Prediction & Zoom-In 
	\end{tabular}
	\addtolength{\tabcolsep}{5.5pt}
	\caption{
		\textbf{Qualitative Comparison to Baselines (3D).} 
		Simple fusion of the 2.5D reconstruction results of Schmitt \etal~\cite{Schmitt2020CVPR} results in artifacts in the geometry and appearance.
		While the model from Nam \etal~\cite{Nam2018SIGGRAPH} recovers detailed normal maps, the material reconstruction fails to capture the appearance of the object which leads to high-frequency artifacts in the predictions.
		Our model estimates normals that are on par with the results of Nam \etal~\cite{Nam2018SIGGRAPH} and predicts consistent appearance with realistic specular reflections.
	}
	\label{fig:results_baselines}
\end{figure*}

\boldparagraph{Comparison to Schmitt et al.~\cite{Schmitt2020CVPR}}
We present a qualitative and quantitative comparison to \cite{Schmitt2020CVPR}, the conference paper that we extend in this paper.
Since the method jointly reconstructs pose, geometry and materials for local 2.5D scene representations, we execute it independently on all keyframes in $K$ and demonstrate that integration of the results to a fused, global 3D mesh is insufficient:
The 2.5D parameter estimates are inconsistent, causing patching artifacts and wrong appearance in the predictions of the integrated 3D models, see \figref{tab:results_CVPR20_a} and \figref{fig:results_baselines}.
The integration in 3D also reveals ambiguities in the estimation of geometric and photometric parameters. 
Our case study in \figref{tab:results_CVPR20_c} shows that the 2.5D model is indeed prone to over-estimate specular reflection: 
since the appearance is highly sensitive to angular changes in the normals, very small deviations of the normals are sufficient to strongly alter the appearance \eg~by removing specular highlights from the predictions. 
Therefore, the model is able to ``cheat'' by tilting normals away instead of decreasing the glossiness of the material.
In contrast, our method resolves such ambiguities by incorporating information from neighboring keyframes and encouraging multi-view consistency.
Specifically, regularization against aggregated neighboring parameter maps prevents a bias in the normals and leads to better material estimates.
\figref{tab:results_CVPR20_b} confirms significantly lower reconstruction errors for our model on held-out test views.

\boldparagraph{Comparison to Nam et al.~\cite{Nam2018SIGGRAPH}}
We compare our method with that of Nam et al.~\cite{Nam2018SIGGRAPH} on a scene with multiple objects.
As shown in \figref{fig:results_baselines}, their method reconstructs 3D appearance components such as normals and diffuse albedo properly but does not manage to recover the specular reflections of the given scene well.
Since they use base materials for reflectance modeling, this indicates that the clustering into surface regions with distinct materials fails, potentially due to an erroneous estimate of the number of base materials.
This leads to non-smooth predictions even for regions with similar appearance and results in uneven specular highlights and high-frequency artifacts.
In contrast, our method reconstructs materials and specular highlights well because our pixel-wise material representation does not involve a model selection step.

\subsection{Reconstruction Results} 
\label{sec:results_experiments}
We show results of our method on captured objects in \figref{fig:results_all_objects1} and the supplement and demonstrate the capabilities of our method to reconstruct accurate geometry and materials for a variety of real objects, scenes and materials.
Please see videos of our reconstructions here:
\href{https://sites.google.com/view/material-fusion/}{https://sites.google.com/view/material-fusion/}
And results on synthetic data can be found in the supplement.

\boldparagraph{Towards Scalable Scene Reconstructions} 
\label{sec:results_room-scale}
We demonstrate the scalability of the proposed approach in \figref{fig:results_room-scale}.
As shown, our method reconstructs scenes on the scale of several meters at a resolution of $\leq2$mm and recovers accurate appearance and geometry, leading to realistic renderings of novel viewpoints and illumination.
But the `Office' scene shows 2 limitations of our model: Since we do not recover missing geometry or fill holes but rely on the completeness of the input geometry, artifacts on the `Teapot' and the `Mug' can not be resolved.
Additionally, we do not model global illumination effects leading to small artifacts in the diffuse albedo map.

\subsection{Limitations}
\label{sec:results_limitations}
\begin{figure*}
	\centering
	\addtolength{\tabcolsep}{-5pt}
	\begin{tabular}{ccccccc}
		\centering
		\rotatebox{90}{$\quad\;\;\;$ 3 Objects} &
		\includegraphics[height=2.8cm]{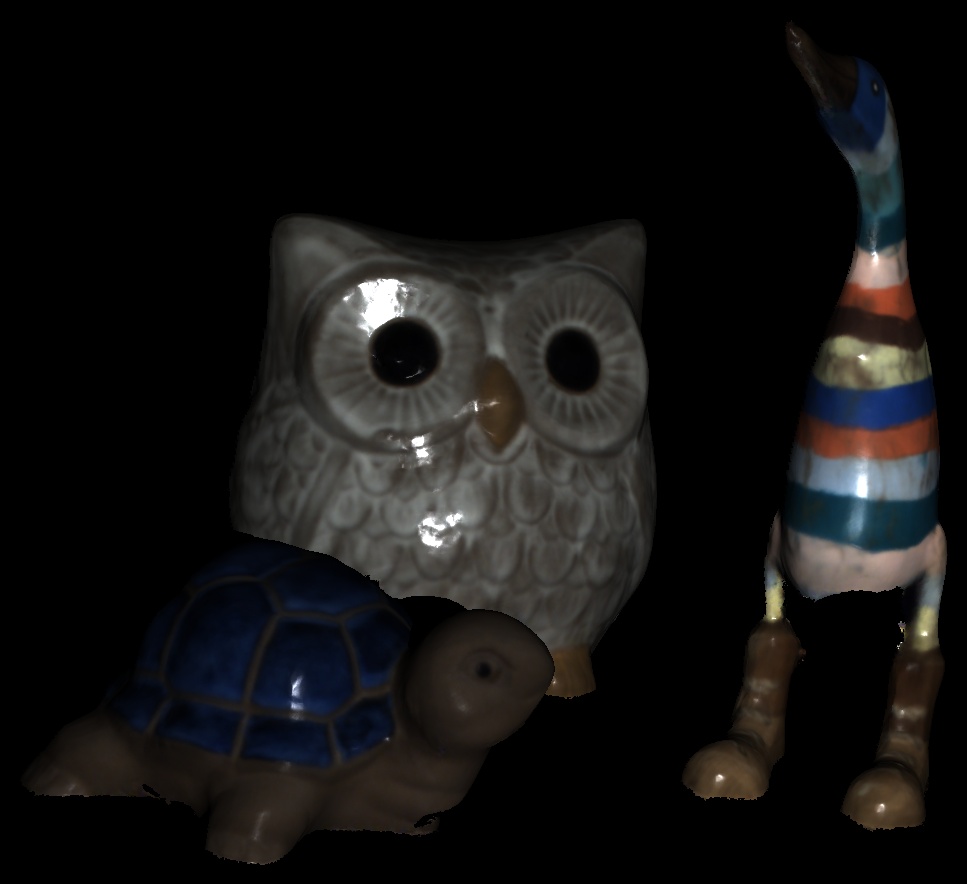} 
		\includegraphics[height=2.8cm]{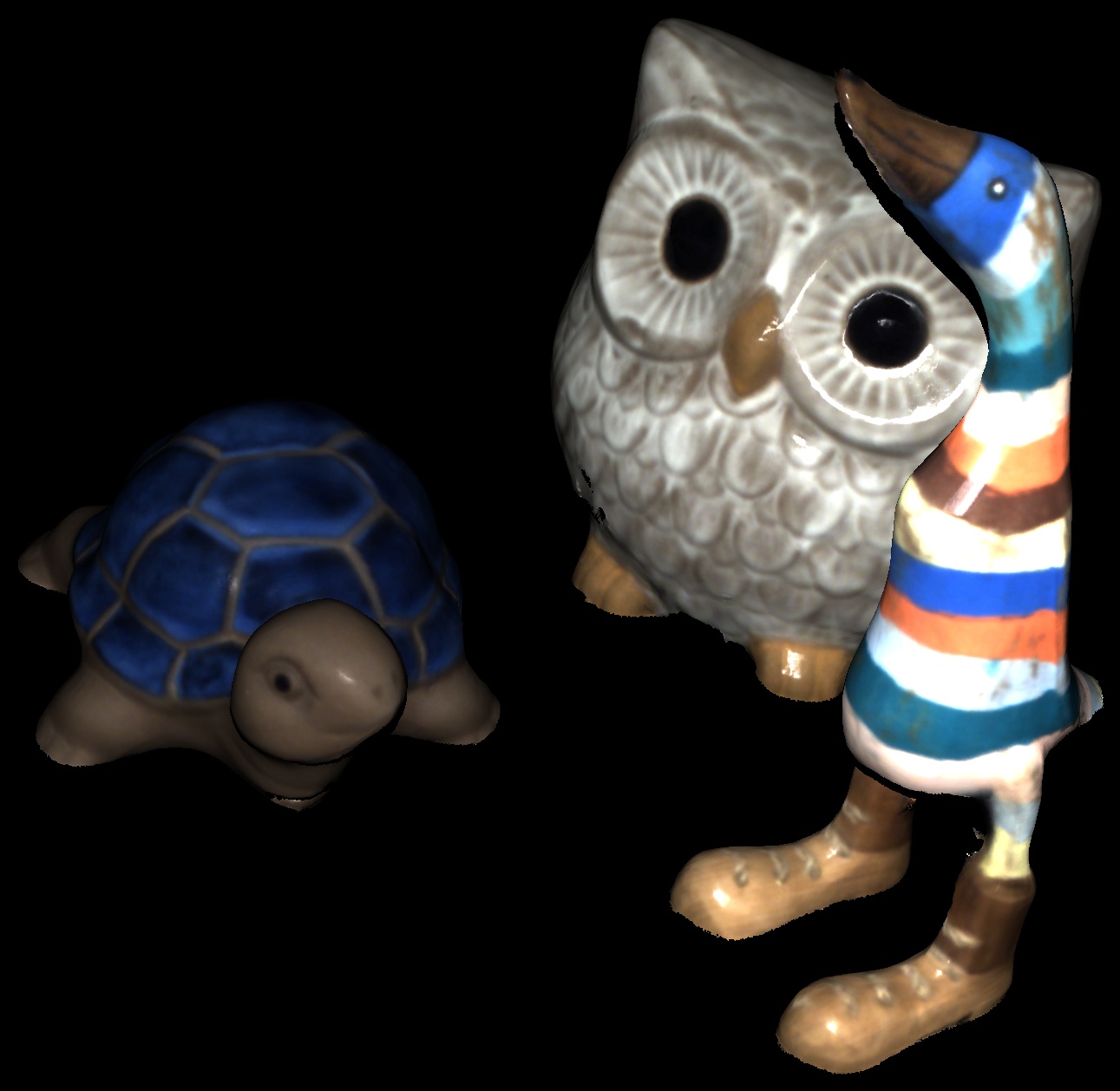} &
		\includegraphics[height=2.8cm]{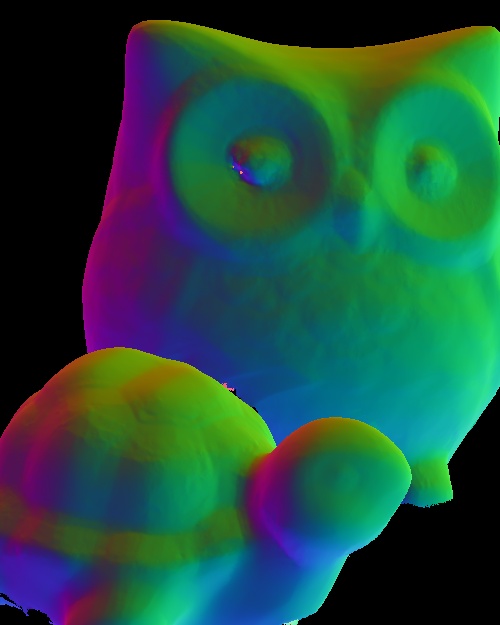} &
		\includegraphics[height=2.8cm]{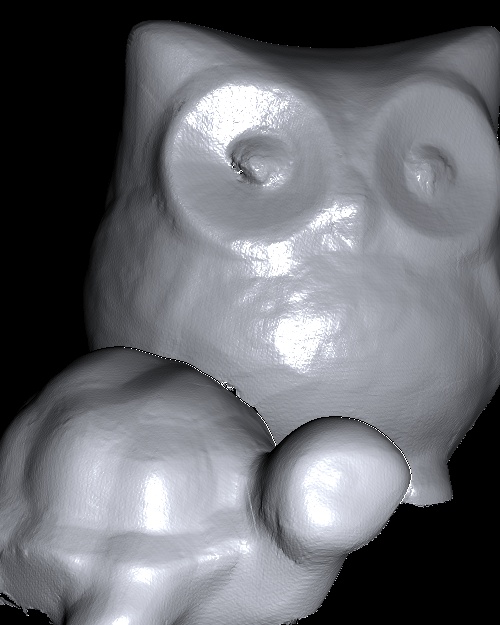} &
		\includegraphics[height=2.8cm]{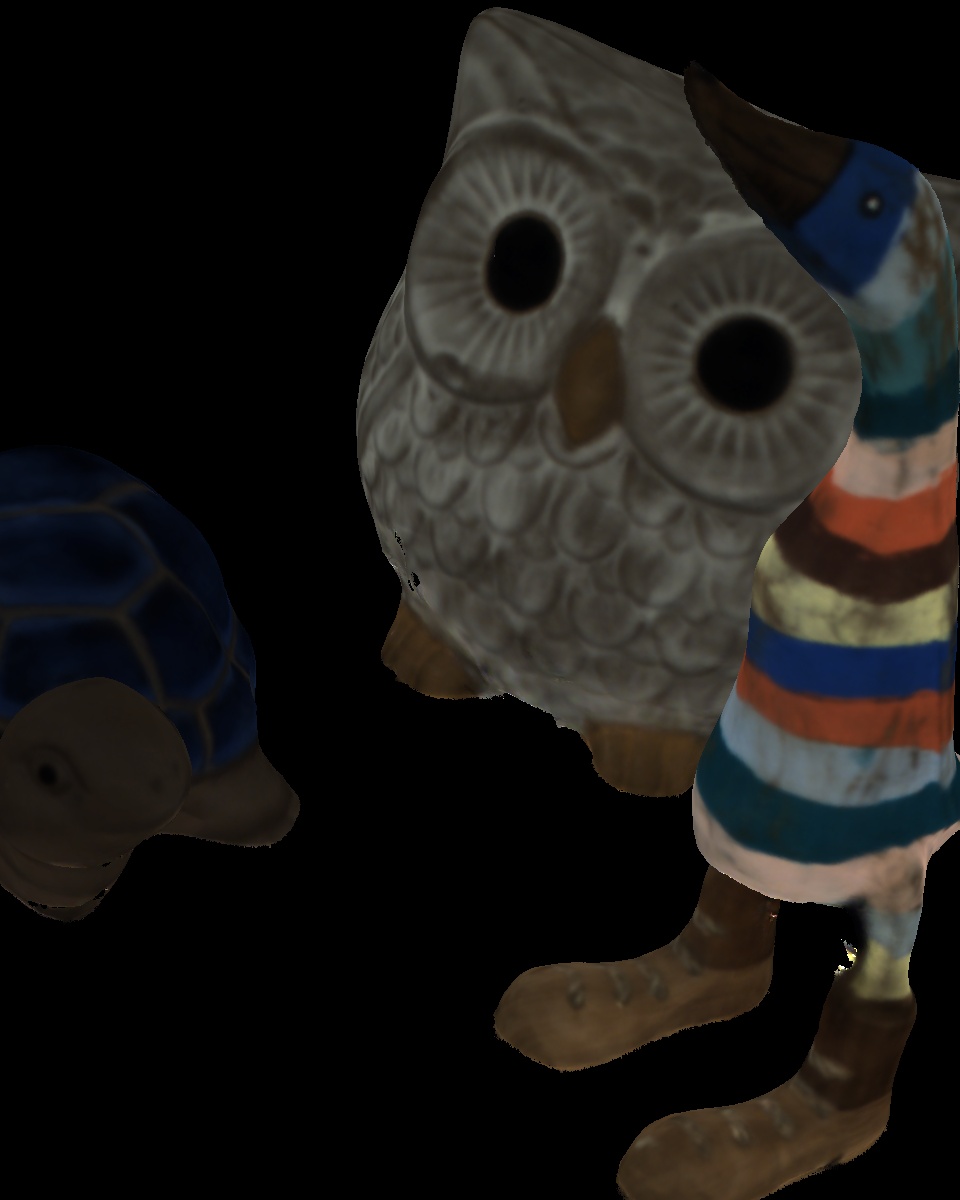} &
		\includegraphics[height=2.8cm]{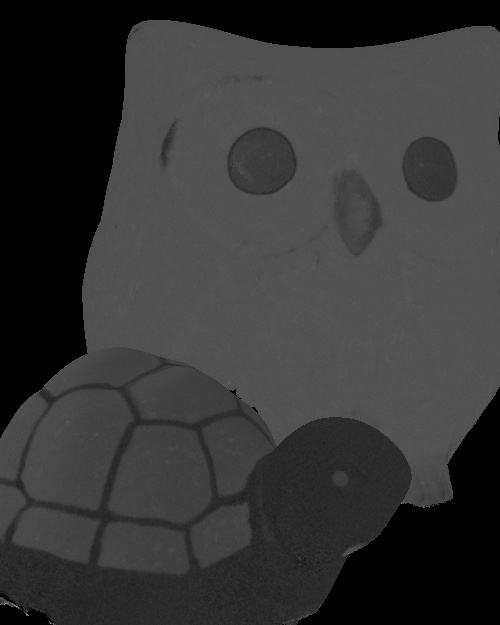} &
		\includegraphics[height=2.8cm]{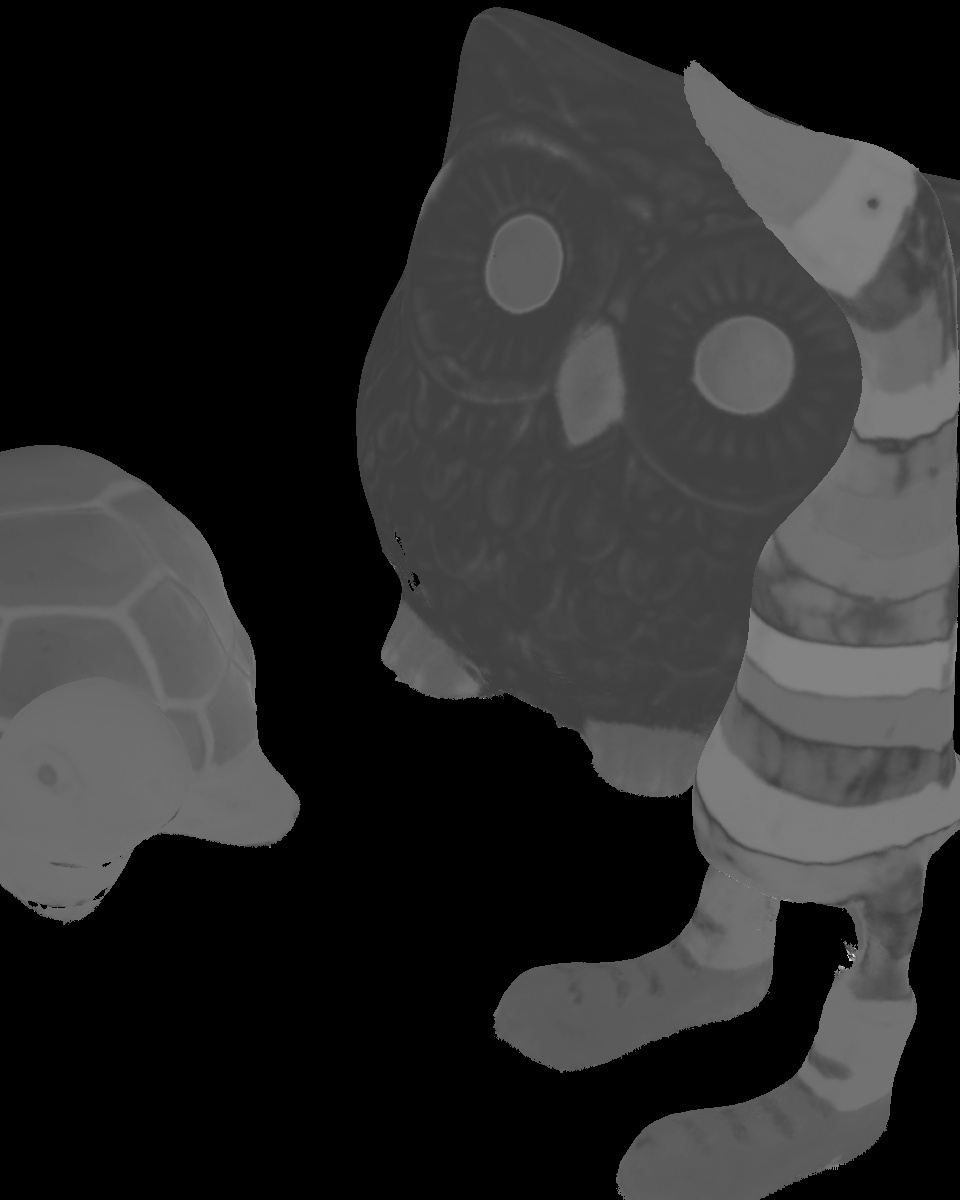} 
		\\
		\rotatebox{90}{$\qquad\quad$ Sofa} &
		\includegraphics[height=2.8cm]{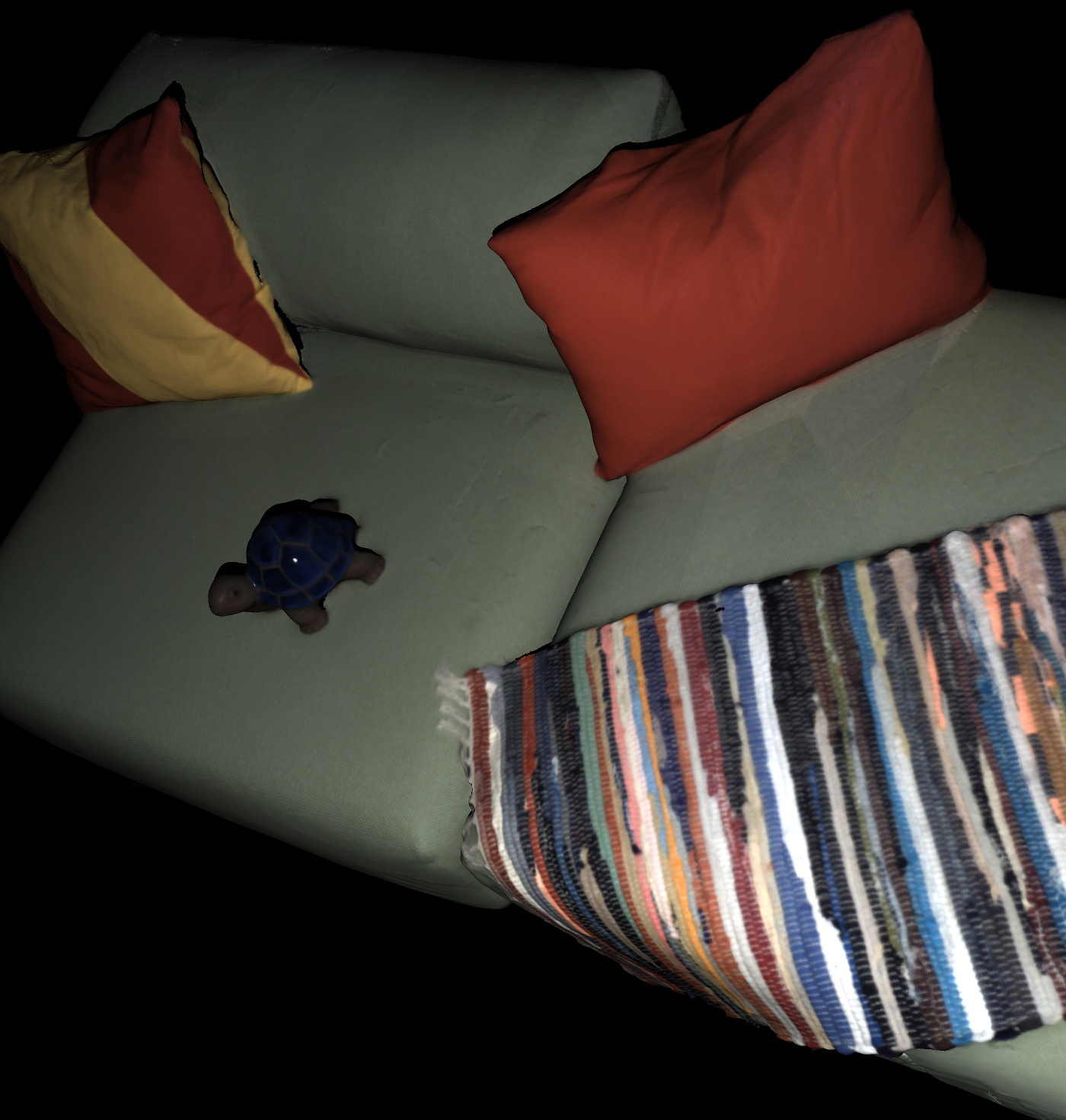} 
		\includegraphics[height=2.8cm]{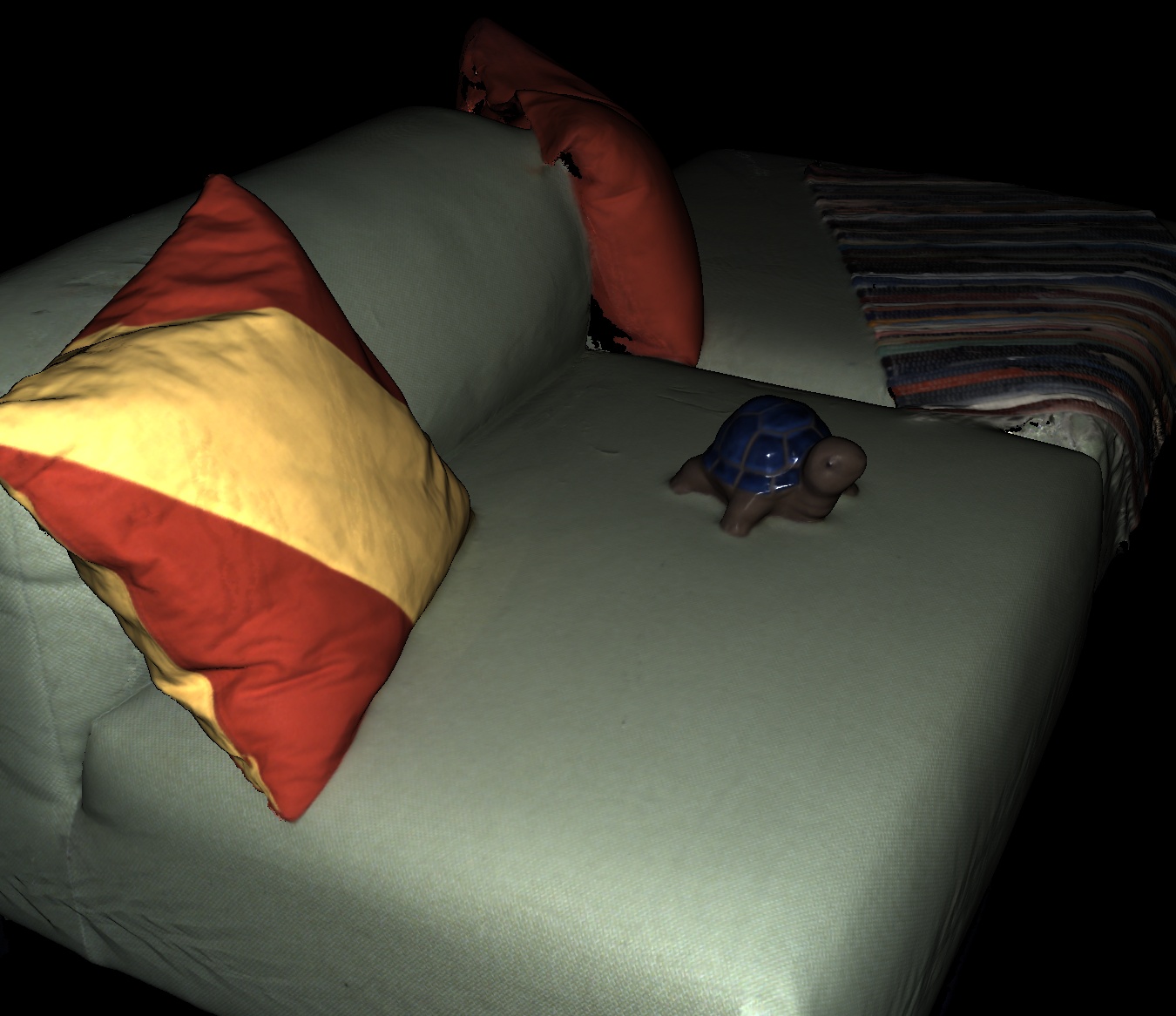} &
		\includegraphics[height=2.8cm]{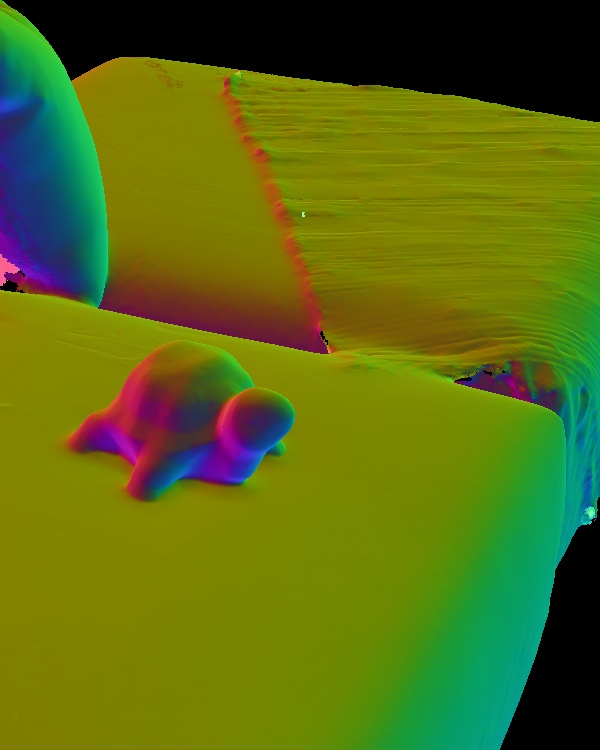} &
		\includegraphics[height=2.8cm]{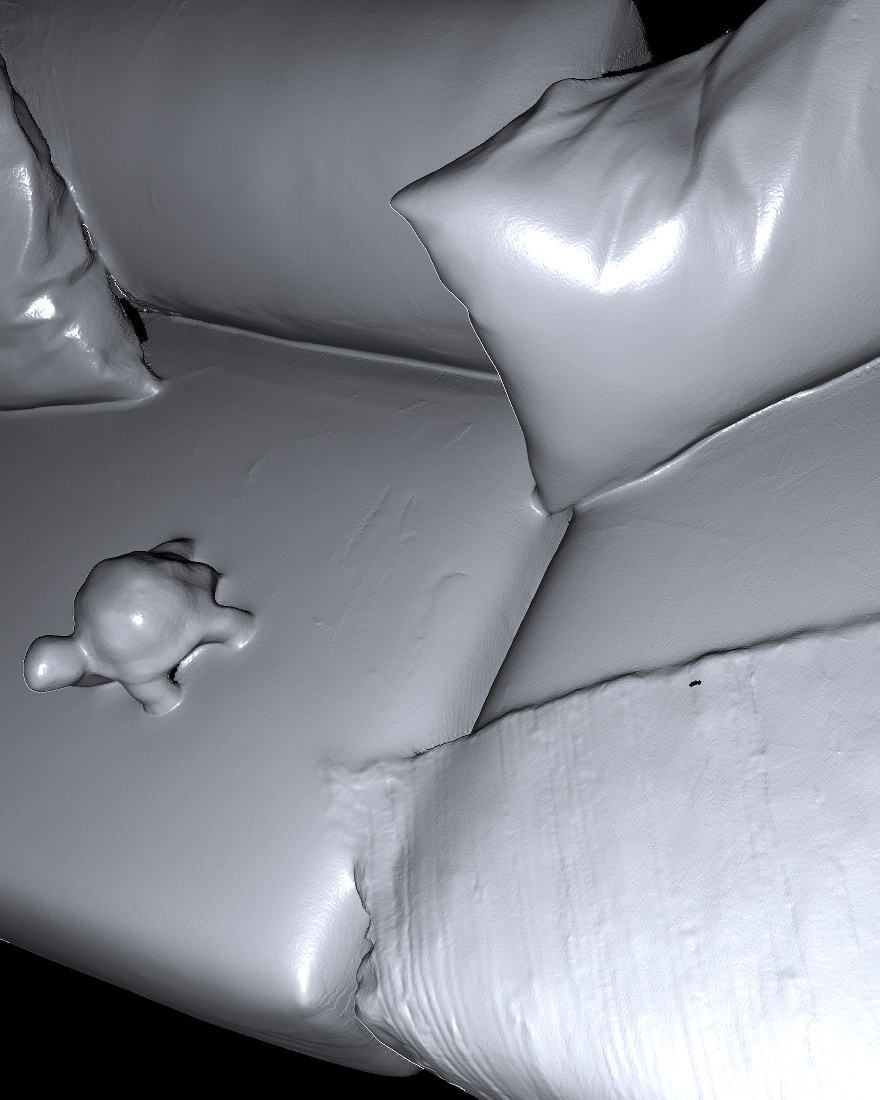} &
		\includegraphics[height=2.8cm]{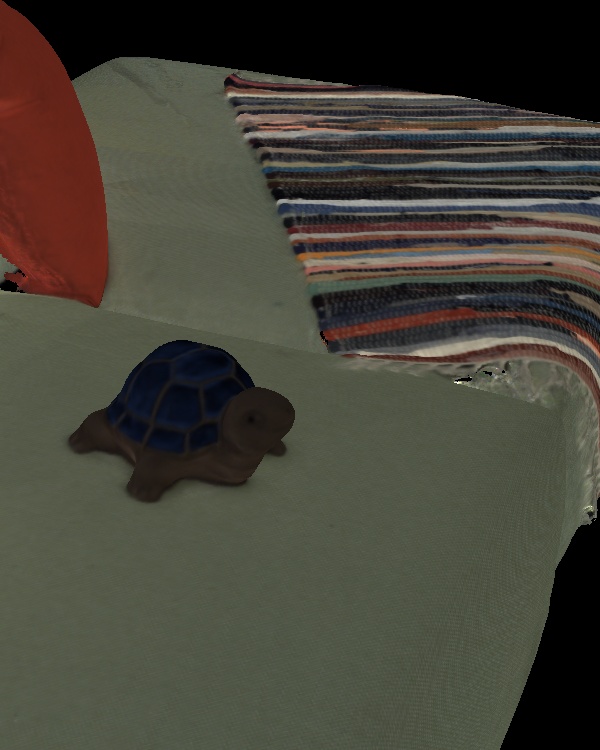} &
		\includegraphics[height=2.8cm]{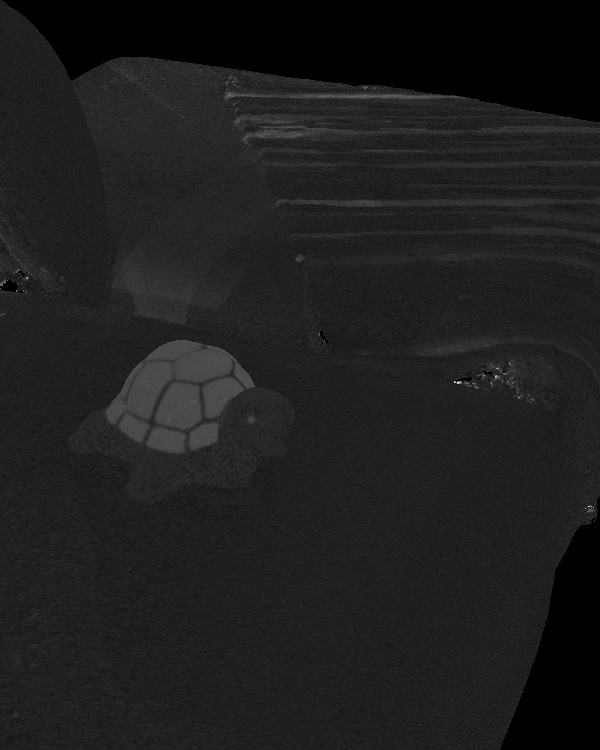} &
		\includegraphics[height=2.8cm]{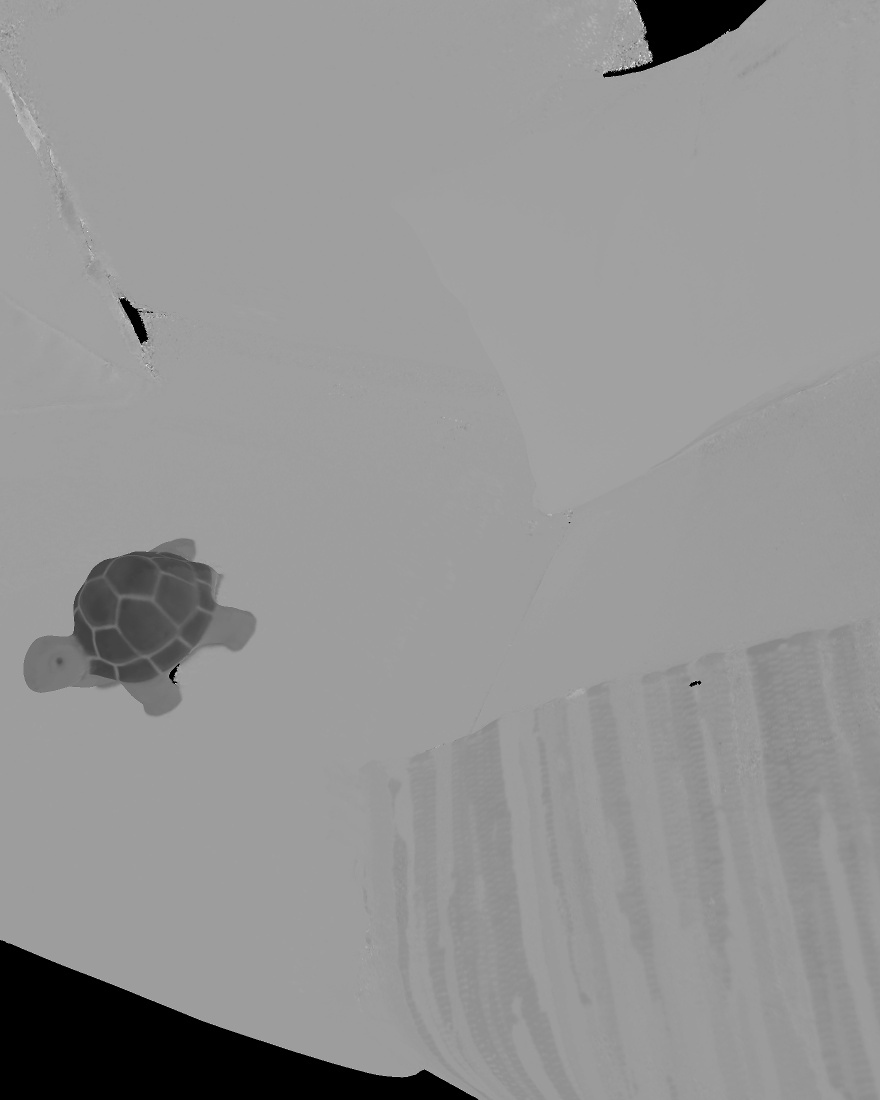} 
		\\
		\rotatebox{90}{$\qquad\;\;$ Office} &
		\includegraphics[height=2.8cm]{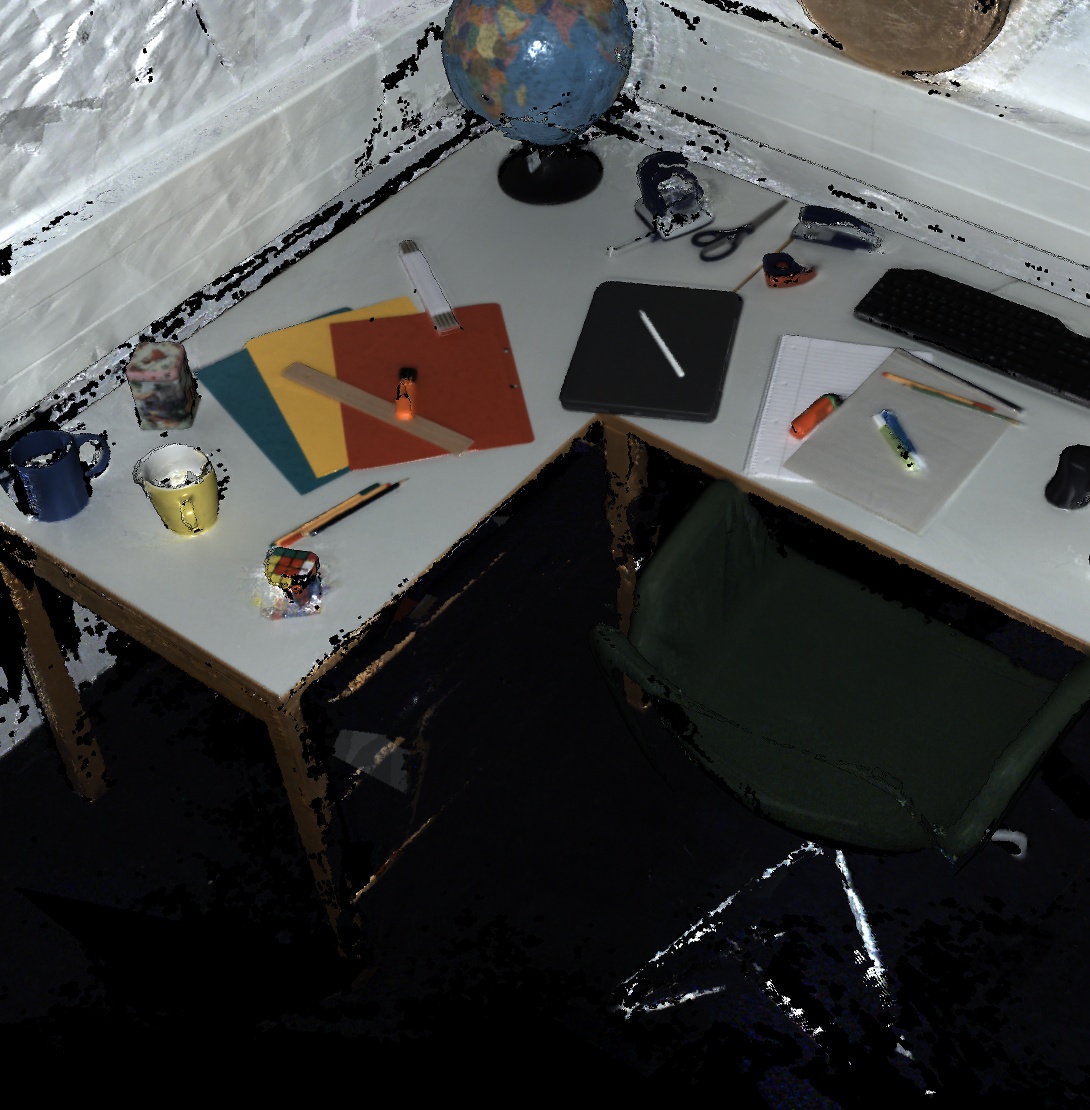} 
		\includegraphics[height=2.8cm]{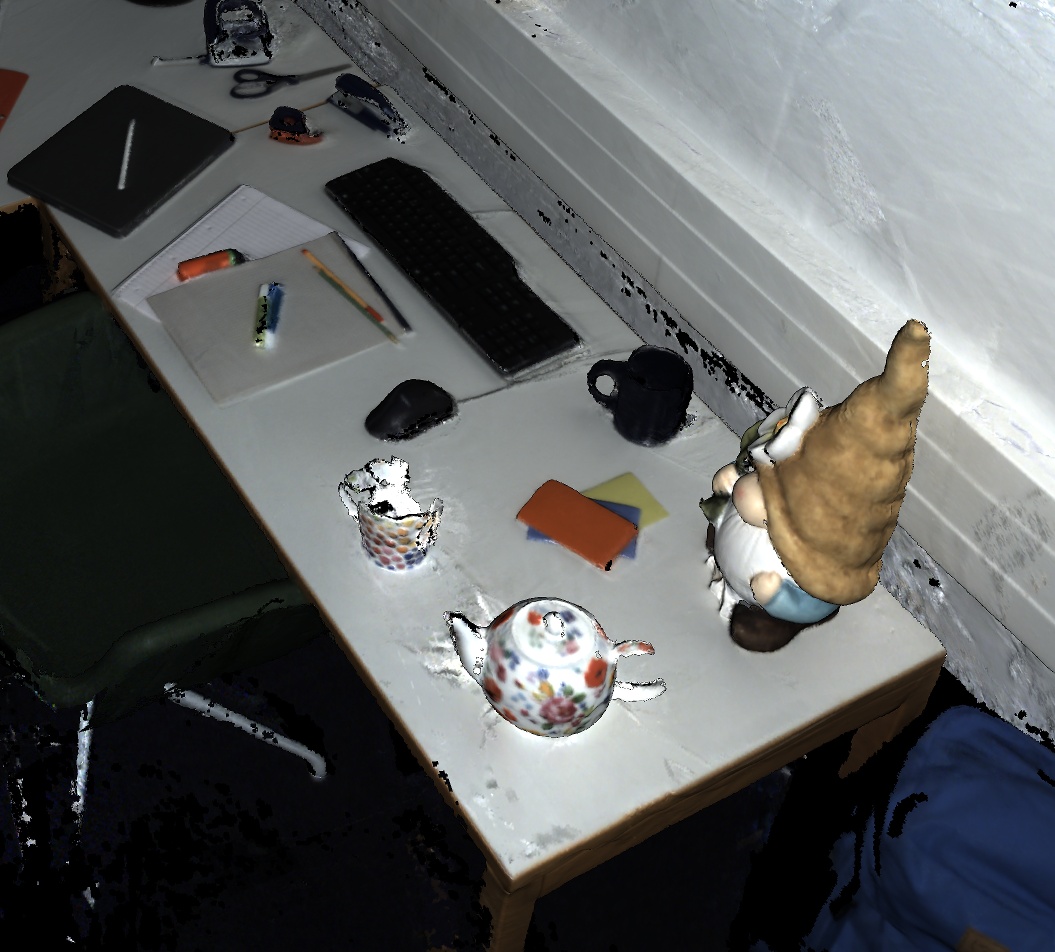} &
		\includegraphics[height=2.8cm]{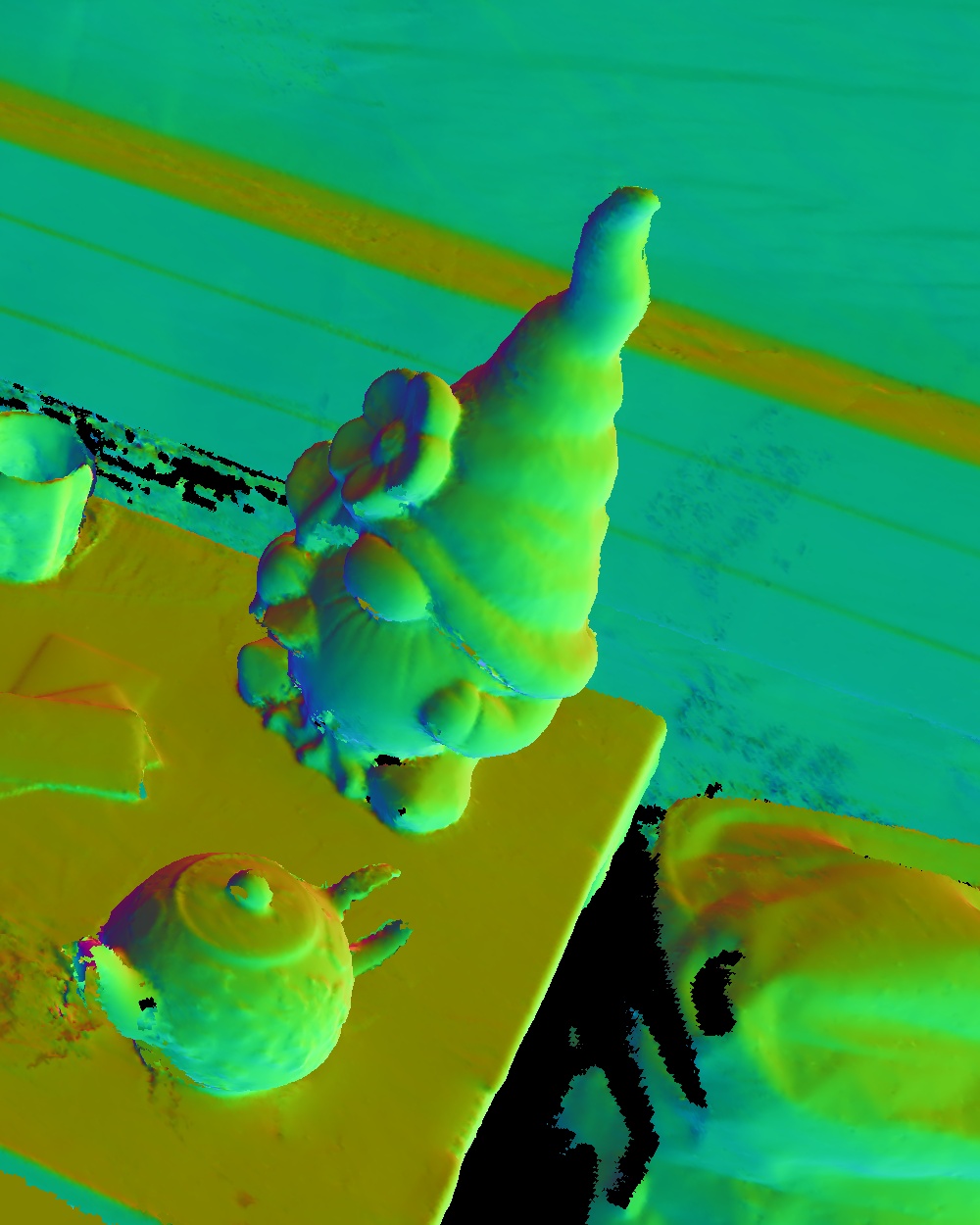} &
		\includegraphics[height=2.8cm]{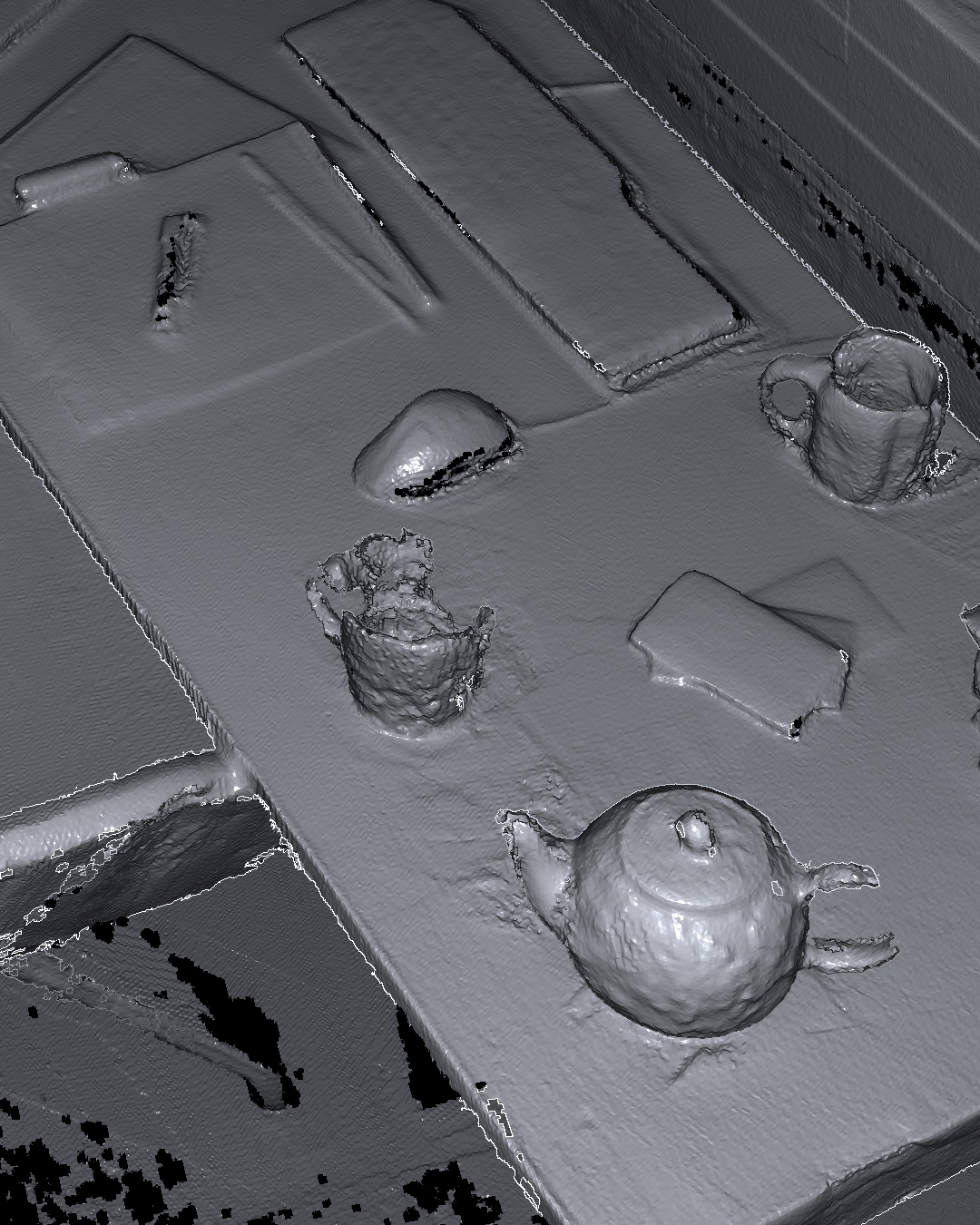} &
		\includegraphics[height=2.8cm]{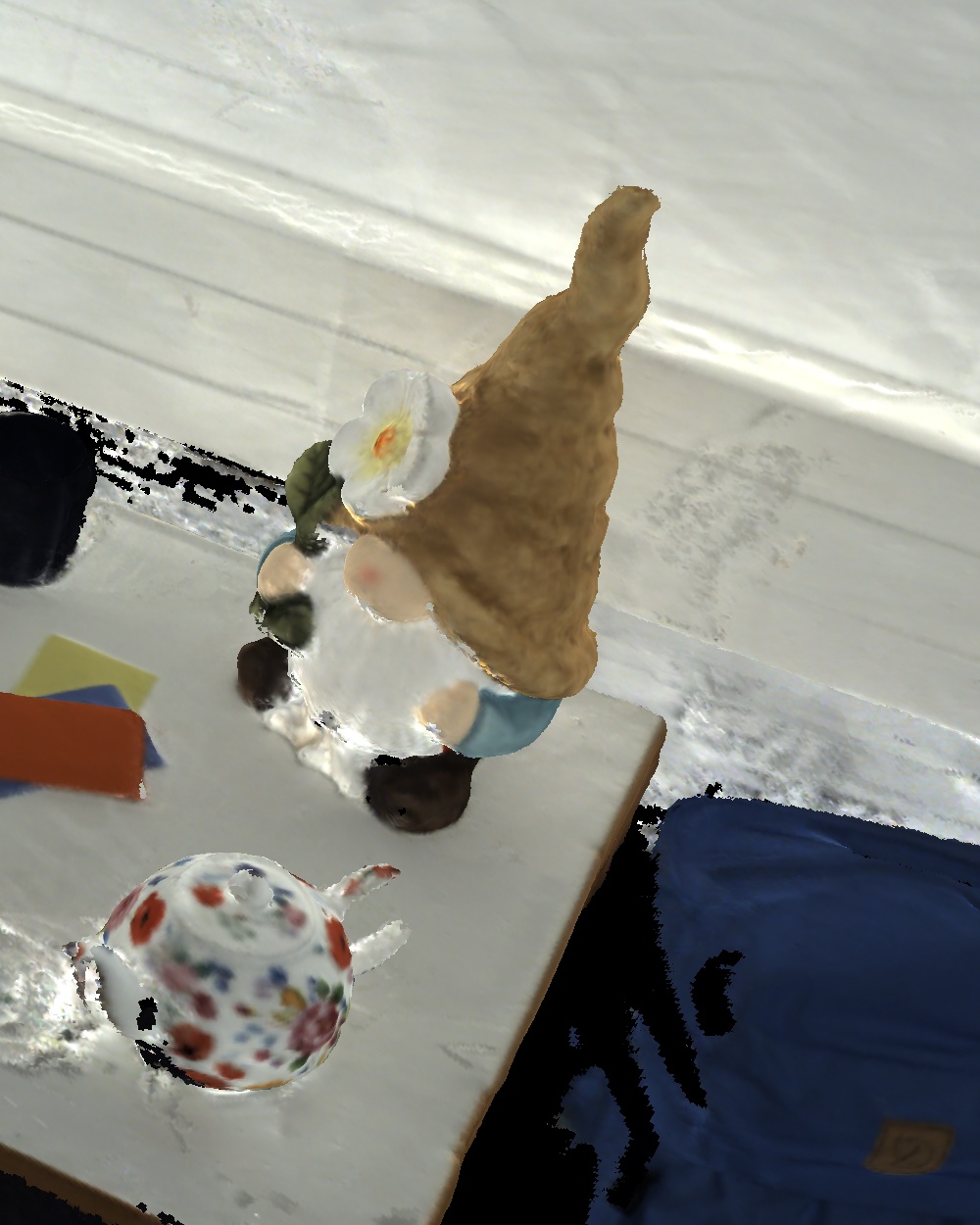} &
		\includegraphics[height=2.8cm]{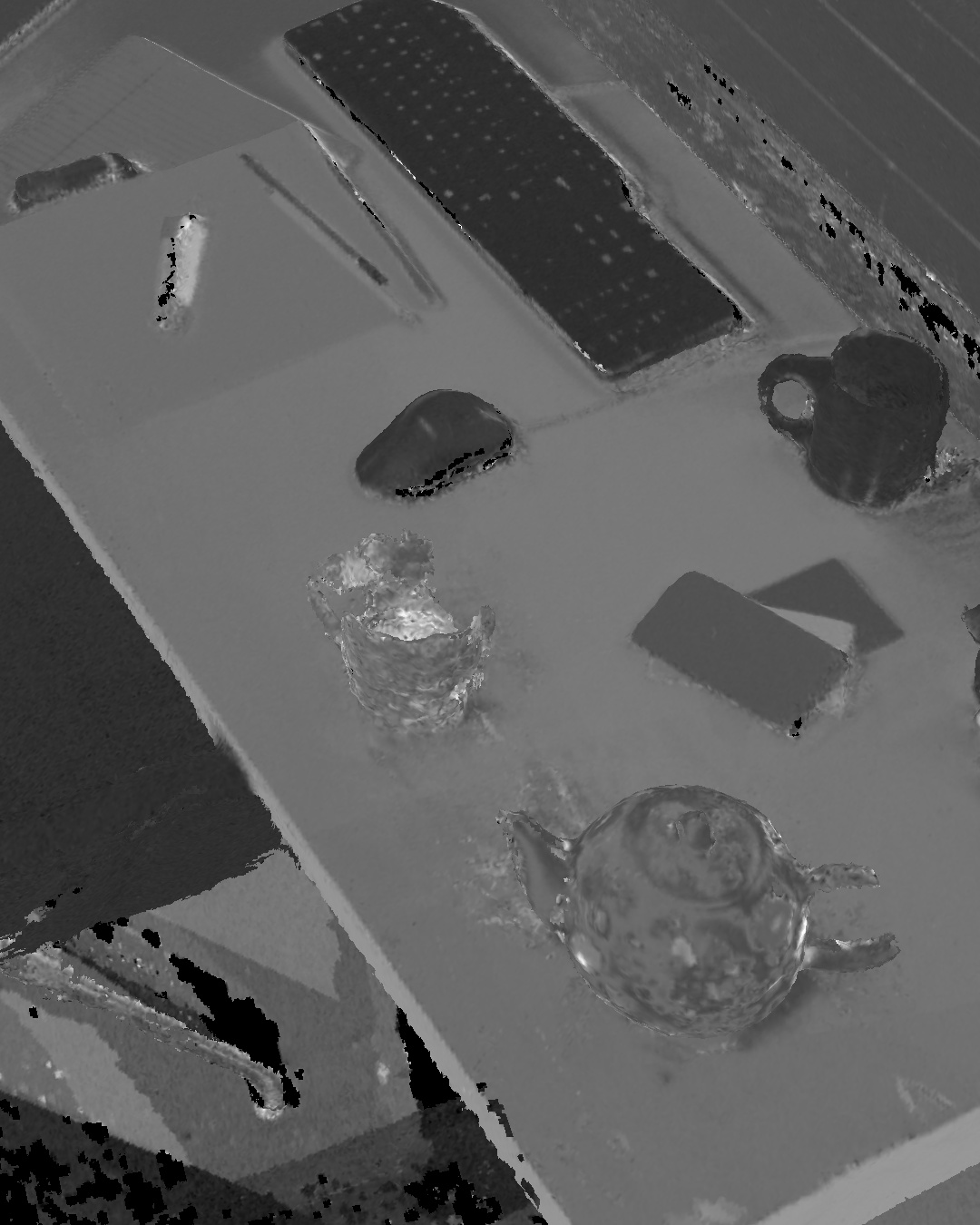} &
		\includegraphics[height=2.8cm]{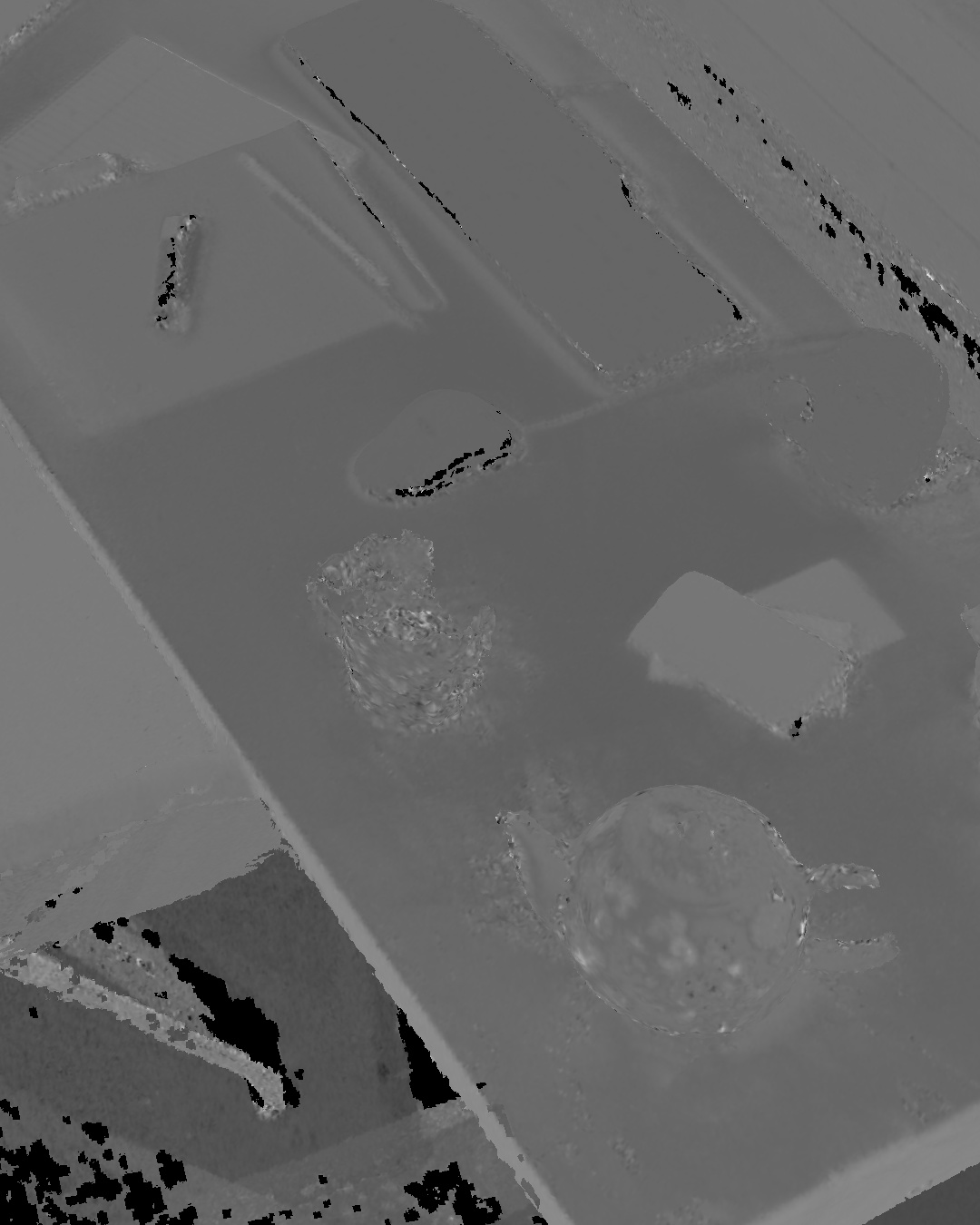} 
		\\		
		& Rendered for Novel Viewpoints & Normals & Geometry & Diffuse Albedo & Specular Albedo & Roughness
	\end{tabular}
	\addtolength{\tabcolsep}{5pt}
	\caption{
		\textbf{Reconstruction Beyond Object-Scale (3D).} 
		We present reconstructions for three challenging scenes that are composed of multiple objects with non-convex shapes, detailed geometries, various different materials, many occlusions and shadows and a spatial size of up to $2 \times 3m$, \eg~the scene `Office'.
		Our method  reconstructs detailed geometry and accurate materials leading to realistic renderings under new illumination and unseen viewpoints.
		Hereby, we observe a clean separation of illumination effects and geometric properties, as the material maps are homogeneous per object parts (see \eg~the `Turtle' in `3 Objects' and `Sofa').
	}
	\label{fig:results_room-scale}
\end{figure*}

\begin{figure*}
	\centering
	\addtolength{\tabcolsep}{-4pt}
	\begin{tabular}{lccccc}
\toprule
{} &                                                                      Duck &                                                                      Gnome &                                                                      Sheep &                                                                      Teapot &                                                                      Turtle \\
\midrule
\rotatebox{90}{$\quad\;$ Prediction}      &  
\includegraphics[height=2.2cm]{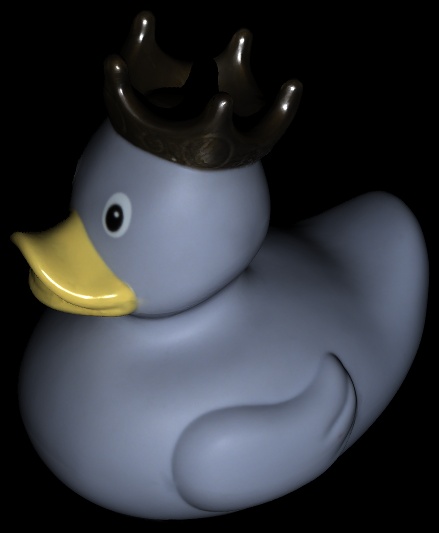} &  \includegraphics[height=2.2cm]{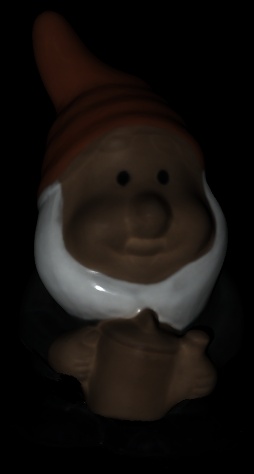} &  \includegraphics[height=2.2cm]{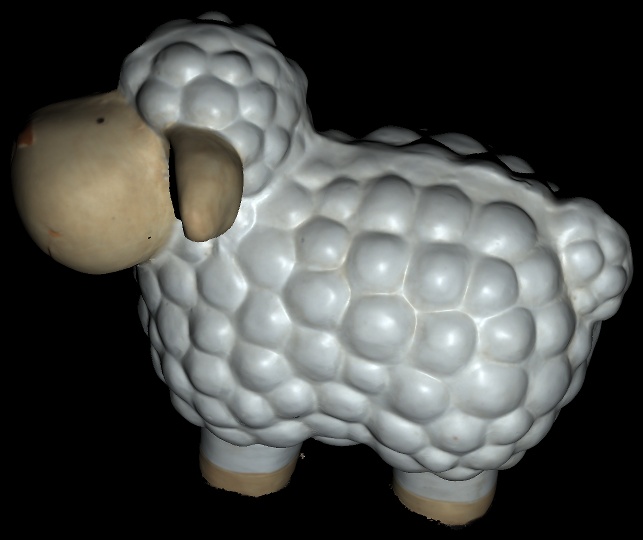} &  \includegraphics[height=2.2cm]{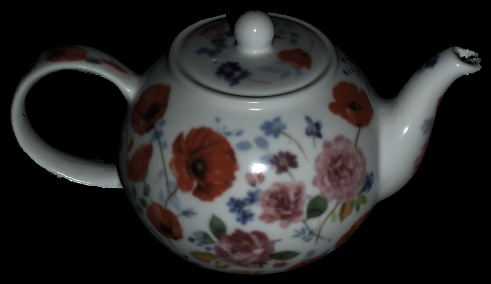} &  \includegraphics[height=2.2cm]{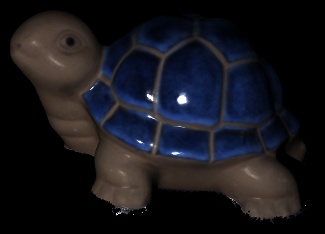} \\
\rotatebox{90}{$\;\;\;$ Observation}     &     
\includegraphics[height=2.2cm]{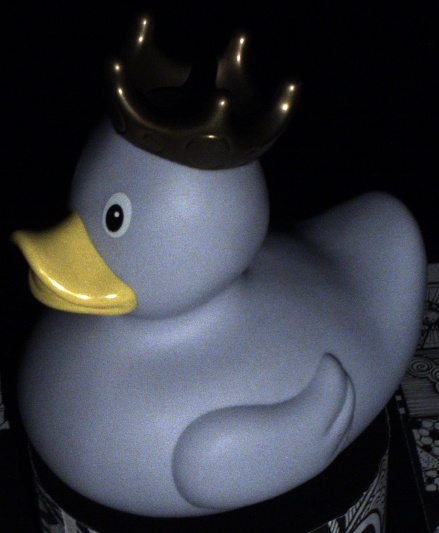} &     \includegraphics[height=2.2cm]{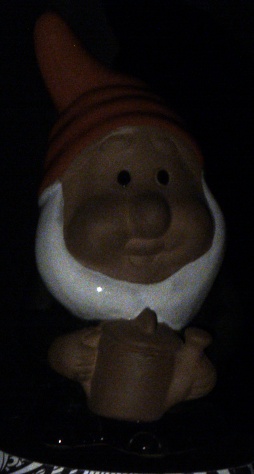} &     \includegraphics[height=2.2cm]{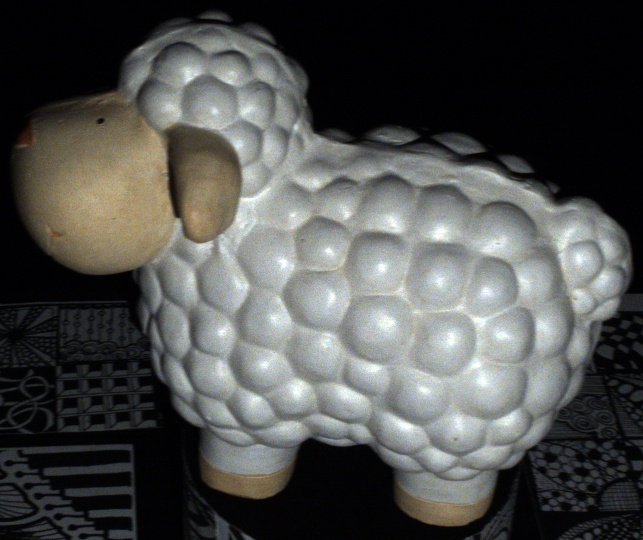} &     \includegraphics[height=2.2cm]{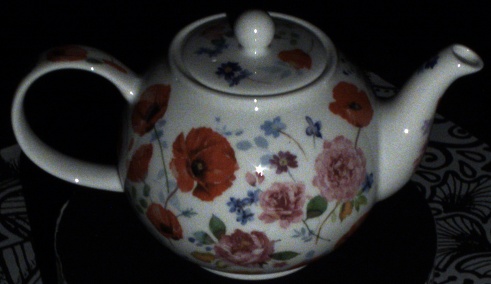} &     \includegraphics[height=2.2cm]{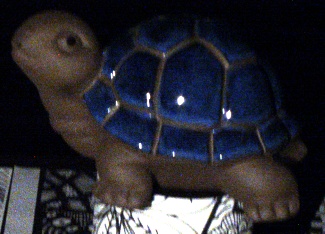} \\
\rotatebox{90}{$\qquad\;$ RMSE}            &         
\includegraphics[height=2.2cm]{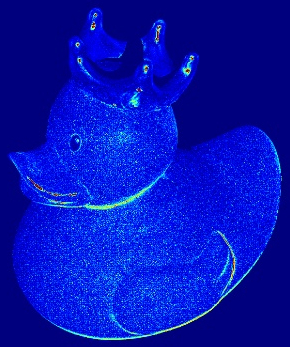} &         \includegraphics[height=2.2cm]{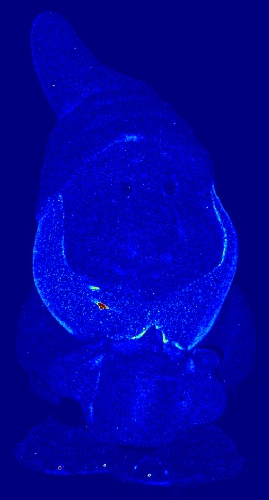} &         \includegraphics[height=2.2cm]{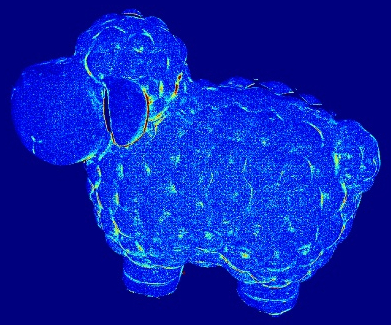} &         \includegraphics[height=2.2cm]{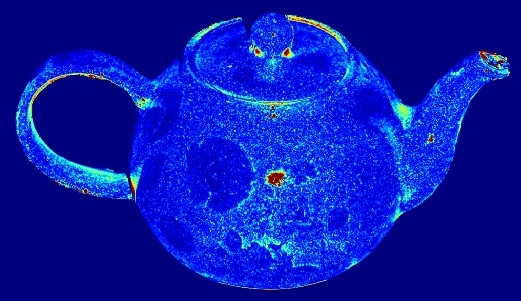} &         \includegraphics[height=2.2cm]{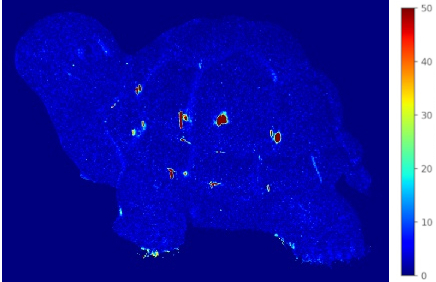} \\
\rotatebox{90}{$\quad\;\;\:$ Normals}         &      
\includegraphics[height=2.2cm]{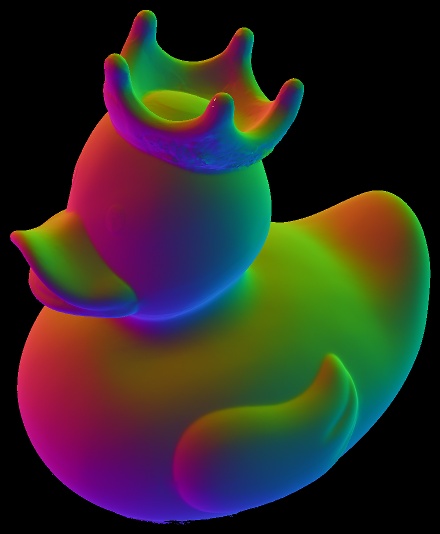} &      \includegraphics[height=2.2cm]{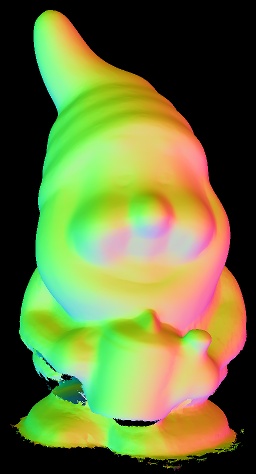} &      \includegraphics[height=2.2cm]{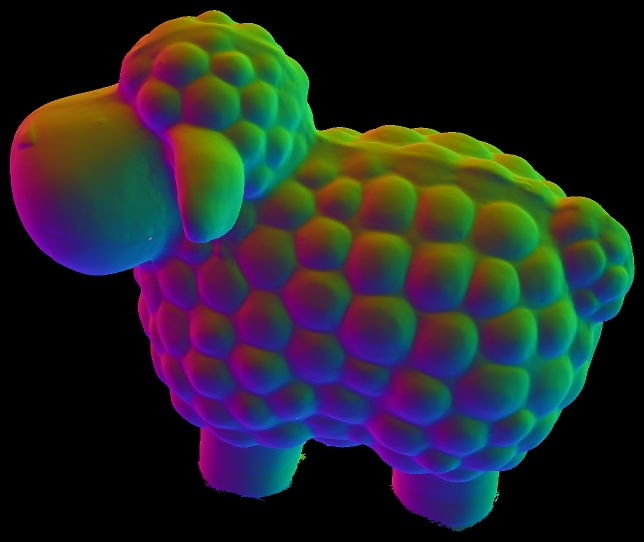} &      \includegraphics[height=2.2cm]{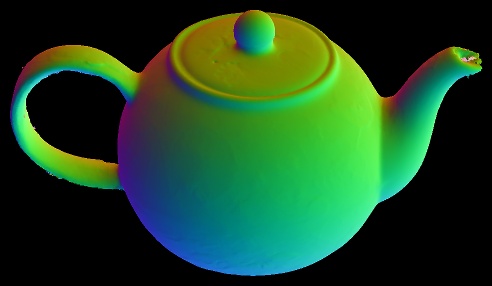} &      \includegraphics[height=2.2cm]{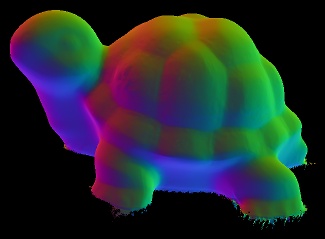} \\
\rotatebox{90}{$\;\:$ Diff. Albedo}  &   
\includegraphics[height=2.2cm]{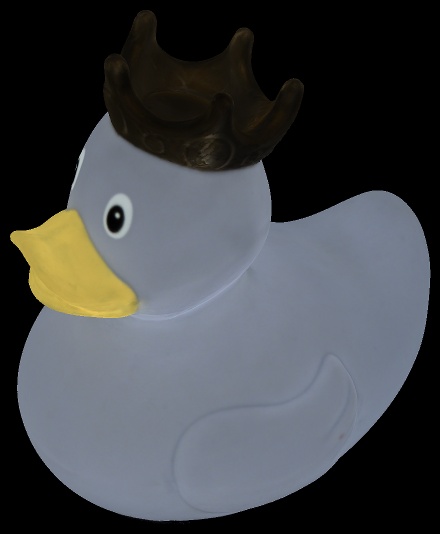} &   \includegraphics[height=2.2cm]{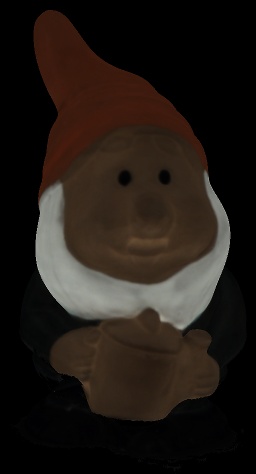} &   \includegraphics[height=2.2cm]{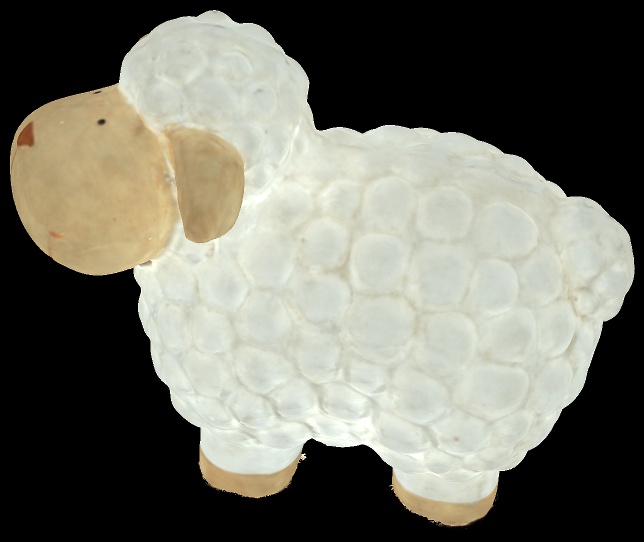} &   \includegraphics[height=2.2cm]{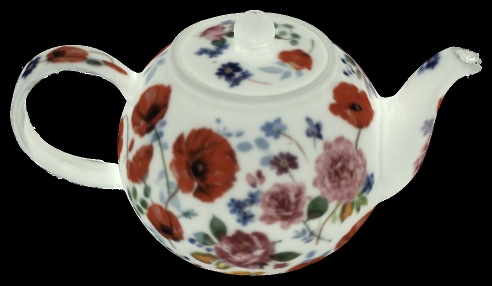} &   \includegraphics[height=2.2cm]{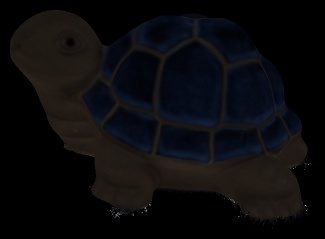} \\
\rotatebox{90}{$\:$ Spec. Albedo} &  
\includegraphics[height=2.2cm]{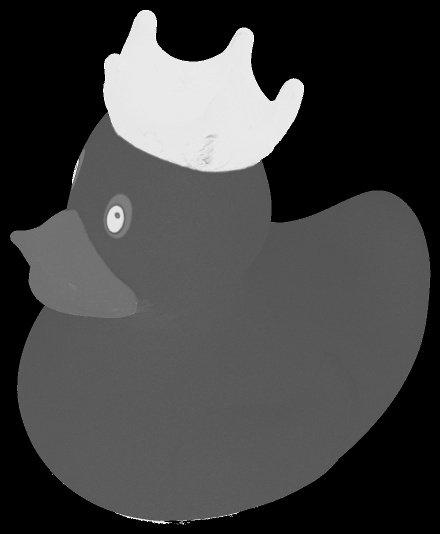} &  \includegraphics[height=2.2cm]{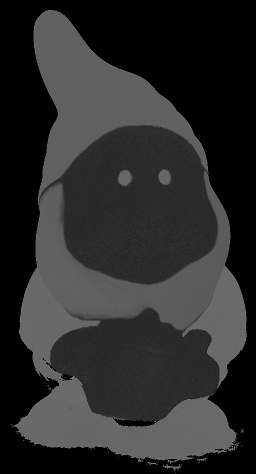} &  \includegraphics[height=2.2cm]{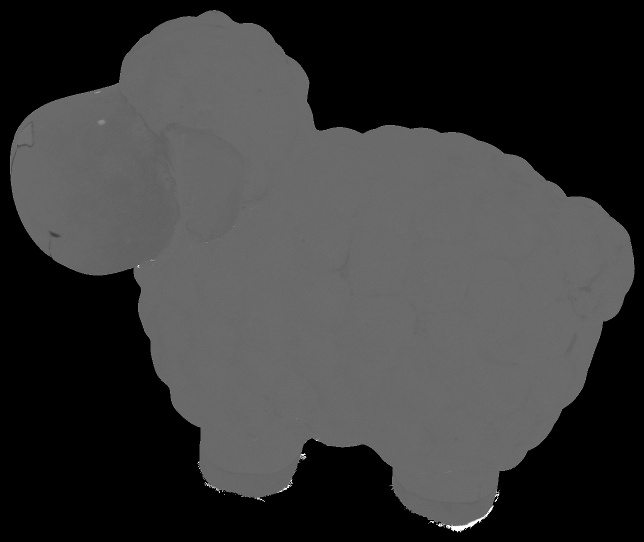} &  \includegraphics[height=2.2cm]{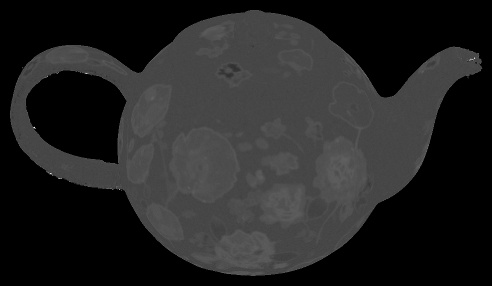} &  \includegraphics[height=2.2cm]{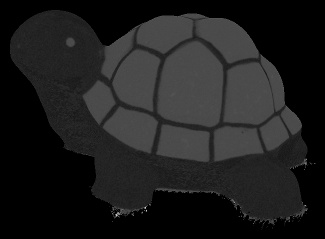} \\
\rotatebox{90}{$\quad\;$ Roughness}       &    
\includegraphics[height=2.2cm]{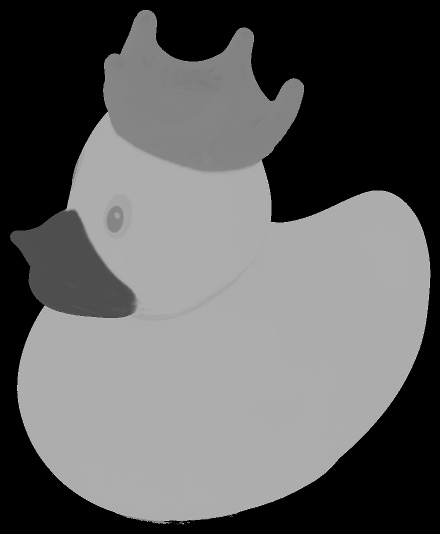} &    \includegraphics[height=2.2cm]{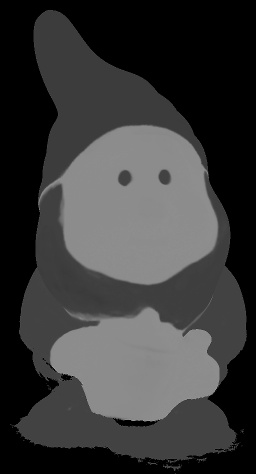} &    \includegraphics[height=2.2cm]{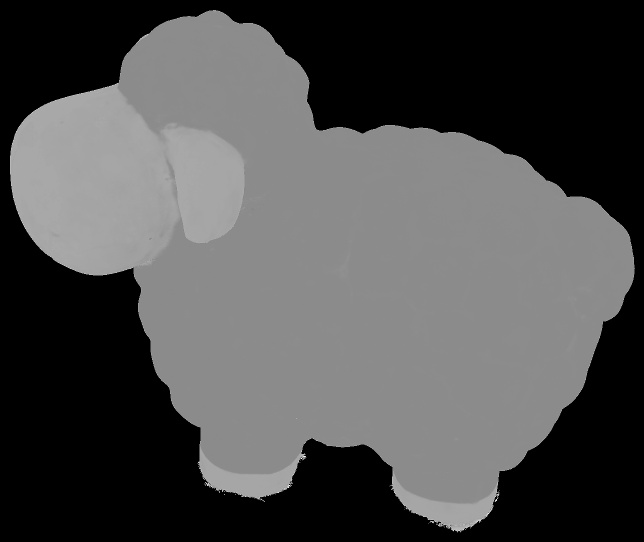} &    \includegraphics[height=2.2cm]{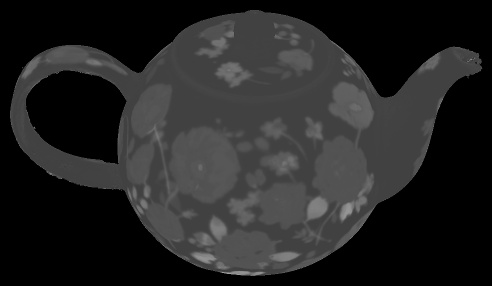} &    \includegraphics[height=2.2cm]{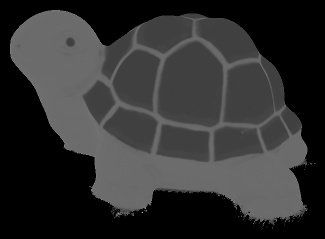} \\
\bottomrule
\end{tabular}

	\addtolength{\tabcolsep}{4pt}
	\caption{
		\textbf{Reconstruction Results (3D)} for the objects `Duck', `Gnome', `Sheep', `Teapot' and `Turtle'.
	}
	\label{fig:results_all_objects1}
	\vspace{-1mm}
\end{figure*}

The proposed method refines depth maps but does not compete missing geometry. Therefore, larger holes in the initial geometry can not be filled, see \eg~the wall in the 'Office', \figref{fig:results_room-scale}.
Further, 
we decided on a renderer that models a single light bounce to keep computation tractable. That means that global illumination cannot be modeled and inter-reflections cause  local errors in the material maps, as can be seen in \figref{fig:results_all_objects1}.
And last,
the expressiveness of our BRDF model is limited. While it is able to represent most objects common in indoor rooms, it does not support anisotropic reflections or subsurface scattering.

\section{Conclusion}
\label{sec:conclusion}

We have proposed a practical approach to estimating geometry and materials from a handheld sensor for full 3D models that exceed object-scale. 
Accurate camera poses are crucial to this task, but are not readily available. 
To tackle this problem, we propose a novel formulation which enables joint optimization of poses, geometry and materials using a single objective. 
Towards large-scale appearance and geometry reconstructions, we represent the scene as 2.5D parameter maps for a set of keyframes and introduce a distributed optimization scheme. 
We demonstrate that our multi-view consistency regularization is key to enable accurate integration of the local 2.5D reconstruction results into a consistent 3D model. 
Our approach recovers accurate geometry and material properties that are globally consistent across the local representations.
We demonstrate on multiple scenes with larger compositions of multiple objects that our method takes a step towards scalable multi-view reconstruction of geometry and materials.
In future work, we plan to extend our model to handle ambient light and global illumination effects.

	\ifCLASSOPTIONcompsoc
	\section*{Acknowledgments}
	\else
	\section*{Acknowledgment}
	\fi
	We thank Giljoo Nam for providing the results of his method \cite{Nam2018SIGGRAPH} on our data.
	We thank the International Max Planck Research School for Intelligent Systems (IMPRS-IS) for supporting Carolin Schmitt.
	This work was supported by the Intel Network on Intelligent Systems (NIS).
	Joo Ho Lee and Andreas Geiger were supported by the ERC Starting Grant LEGO-3D (850533) and the DFG EXC number 2064/1 - project number 390727645.

	\bibliographystyle{IEEEtran}
	\bibliography{bibliography_short,bibliography,bibliography_custom}

\end{document}